\documentclass[11pt,a4paper]{article}

\RequirePackage[T1]{fontenc}
\RequirePackage[utf8]{inputenc}
\RequirePackage{lmodern}
\RequirePackage[a4paper,margin=27mm]{geometry}
\RequirePackage{amsmath,amssymb,amsthm,mathtools}
\RequirePackage{microtype,multicol,enumitem}
\RequirePackage[hidelinks,bookmarksnumbered]{hyperref}
\RequirePackage{subcaption}
\RequirePackage{xcolor}

\theoremstyle{plain}
\newtheorem{theorem}{Theorem}
\newtheorem{proposition}{Proposition}
\newtheorem{lemma}{Lemma}
\newtheorem{corollary}{Corollary}
\newtheorem{conjecture}{Conjecture}
\theoremstyle{definition}

\theoremstyle{remark}

\theoremstyle{assumption}
\newtheorem{assumption}{Assumption}
\allowdisplaybreaks[1]
\newcommand{\tin}{\mathrm{in}}
\newcommand{\tout}{\mathrm{out}}

\newcommand{\EE}{\mathbb{E}}

\newcommand{\NN}{\mathbb{N}}

\newcommand{\PP}{\mathbb{P}}

\newcommand{\RR}{\mathbb{R}}

\newcommand{\Bb}{\mathcal{B}}

\newcommand{\Hh}{\mathcal{H}}

\newcommand{\Ll}{\mathcal{L}}

\newcommand{\Nn}{\mathcal{N}}

\newcommand{\Uu}{\mathcal{U}}
\newcommand{\Vv}{\mathcal{V}}

\newcommand{\Xx}{\mathcal{X}}
\newcommand{\Yy}{\mathcal{Y}}

\usepackage{amsmath,amssymb}
\usepackage{graphicx}
\usepackage{placeins}
\usepackage{booktabs,array,longtable}

\DeclareMathOperator{\HS}{HS}

\DeclareMathOperator{\im}{Im}

\newcommand{\loss}{\mathrm{loss}}

\usepackage{needspace}

\newcommand{\Thetainf}[1]{\Theta_{#1}^{\mathrm{unif}}}

\usepackage{authblk}

\title{Infinite ResNets are Gaussian Mean ODEs}

\author[1]{Louis-Pierre Chaintron}
\author[2]{Lénaïc Chizat}
\author[1]{Javier Maass}

\affil[1]{Institute of Mathematics,
École Polytechnique Fédérale de Lausanne (EPFL), Switzerland}

\affil[2]{Department of Mathematics,
National University of Singapore,
Singapore}

\date{\today}

\begin{document}
\maketitle

\begin{abstract}
Understanding the training dynamics of deep neural networks in large-scale limits is a question both of foundational and of practical importance. Prior works have obtained such descriptions through dynamical mean field theory (DMFT), yielding a system of non-Markovian dynamics that are challenging to analyze and that obscure the optimization structure of the original dynamics. In this paper, we obtain instead a simple description of the large-scale limit training dynamics for ResNets in  standard shape scalings and in the $1/\text{depth}$ branch scaling that allows maximal (local) feature updates. For a model trained with gradient descent (GD) from a random initialization, the limit dynamics is simply GD on a Neural Mean ODE with Gaussian initial weights. Our analysis covers population or empirical risk GD in the joint limit of large width, depth and embedding, input, and output dimensions. We obtain approximation scaling laws that depend on the shape of the ResNet and the covariance spectrum of the data. We also illustrate the flexibility of our approach by discussing Mixture of Expert models, where likewise the limit admits an intuitive (nested) Mean ODE description.
\end{abstract}

\setcounter{tocdepth}{2}
\begin{multicols}{2}
  \small
  \tableofcontents
\end{multicols}

\section{Introduction}\label{sec:introduction}
Over the past decade, the analysis of training dynamics of large-scale
neural networks has addressed architectures of increasing complexity:
from shallow neural networks~\cite{chizat2018global,mei2018mean,
rotskoff2018parameters,sirignano2020mean}, to multi-layer
perceptrons~\cite{yang2021feature,bordelon2022self}, to
ResNets~\cite{lu2020mean,yang2024depth,bordelon2024depthwise,
chizat2025hidden,chaintron2026resnets} and
Transformers~\cite{bordelon2024transformer}.
These analyses aim to guide the scaling of hyperparameters, understand how collective effects emerge from interactions
between neurons, and derive more tractable models of neural network
training dynamics.
For the latter aim, while mean-field limits of shallow networks have provided useful models, the derived limiting equations for deeper networks in
feature-learning regimes have proved difficult  to interpret and
analyze.
In particular, dynamical mean-field theory (DMFT), and its algorithmic formulation Tensor Programs~\cite{yang2021feature}, describe these limits through equations involving the
history of training, where the optimization structure is no longer apparent.
In this paper, we study the large-scale limit of
ResNets~\cite{he2016deep} with two-layer perceptron blocks, motivated
by the residual and depth-two structure also used in
Transformers~\cite{vaswani2017attention}.
We obtain an intuitive and tractable description of the limit as a gradient descent dynamics, together with quantitative convergence rates.

\subsection{Setting}
Consider a supervised learning problem with data distribution $\mu$ on $\Xx\times\Yy$. In our analysis, the input and output spaces may be infinite dimensional; for concreteness, in this introduction take $\Xx=\RR^{d_\tin}$ and $\Yy=\RR^{d_\tout}$. Fix an activation function $\rho:\RR\to\RR$ with $\rho'$ and $\rho''$ bounded. Our model of interest is a residual neural network (ResNet)  with two-layer blocks where the hidden states and output are defined recursively by
\begin{align}\label{eq:resnet}
\widehat h^0(x)&=\widehat W_\tin x,\notag\\
 \widehat h^\ell(x)&=\widehat h^{\ell-1}(x)
 +\frac1{LM}\sum_{j=1}^M\widehat v^{j,\ell}
       \rho\big((\widehat u^{j,\ell})^\top\widehat h^{\ell-1}(x)\big),\\
       \widehat f_{\hat \theta}(x)&=\widehat W_\tout^* \widehat h^L(x).\notag
\end{align}
The shape of this architecture is specified by the depth $L$,
the embedding dimension $D$, and the width $M$ of each residual block. The input and output embeddings are
$\widehat W_\tin\in\RR^{D\times d_\tin}$ and
$\widehat W_\tout\in \RR^{D\times d_\tout}$; the internal weights are
$(\widehat u^{j,\ell},\widehat v^{j,\ell})_{j,\ell}\in(\RR^D\times \RR^D)^{LM}$.
The forward pass uses the adjoint $W_{\tout}^*$ which is simply the transpose matrix in this introduction.
We write $\widehat\theta\in\RR^\Bb$ for the vector of all parameters/weights, where $\Bb=2LMD+(d_\tin+d_\tout)D$. We use hats for the finite ResNets and drop them for the limit model.

Given a family of functions $\loss_y:\Yy\to\RR$, the risk is
\begin{align}\label{eq:loss-resnet}
\widehat  \Ll(\widehat\theta)=\int\loss_y(\widehat f_{\hat \theta}(x))\,d\mu(x,y).
\end{align}
Depending on the choice of $\mu$, this formulation covers a population distribution and a finite training set, in which case we write $n$ for the number of training samples. The objective of this paper is to study a fixed number of gradient descent (GD) steps from a standard random initialization in the large scale regime $L,M,D\to \infty$. We seek to determine the limit dynamics and obtain quantitative convergence bounds that are uniform in $d_\tin, d_\tout$ and $n$.

We focus on the following scaling regime:
\begin{itemize}
 \item \textbf{Initialization scaling.} The initialization scalings allow \emph{maximal local (feature) updates} (MLU), where individual layers asymptotically retain their nonlinear dependence on their learned parameters. This regime is also known as ``Complete Feature Learning''~\cite{dey2025don} or the $1/\text{depth}$ or ODE scaling. Here we use the descriptive terminology MLU because it hints at the relation to the maximal update parameterization and makes no claim on the quality of the learned features. In this paper, we use the somehow unusual convention that the $\ell_2$ norm (rather than the usual RMS norm) of the hidden states in~\eqref{eq:resnet} is $\Theta(1)$. Under this convention, the MLU regime corresponds to iid centered initial internal weights $\hat u^{j,\ell}_0,\hat v^{j,\ell}_0$ with coordinatewise standard deviation of order one, and an initial output embedding $W_\tout$ with bounded operator norm $\Yy\to(\RR^D,\|\cdot\|_2)$\footnote{In practice and previous works, it is more common to use hidden states of RMS norm in $\Theta(1)$ (which is also more suitable to low-precision computations). Both conventions are equivalent up to fixed rescalings, and lead to the same dynamics of predictors. Here we opted for $\ell_2$ normalization as it simplifies notations. } Our analysis requires the slightly stronger condition of bounded Hilbert-Schmidt norm for $W_\tout$.
 \item \textbf{Learning-rate scaling.} We adopt the unique scaling of block-wise learning rates (LR) that ensures that each block contributes asymptotically to loss decay, as in~\cite{chizat2024featurespeed}\footnote{For gradients computed in the ordinary $\ell_2$ geometry, the LR scalings for the internal weights, input embedding, and output embedding are respectively $\eta LM$, $\eta$, and $\eta$ for some fixed $\eta$. See also Section~\ref{sec:main-result}.}.
 \item \textbf{Shape scaling.} The shape satisfies $D\ll LM$ (up to logarithmic factors, ignored in this introduction). This covers the  standard practice of using $D\propto M$ and $L\to \infty$.
\end{itemize}

The MLU initialization scale and the shape transition at $D\propto ML$ were determined
in~\cite{chizat2025hidden} for ResNets in the large-$L,M,D$ regime,
building on many prior work on depth scaling and feature
learning, in particular~\cite{yang2024depth,bordelon2024depthwise,dey2025don} which noticed that a lazy regime can also arise with depth-scaling. In the MLU regime and with this shape constraint, averaging across units and depth makes the forward and backward states asymptotically deterministic conditional on the initial embeddings.

\subsection{The limit model}\label{sec:intro-limit-model}

\begin{figure}[t]
 \centering
 \includegraphics[width=\textwidth]{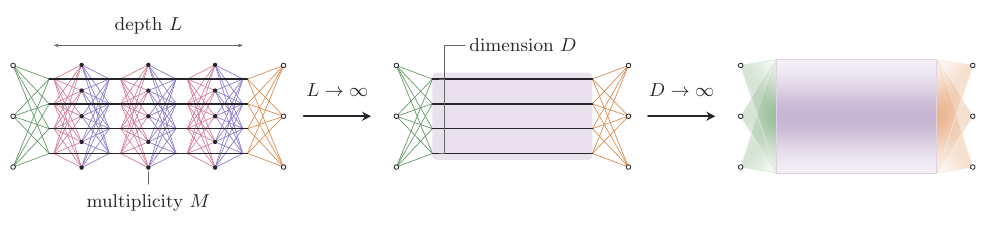}
 \caption{From finite ResNets to finite-dimensional Mean ODEs, to Gaussian Mean ODEs.}
 \label{fig:resnet-limits}
\end{figure}

Our main theorems deal with joint $L,M,D\to \infty$ limits. In proofs, this limit is realized in two stages, illustrated in Figure~\ref{fig:resnet-limits}.

\paragraph{First stage of the limit: Mean ODE.}
 In the first stage, as depth diverges ($L\to \infty$), the forward pass becomes an ODE and the average over $M$ units in each layer becomes an expectation, as proved in~\cite{chizat2025hidden}. We used AI to strengthen this result, see Section~\ref{sec:finite-limit}.  The model after this stage\footnote{All our convergence statements, simplified for the purpose of this introduction, are convergence in probability of the training dynamics (signal and weights) after a bounded number of training steps.} is a \emph{Mean ODE} in $\RR^D$, defined as
\begin{align}\label{eq:MeanODE-D}
h_D(0,x)&=W_{\tin,D}x,\notag\\
 \partial_s h_D(s,x)&=
 \EE\big[v_D(s)\rho(u_D(s)^\top h_D(s,x))\big],
 \qquad s\in[0,1],\\
 f_D(x)&=W_{\tout,D}^*h_D(1,x).\notag
\end{align}
Here the expectation $\EE$ is taken over the internal neuron randomness $(u_D,v_D)$, so conditioning on the  embeddings $W_{\tin, D}$ and $W_{\tout, D}$ the forward pass is deterministic for a given input. The block width $M$ has disappeared from this formulation: even if $M=1$, taking $L\to \infty$ makes the model behave as if $M\to \infty$, at least if $D\ll LM$.

\paragraph{Second stage of the limit: Gaussian Mean ODE.}
The second stage takes the embedding dimension to infinity ($D\to \infty$). While the first stage relied only on the residual structure of the architecture, this stage uses a more specific structure of the model~\eqref{eq:resnet}: the nonlinearity $\rho$ is never directly applied to the hidden representations $h^{\ell}$; instead $h^{\ell}$ only interacts with projections against internal weights or the adjoint operation. As a consequence, for an independent Gaussian initialization of the internal weights following the MLU scaling
\begin{equation}\label{eq:gaussian-init}
 \widehat u^{j,\ell}_0\sim\Nn(0,\sigma_u^2I_D),
 \qquad
 \widehat v^{j,\ell}_0\sim\Nn(0,\sigma_v^2I_D),
\end{equation}
the distribution of these initial projections is invariant under rotations of the hidden space $\RR^D$. Because of this symmetry and since the contribution of the initial residual blocks vanish, we can show that the hidden state never leaves the subspace $S= \im(W_\tin) + \im(W_\tout) \subset \RR^D$ during training (Lemma~\ref{lem:subspace}).

This remark is the starting point of our analysis: we show that the the limit model is exactly a Mean ODE~\eqref{eq:MeanODE-D} in dimension $r= d_\tin+d_\tout$ with the Gaussian initial internal weights~\eqref{eq:gaussian-init}. For independent Gaussian embedding matrices with entry variance $1/D$,\footnote{For tied initial embeddings $W_{\tin}=W_\tout$, with $d_\tin=d_\tout=r$, the limit can be represented in $\RR^r$ with both embeddings initialized with the identity matrix.} the initial embedding matrices in the limit dynamics are the isometric inclusions
\begin{align}\label{eq:GMODE-reduced}
W_{\mathrm{in}}= \begin{bmatrix}I_{d_{\mathrm{in}}}\\0\end{bmatrix},
\qquad
W_{\mathrm{out}}=\begin{bmatrix}0\\ I_{d_{\mathrm{out}}}\end{bmatrix}.
\end{align}
Importantly, we show that this limit holds even when the embeddings are trained and if the initial distribution of the internal weights is not Gaussian.

\paragraph{Hilbert space representation.} In Section~\ref{sec:population}, we introduce a more general formulation of the Gaussian Mean ODE where the hidden states belong to a Hilbert space that is potentially infinite-dimensional. This enables infinite dimensional input and/or output spaces $\Xx$ and $\Yy$ and to derive scaling laws. Moreover, this formulation present some conceptual advantages:
\begin{itemize}
\item Both the Hilbert and the reduced formulation~\eqref{eq:GMODE-reduced} yield the same dynamics of the output predictor. However, the behavior of the internal weights, including their diverging initial norm, is more faithfully captured by the Hilbert space formalism, where it becomes necessary to separate the initial weights from their updates, which are objects of a different nature. The reduced formulation only shows the relevant $r$-dimensional slice of the limit model.
\item The finite-dimensional Mean ODE~\eqref{eq:MeanODE-D} with Gaussian initialization is a particular case of that general construction with $\Hh_D=(\RR^D,\langle\cdot,\cdot\rangle_2)$. Fixing an abstract Hilbert space $\Hh$ is convenient as we can isometrically embed the dynamics with growing dimensions $D$ and the limit in the same space. By abstracting away the embedding space, this formulation clarifies that the limit Mean ODE evolves in a geometry determined by the covariance structure of the data and that the Gaussian Mean ODE is better view as a dimension-free object.
\end{itemize}

\subsection{Contributions}
Our main contributions are the following:
\begin{itemize}
\item  \textbf{Quantitative limit.} Our main result is the convergence of the training dynamics of ResNets towards that of the Gaussian Mean ODE (Theorem~\ref{thm:main}). Over bounded training horizons, we obtain errors between the output of the ResNet and its infinite limit of the following form (with high probability, up to logarithmic factors):
\begin{equation}\label{eq:intro-rate}
 \frac1L+\sqrt{\frac{D}{LM}}
 +\min_{0\le r\le D}
       \left\{\left(\sum_{j>r}\lambda_j\right)^{1/2}
                       +\sqrt{\frac rD}\right\}
\end{equation}
where $\lambda_1>\lambda_2>\dots$ are the eigenvalues of a data-dependent covariance operator.
The terms respectively measure the depth discretization, the sampling of $LM$ neurons, and the approximation of the initial embeddings. The two first terms are as in~\cite{chizat2025hidden} and the new last term can be seen as a geometry distortion error, it recovers the rate $O(1/\sqrt{D})$ from~\cite{chaintron2026resnets} for finite-dimensional data. Intuitively, it expresses that only a few principal directions can be faithfully represented when $D$ is small. The term $\sqrt{\frac{r}{D}}$ is due to the randomness of Gaussian embeddings and/or the non-Gaussianity of the initialization. It disappears when using PCA embeddings instead and Gaussian internal initialization (then take $r=D$ in the bound). %In Section~\ref{sec:non-gaussian} we show that non-Gaussian internal initialization leads to the same limit, with an additional error term $(r/D)^{\alpha}$ with $\alpha=1$ if the initialization is symmetric and $\alpha=1/2$ otherwise.
\end{itemize}
Our analysis yields in fact a slightly sharper version of~\eqref{eq:intro-rate} (see Proposition~\ref{prop:output-stability}) which we observe empirically to be tight for the errors in hidden space. For the error on outputs, we conjecture and observe empirically a faster decay than this upper bound, see Conjecture~\ref{conj:prediction}. Our motivation for obtaining quantitative rates is not that we believe that the precise scaling is important, but rather that (i) a sharp rate indicates that the theory properly captures the mechanism behind convergence, and (ii) quantitative bounds directly show which limits commute, and further inform on the role of the shape scaling. We complement this main result with two side results:
\begin{itemize}
\item
\textbf{Quasi-smoothness of the limit objective.} To illustrate the usefulness of a having tractable limit model, we analyse the smoothness of the limit objective, see Section~\ref{sec:regularity-gradient}. Besides analyses in the lazy regime, we are not aware of any other scale-free (in particular uniform in $D$) smoothness results for such architectures.
\item
\textbf{Mixture of Expert architecture.} To demonstrate the flexibility of our approach, we write at a formal level the limit model for the mixture of expert limit in the same formalism, see Section~\ref{sec:experts}.
\end{itemize}

\subsection{Related work}

Following the mean-field description of shallow networks~\cite{chizat2018global,mei2018mean,rotskoff2018parameters,sirignano2020mean,nitanda2017stochastic}, several works have proposed mean-field descriptions of deep residual architectures~\cite{lu2020mean,ding2022overparameterization}. However, these analyses required $M$ to diverge while $D$ remained fixed, which makes each layer converge to its infinite-width mean-field limit. As this differs from practical shape scalings where $M\propto D$, and where each layer \emph{does} keep fluctuations of constant order, this limit was generally regarded as unrealistic.

A parallel line of work uses dynamical mean-field theory (DMFT) to characterize feature-learning limits at infinite width \cite{yang2021feature,bordelon2022self}, with subsequent works studying the large-depth limit \cite{yang2024depth,bordelon2024depthwise,bordelon2024transformer}.  These approaches cover a broad range of architectures and parameterizations, including scalings with stochastic continuous-depth limits, which are outside the scope of the present paper. However, they have also shortcomings: they are difficult to fully justify, require $D\propto M$ and most importantly yield limit systems that are difficult to analyse, involving correlations across training steps, and do not have \emph{a priori} a gradient-descent structure.

Our work connects these perspectives and continues the program developed in \cite{chizat2025hidden,chaintron2026resnets}. First, \cite{chizat2025hidden} established that the training dynamics of deep ResNets converge to those of Mean-Field Neural ODEs---which we call  Mean ODEs to emphasize on the different convergence mechanism---under practical shape scalings, even though individual layers need not converge to their mean-field limits. Second, in~\cite{chaintron2026resnets} we analyzed the subsequent limit $D\to\infty$ and established quantitative convergence and universality for a training set of fixed size. There, the limiting system has a DMFT formulation corresponding to a special case of the dynamics derived in \cite{bordelon2024transformer}.

The present paper identifies this large-scale limit as gradient descent on a dimension-free Gaussian Mean ODE. This formulation makes the optimization structure explicit and reduces the large-$D$ analysis under Gaussian internal initialization to stability with respect to the initial embeddings. It also yields quantitative bounds that reflect the spectral structure of the embeddings and are uniform in the input and output dimensions and the number of training examples. For non-Gaussian initialization, we build on the  skeleton map and Lindeberg replacement arguments of \cite{chaintron2026resnets} and refine the dependence on the effective rank.

\subsection{Notation}
We use hats for quantities associated to finite ResNets, a subscript $D$ for finite-dimensional
Mean ODEs, and plain symbols for the large-scale limit model.
The indices $k,\ell,j$ denote a training step, a discrete layer and a unit/neuron, and
$s\in[0,1]$ is the continuous depth parameter. For $a,b\in\RR^D$, we use the standard $\ell_2$ geometry:
\[
 \langle a,b\rangle_2=a^\top b,\qquad
 \|a\|_2=(\textstyle\sum_{d=1}^Da_d^2)^{1/2},\qquad
 \Hh_D=(\RR^D,\langle\cdot,\cdot\rangle_2).
\]
All Hilbert spaces are real and separable.
Unless otherwise specified, $\EE$ takes the expectation over internal neurons randomness only; data averages are written $\int\cdot\,d\mu$.
Constants $C_{B,K}$ may change between lines and do not depend on the shape parameters, dimensions or sample size.

\subsection{Methodology and AI use} This work began with our realization that, conditioned on the embeddings, the Gaussian Mean ODE limit is achieved at finite $D$ when $d_\tin,d_\tout$ are finite (Lemma~\ref{lem:subspace}). It then became clear to us that (i) all limits could be handled by embedding the dynamics in a common Hilbert space and using stability arguments, and (ii) that GD on this model was just one integration by part away from the DMFT equations (Section~\ref{sec:dmft}). Given this strategy, the results in our main thread follow from arguments that are either routine or adapted from our previous works~\cite{chizat2025hidden, chaintron2026resnets}. They were obtained by interactions with AI (GPT-6 Astra), mostly just to speed up the writing and verification process. However, AI's contribution has been critical for two results which are improvements of our previous analyses in~\cite{chizat2025hidden, chaintron2026resnets}:
\begin{itemize}
    \item \textbf{From ResNets to Mean ODE.} The convergence of finite ResNets to Mean ODEs in our regime with the rate $O(1/L+\sqrt{D}/\sqrt{ML})$ was proved in~\cite{chizat2025hidden}. In Proposition~\ref{thm:finite-mean-ode}, we present a strengthening of this result with fewer assumptions. Some of the changes, like the removal of the backward-pass clips and handling the population case, are mere technicalities. However, the removal of the condition $D\lesssim M$ is a significant innovation compared to~\cite{chizat2025hidden}. We had previously considered the possibility of grouping layers to avoid this assumption (see Lemma~\ref{lem:finite-apriori}), but quickly realized that preserving the sharp $D$ dependence would be challenging and require a delicate tracking of the dependencies in the initial randomness. The complete argument was found by GPT-6 Astra in one prompt.
    \item  \textbf{Universality/Gaussianization.} Our prior work~\cite{chaintron2026resnets} dealt with general centered subgaussian initialization, proving universality of the large $D$ limit. However, back then we did not attempt to obtain uniform error bounds in $n$. To refine the analysis and extend it to our present setting, we asked AI to derive the Gaussianization argument in the present setting. Several back-and-forths were needed to obtain the stated, empirically sharp, dependency in $r/D$. The argument uses coordinate-by-coordinate Lindeberg replacement with derivatives propagated through the training dynamics as in our previous work~\cite{chaintron2026resnets}, but with subtle moment bounds to obtain the sharp rate (Lemmas~\ref{lem:seed-responses} and~\ref{lem:replacement}).
\end{itemize}

\section{Gradient Descent on dimension-free Mean ODEs}\label{sec:population}
In this section, we define the Mean ODE and study the dimension-free stability of its gradients. We also connect the GD dynamics with its existing DMFT description.

\subsection{Mean ODE in Hilbert space}
Let $\Xx,\Yy,\Hh$ be the input, output and hidden Hilbert spaces (assumed real and separable), and let
$\mu$ be the data probability law on $\Xx\times\Yy$. We write $z=(x,y)$
for a data sample.

\paragraph{Initial weight projections.}
Let $(\Omega,\mathcal F,\PP)$ be a probability space handling the randomness of the initial internal weights. Unless otherwise specified, $\EE$ always means expectation with respect to $\PP$ on $\Omega$.

Fix bounded linear maps $u_0,v_0:\Hh\to L^2(\Omega;\RR)$ of operator norm at
most $B$. For $a\in\{u,v\}$, the scalar random variable $a_0(h)$ describes
the projection of an initial weight against a deterministic hidden vector
$h$. Its adjoint satisfies
\begin{equation}\label{eq:initialization-adjoint}
 \langle a_0^*\phi,h\rangle_\Hh=\EE[\phi a_0(h)],
 \qquad \|a_0^*\phi\|_\Hh\le B\|\phi\|_{L^2(\Omega)},\qquad a\in\{u,v\}.
\end{equation}
When the map is represented by an initial weight in $L^2(\Omega;\Hh)$,
we use the same notation $a_0$ for that random vector, so that
$a_0(h)=\langle a_0,h\rangle_\Hh$ and $a_0^*\phi=\EE[\phi a_0]$.
This identification applies in particular in finite dimension. The operator
formulation also permits initial projections that do not come from an
$\Hh$-valued random vector, as in the Gaussian case (see end of section).

\paragraph{Parameters and embeddings.}

Write $\theta=(\theta_0,\Delta\theta)$, where
$\theta_0=(W_{\tin,0},W_{\tout,0},u_0,v_0)$ is the initialization and
$\Delta\theta=(\Delta W_{\tin},\Delta W_{\tout},\Delta u,\Delta v)$
contains the learned increments. The initialization is held fixed during
training. The input and output embeddings are
\begin{equation}\label{eq:initial-feature}
 \begin{aligned}
 W_{\tin}(x)&=W_{\tin,0}(x)+\Delta W_{\tin}x,&
 W_{\tout}&=W_{\tout,0}+\Delta W_{\tout},\\
 \Delta W_{\tin}&\in\HS(\Xx,\Hh),&
 \Delta W_{\tout}&\in\HS(\Yy,\Hh).
 \end{aligned}
\end{equation}
Here $W_{\tin,0}:\Xx\to\Hh$ is measurable (not necessarily linear) and
$W_{\tout,0}\in\mathcal B(\Yy,\Hh)$ is bounded and linear. Keeping
$W_{\tin,0}$ separate permits an initial embedding that is not
Hilbert--Schmidt (HS), which is useful since standard random embeddings
are not uniformly HS. We use the outer-product notation $(a\otimes_\Uu b)w=\langle b,w\rangle_\Uu a$,
so $\|a\otimes_\Uu b\|_{\HS}=\|a\|_\Vv\|b\|_\Uu$.
For $2\le p<\infty$, the space of increments is
\begin{equation}\label{eq:parameter-hilbert}
 \Theta_p=\HS(\Xx,\Hh)\oplus\HS(\Yy,\Hh)
 \oplus L^p([0,1]\times\Omega;\Hh)^{\oplus2},
\end{equation}
with the product norm
\begin{equation}\label{eq:parameter-norm}
 \|\Delta\theta\|_{\Theta_p}^2
 =\|\Delta W_{\tin}\|_{\HS}^2+\|\Delta W_{\tout}\|_{\HS}^2
 +\|\Delta u\|_{L^p(ds\,d\PP)}^2+\|\Delta v\|_{L^p(ds\,d\PP)}^2.
\end{equation}
In particular, $\Theta_2$ is exactly the Hilbert geometry used for GD.
The initialization components carry respectively the $L^2(\mu;\Hh)$ norm,
the operator norm, and the operator norms
$\mathcal B(\Hh,L^p(\Omega))$ for $u_0,v_0$ whenever these moments are used.
For comparisons with common $u_0,v_0$, we write
\begin{equation}\label{eq:full-parameter-distance}
 \|\theta-\widetilde\theta\|_p^2
 =\|W_{\tin,0}-\widetilde W_{\tin,0}\|_{L^2(\mu;\Hh)}^2
 +\|W_{\tout,0}-\widetilde W_{\tout,0}\|_{\mathrm{op}}^2
 +\|\Delta\theta-\Delta\widetilde\theta\|_{\Theta_p}^2.
\end{equation}

We do not include different initial internal weights in this norm; the stability under different internal
initialization is treated separately in Section~\ref{sec:non-gaussian}.

\paragraph{Regularity assumptions.}
Let $\loss_y:\Yy\to\RR$ be a differentiable loss indexed by the data label $y\in \Yy$ and such that $y\mapsto\loss_y(a)$ and $y\mapsto\nabla\loss_y(a)$ are measurable for every fixed $a$.

\begin{assumption}[Data, loss and activation]\label{ass:data-loss}\label{ass:bounded-data}
For some $B\ge1$,
\[
 \|x\|_\Xx+\|y\|_\Yy+\|{W_{\tin,0}}(x)\|_\Hh
 +\|\nabla\loss_y(0)\|_\Yy\le B\quad\text{for $\mu$-a.e. }z,
 \qquad\|{W_{\tout,0}}\|_{\mathrm{op}}\le B,
\]
and $\int|\loss_y(0)|\,d\mu<\infty$. The loss gradient is uniformly
Lipschitz:
\[
 \|\nabla\loss_y(a)-\nabla\loss_y(\widetilde a)\|_\Yy
 \le B\|a-\widetilde a\|_\Yy,\qquad a,\widetilde a\in\Yy.
\]
The activation satisfies $\rho\in C^2(\RR)$ and
$|\rho(0)|+\|\rho'\|_\infty+\|\rho''\|_\infty\le B$.
\end{assumption}

\paragraph{Definition of the Mean ODE in Hilbert space.}
For $\theta=(\theta_0,\Delta\theta),\ \Delta\theta\in\Theta_2$, define the learned projection maps $A_a(s):\Hh\longrightarrow L^2(\Omega;\RR)$ and their
adjoints, for $a\in \{u,v\}$, by
\begin{align}
 A_a(s)h&=a_0(h)+\langle\Delta a(s),h\rangle_\Hh,
 &A_a(s)^*\phi&=a_0^*\phi+\EE[\phi\Delta a(s)] \label{eq:learned-projections}
\end{align}
Their operator norm is bounded by $ B+\|\Delta a(s)\|_{L^2}$. Let us insist that we do not assume independence between the increments and initialization.

Then the Mean ODE in Hilbert space is defined, with $s\in [0,1]$ as
\begin{align}\label{eq:hilbert-forward}
 h(0,x)&={W_{\tin}}(x),& p(s,x)&=A_u(s)h(s,x),\notag\\
 \partial_s h(s,x)&=A_v(s)^*\rho(p(s,x)),&f_\theta(x)&={W_{\tout}}^*h(1,x),
\end{align}
The hidden state is deterministic with respect to $\omega \in \Omega$.
It is sufficient to use a single pair $(u_0,v_0)$ for every depth (rather than independent copies for each $s\in [0,1]$) because only the marginal law at that depth enters the equations. Finally, the objective for the limit model is
\begin{equation}\label{eq:population-risk}
 \Ll(\theta)=\int\loss_y(f_\theta(x))\,d\mu(z).
\end{equation}

\paragraph{Gaussian initialization.}
We say that the model is a \emph{Gaussian Mean ODE} when $(u_0,v_0)=(G_u,G_v)$, where
$G_u,G_v:\Hh\to L^2(\Omega;\RR)$ are independent centered Gaussian linear maps with
\begin{equation}\label{eq:isonormal}
 \EE[G_a(h)G_a(\widetilde h)]
 =\sigma_a^2\langle h,\widetilde h\rangle_\Hh,
 \qquad a\in \{u,v\},\quad0\le\sigma_a\le B.
\end{equation}
Such a map $G_a$ is known as an isonormal process or a Gaussian white noise in the literature~\cite{Legall}. Observe that $\Vert G_a\Vert_{\mathrm{op}}\leq \sigma_a$. In finite dimension, this is just an inner-product with a Gaussian vector. In infinite dimension, such a map can be constructed by taking an orthonormal basis $(e_j)$ and independent standard Gaussian
variables $(\xi_j^u,\xi_j^v)$, and letting
$G_a(h)=\sigma_a\sum_j\xi_j^a\langle h,e_j\rangle$, with convergence
in $L^2(\Omega)$.
Its adjoint is characterized by
\begin{equation}\label{eq:gaussian-adjoint}
 \langle G_a^*\phi,h\rangle_\Hh=\EE[\phi G_a(h)],
 \qquad \|G_a^*\phi\|_\Hh\le\sigma_a\|\phi\|_{L^2}.
\end{equation}
In infinite dimension and for $\sigma_a>0$, the formal Gaussian vector does not belong to $\Hh$, but its projections nevertheless remain well defined. The isometry property of Gaussian initial weights (up to a scalar rescaling by $\sigma_a$) is crucial for our analysis. Moreover, Gaussianity allows explicit integration by parts, yielding the connection to DMFT equations below.

\subsection{Gradient descent dynamics}

With a slight abuse of notation, $\nabla\Ll(\theta)$ denotes the gradient with respect to $\Delta\theta$, with $\theta_0$ held fixed. It uses the Hilbert inner product of $\Theta_2$, inherited from the finite ResNet $\ell_2$ training geometry:

\begin{equation}\label{eq:gradient-pairing}
 \langle\xi,\zeta\rangle_2
 =\langle\xi_{W_{\tin}},\zeta_{W_{\tin}}\rangle_{\HS}+\langle\xi_{W_{\tout}},\zeta_{W_{\tout}}\rangle_{\HS}
 +\int_0^1\EE[\langle\xi_u,\zeta_u\rangle_\Hh
                         +\langle\xi_v,\zeta_v\rangle_\Hh]\,ds.
\end{equation}

The following lemma gives the expression of the gradient. Its proof in Appendix~\ref{ssec:gradient_expression_proof} follows the standard adjoint state computations from optimal control.
\begin{lemma}[Gradient]\label{lem:gradient}
Under Assumption~\ref{ass:data-loss}, if
$\|a_0(h)\|_{L^4}\le B_4\|h\|_\Hh$ for $a\in \{u,v\}$ and $h\in \Hh$, the risk is finite and
Fr\'echet differentiable with respect to $\Delta\theta\in\Theta_2$. Set
$r(z)=\nabla\loss_y(f_\theta(x))$ and define the backward pass (a.k.a.\ adjoint equations) for $s\in [0,1]$ by
\begin{align}\label{eq:hilbert-backward}
 b(1,z)&={W_{\tout}}r(z),&q(s,z)&=A_v(s)b(s,z),\notag\\
 -\partial_s b(s,z)&=A_u(s)^*[\rho'(p(s,x))q(s,z)].
\end{align}
Then both the forward and backward equations have unique absolutely continuous solutions and the derivative satisfies
$D_{\Delta\theta}\Ll(\theta)[\xi]=\langle\nabla\Ll(\theta),\xi\rangle_2$, where $\nabla\Ll(\theta)\in \Theta_2$ has components
\begin{align}\label{eq:population-gradients}
 \nabla_{W_{\tin}}\Ll&=\int b(0,z)\otimes_\Xx x\,d\mu,
 &\nabla_{W_{\tout}}\Ll&=\int h(1,x)\otimes_\Yy r(z)\,d\mu,\notag\\
 \nabla_u\Ll(s)&=\int\rho'(p(s,x))q(s,z)h(s,x)\,d\mu,
 &\nabla_v\Ll(s)&=\int\rho(p(s,x))b(s,z)\,d\mu.
\end{align}
\end{lemma}

Fix one learning rate (LR) $\eta\ge0$. Starting
with zero increments, the gradient descent (GD) dynamics is defined by
\begin{equation}\label{eq:exact-population}
 \Delta\theta_0=0,\qquad\Delta\theta_{k+1}=\Delta\theta_k-\eta\nabla\Ll(\theta_0,\Delta\theta_k).
\end{equation}
We write $\theta^{(k)}=(\theta_0,\Delta\theta_k)$ when the full parameter at training step $k$ is needed. Note that our analysis can handle different LR for the embeddings and across layer index $s\in [0,1]$, which are important degrees of freedom in practice, but we consider a single LR $\eta$ for notational simplicty.

\paragraph{Subspace stability} In many cases the Gaussian Mean ODE remains in a finite dimensional subspace of $\Hh$, and therefore can be realized exactly up to isometry at finite $D$.  Indeed, writing $n \leq +\infty$ for the number of data, Lemma~\ref{lem:finite-horizon} (Appendix~\ref{app:exact-horizon}) shows that projecting on the subspace
 \begin{equation}\label{eq:exact-rank}
 S=\operatorname{span}\{{W_{\tin,0}}(x_i):i\le n\}
   +\operatorname{span}\{{W_{\tout,0}}r_k(x_i,y_i):i\le n,\ k<K\}
   \end{equation}
   does not change the dynamics. This subspace has dimension bounded by
   $$
   \dim S \leq \min\{ n, d_\tin\} + \min \{ Kn, d_\tout\}.
   $$
   if the input embedding is linear, and $   \dim S \leq n + \min \{ Kn, d_\tout\}$ otherwise.

\paragraph{Interpretation as a dynamics over measures.} As in the shallow NN case~\cite{chizat2018global,mei2018mean,
rotskoff2018parameters,sirignano2020mean}, the limit dynamics can be written as a dynamics in the space of measures. However, in the case where $\Hh$ is infinite-dimensional, there is a slight subtlety that, although the initialization law does not evolve, one must keep track of its dependency with the increments in order to close the equations.
Fix an orthonormal basis $(e_j)$ of $\Hh$ and represent each Gaussian map by
$\xi_a=(a_0(e_j))_{j\ge1} \in \RR^\NN$ for $a\in\{u,v\}$. At depth $s\in [0,1]$, define the joint probability law
$\nu_{k,s}=\operatorname{Law}(\xi_u,\xi_v,\Delta u_k(s),\Delta v_k(s))$ on $(\RR^\NN)^2\times \Hh^2$.
Then $\nu_{k+1,s}=(T_{k,s})_\#\nu_{k,s}$, where
$T_{k,s}(\xi_u,\xi_v,\Delta u,\Delta v)
=(\xi_u,\xi_v,\Delta u-\eta\nabla_u\Ll,\Delta v-\eta\nabla_v\Ll)$
with the gradient given in~\eqref{eq:population-gradients}.

\subsection{Recovering the DMFT equations}\label{sec:dmft}

Several prior works have obtained description of large-scale ResNet in terms of DMFT equations
\cite{bordelon2024depthwise,chaintron2026resnets} which involve the history of training to compute each next iteration. This might appear a priori incompatible with our simple description as GD on the Gaussian Mean ODE. In this section, we explain how to recover the DMFT formulation from our gradient-descent description using Gaussian integration by parts.

\paragraph{Forward projection after $k$ GD steps.} Take Gaussian initial weights $a_0=G_a$ for $a\in \{u,v\}$, fix a depth $s$, omitted in the notation, and write $h_k(z)=h_k(x)$ for a sample $z=(x,y)$. With the forward/backward state histories $(h_t,b_t)_{t\le k}$ fixed, define $p_k[g_u,g_v]$ recursively by
\begin{align}\label{eq:dmft-forward-projections}
 p_k[g_u,g_v](z)={}&g_u(h_k(z))
 -\eta\sum_{t<k}\int\langle h_t(z'),h_k(z)\rangle
\rho'(p_t[g_u,g_v](z'))q_t[g_u,g_v](z')\,d\mu(z').
\end{align}
This function encodes how the projections/pre-activations after $k$ GD steps depend on the initial weights, so that $p_k=p_k[G_u,G_v]$. Here  $q_t[g_u,g_v]$ is similarly defined recursively to represent the projections in the backward pass. We note that, except at $k=0$, these projections are in general non-Gaussian.

\paragraph{Forward pass with response terms.}
For $w\in\Hh$, differentiate the functional in
\eqref{eq:dmft-forward-projections} while holding the states fixed (this is the Malliavin derivative in stochastic calculus):
\[
 \langle D_vp_k(z),w\rangle_\Hh
 =\left.\frac{d}{d\varepsilon}\right|_{\varepsilon=0}
 p_k[G_u,G_v+\varepsilon\langle w,\cdot\rangle_\Hh](z).
\]
For the map
\begin{equation}\label{eq:dmft-phi}
 \Phi^h_{k,z}(G_u,G_v)=\rho\bigl(p_k[G_u,G_v](z)\bigr),
\end{equation}
Gaussian integration by parts~\cite[Theorem 4.15]{nourdin2012lectures}
gives
\begin{equation}\label{eq:gaussian-ibp}
 G_v^*\Phi=\sigma_v^2\EE[D_v\Phi],
 \qquad D_v\Phi^h_{k,z}=\rho'(p_k(z))D_vp_k(z).
\end{equation}
Substituting this into the forward-pass equation~\eqref{eq:hilbert-forward}, and replacing $\Delta v_k$ by the sum of its updates, we get
\begin{equation}\label{eq:dmft-equations}
 \partial_s h_k(z)=\sigma_v^2\EE[\rho'(p_k(z))D_vp_k(z)]
 -\eta\sum_{t<k}\int b_t(z')
       \EE[\rho(p_t(z'))\rho(p_k(z))] \,d\mu(z').
\end{equation}
Combined with~\eqref{eq:dmft-forward-projections} and with the analogous equations for the backward pass $(b_k)$ (omitted here for concision) this yields a closed system of DMFT equations.
While the representation~\eqref{eq:dmft-equations} is where one naturally arrives at when using prior proof techniques (such as the cavity method as in~\cite{chaintron2026resnets} or the path integral method as in~\cite{bordelon2024transformer}), it is not clear to us that there is any advantage using it once the Mean ODE form is available.

\subsection{Regularity of the gradient}\label{sec:regularity-gradient}
To illustrate the interest of having a simple description of the limit model, we propose in this section a preliminary analysis of the regularity of its gradient. Such estimates are of interest from an optimization viewpoint, as they guarantee a descent lemma for GD, the starting point of many optimization analyses. The results in the section are not used in the rest of the paper in this precise form, although the proof of Theorem~\ref{thm:main} involves several related stability estimates, see Proposition~\ref{prop:output-stability}).

\begin{proposition}[Gradient is Quasi-Lipschitz locally]\label{prop:gradient-regularity}\label{lem:gradient-regularity}
Suppose two parameters $\theta,\widetilde\theta$ satisfy
Assumption~\ref{ass:data-loss}, share the same initial internal weights (or maps) $u_0,v_0$,
and, for some $A\ge1$, satisfy the subgaussian moment bounds
\begin{equation}\label{eq:parameter-subgaussian}
 \max_{a\in\{u,v\}}\|a_0\|_{\mathcal B(\Hh,L^p(\Omega))}
 +\|\Delta\theta\|_{\Theta_p}+\|\Delta\widetilde\theta\|_{\Theta_p}
 \le A\sqrt p,\qquad p\ge2.
\end{equation}
Let $\omega(0)=0$ and $\omega(t)=t\sqrt{\log(e/t)}$ for $0<t\le1$.
Then, whenever $\|\theta-\widetilde\theta\|_2\le1$,
\begin{equation}\label{eq:feedback-gradient-moduli}
 \|\nabla\Ll(\theta)-\nabla\Ll(\widetilde\theta)\|_{\Theta_2}
 \le C_{B,A}\omega\!\left(\|\theta-\widetilde\theta\|_2\right).
\end{equation}
\end{proposition}
Gaussian initialization satisfies
\eqref{eq:parameter-subgaussian}, and Lemma~\ref{lem:moment-propagation}
shows that the bound remains satisfied through every fixed
number of GD steps from zero. The proof is in
Appendix~\ref{app:gradient-regularity}.

We can make the following remarks:
\begin{itemize}
\item \emph{Well-posedness of gradient flow.}
Since the modulus of continuity $\omega$ is quasi-Lipschitz and satisfies the Osgood condition
$\int_0^1\omega(r)^{-1}\,dr=\infty$, Proposition~\ref{lem:gradient-regularity} (combined with suitable propagation of moments bounds) implies that there exists a unique gradient flow for the loss, at least locally in time.

\item \emph{Towards Lipschitz regularity.} For mean-field ReLU shallow networks with balanced initialization, i.e.\ $\Vert u_0\Vert_2=\Vert v_0\Vert_2$ in finite dimension (this holds automatically in infinite dimension), it can be shown that the objective has a locally Lipschitz gradient (rather than merely quasi-Lipschitz), as long as the initialization has bounded second moments. This follows from the reduction to measures on a product of spheres in~\cite{chizat2020implicit,chizat2026kmd}. This result exploits homogeneity and the regularity of the input data. It is likely that a similar improvement from quasi-Lipschitz to Lipschitz could be obtained here by exploiting a similar structure, suggesting that such a balanced initialization is beneficial for regularity.

\item \emph{A Lipschitz preconditioned direction.} It is also possible to precondition the gradient to make it locally Lipschitz. For a given neuron, write $Q(s)=(\int|q(s,z)|^2\,d\mu(z))^{1/2}$ and define $g^{\mathrm{cond}}(\theta)$ by keeping all components of $\nabla\Ll(\theta)$ except its $u$ component, replacing it by
\begin{equation}\label{eq:adaptive-feedback-update}
 g_u^{\mathrm{cond}}(\theta)(s)
 =\frac{\nabla_u\Ll(\theta)(s)}{\sqrt{1+Q(s)^2}}.
\end{equation}
Under the assumptions of Proposition~\ref{prop:gradient-regularity}, it can be shown that
\begin{equation}\label{eq:normalized-lipschitz}
 \|g^{\mathrm{cond}}(\theta)-g^{\mathrm{cond}}(\widetilde\theta)\|_{\Theta_2}
 \le C_{B,A}\|\theta-\widetilde\theta\|_2.
\end{equation}
This preconditioner ensures that the size of the updates of $u$ is controlled.
\end{itemize}

\section{Main theorem}\label{sec:applications}

\subsection{Initialization of embeddings}
\label{sec:main-theorem}

\paragraph{Limit embeddings and their covariance.}
Define the uncentered input covariance $C_\tin$ and choose a output embedding covariance $Q$:
\[
 C_\tin=\int x\otimes x\,d\mu,\qquad Q:\Yy\to\Yy,\qquad
 Q\ge0,\quad\operatorname{tr}Q\le B^2.
\]
The operator $Q$ is part a property of the initialization of the output embedding: it is required to ensure bounded HS norm for $W_{\tout,0}$, since bounded operator norm is not enough to get estimates that are uniform in $d_\tout$.\footnote{This requirement is specific to our Hilbertian setting (it is in particular not needed with one-hot encoding).}
For the reference limit model, we take
\begin{equation}\label{eq:reference-boundaries}
 \Hh=\Xx\oplus\Yy,\qquad {W_{\tin,0}}x=(x,0),\qquad {W_{\tout,0}}y=(0,Q^{1/2}y).
\end{equation}
The initial output embedding satisfies
$\|{W_{\tout,0}}\|_{\mathrm{op}}\le\|{W_{\tout,0}}\|_{\HS}=(\operatorname{tr}Q)^{1/2}\le B$.
The initial prediction is zero, and the initialization of the backward pass is
$b_0(1,z)=(0,Q^{1/2}\nabla\loss_y(0))$.

The covariance governing approximation of these initial embeddings is
\begin{equation}\label{eq:task-covariance}
 C=\int {W_{\tin,0}}x\otimes {W_{\tin,0}}x\,d\mu+{W_{\tout,0}}{W_{\tout,0}}^*=C_\tin\oplus Q.
\end{equation}
It is positive and trace class, with trace at most $2B^2$.
Write $(\lambda_j)_{j\ge1}$ for its eigenvalues in decreasing order,
counted with multiplicity and completed by zeros when necessary. Although the embedding evolve under GD, our result only depend on the spectrum $(\lambda_j)_j$ of the initial embeddings.

\paragraph{Embeddings in finite dimension $D$.} In our analysis, we consider two possible choices for the initial embeddings of the finite ResNet. They both approximate the same limit embeddings
$W_{\mathrm{in},0}x=(x,0)$ and $W_{\mathrm{out},0}y=(0,Q^{1/2}y)$):
\begin{itemize}
\item \textbf{Random Gaussian embeddings.} This is a standard random embedding with Gaussian iid weights, followed by a norm clipping. For an embedding dimension $D$ and finite-dimensional input/output spaces, let $\Hh_D=(\RR^D,\langle\cdot,\cdot\rangle_2)$. Draw independent matrices $G_{\mathrm{in}}\in\mathbb R^{D\times d_{\mathrm{in}}}$ and $G_{\mathrm{out}}\in\mathbb R^{D\times d_{\mathrm{out}}}$ with independent standard Gaussian entries on a probability space $\Omega_{\mathrm{emb}}$ (hence independent of the internal weights) and set,
\begin{align}\label{eq:init_gaussian_embeddings}
W_{\mathrm{in},0,D}(x) =\Pi_{2B}^{\Hh_D}\Big(\frac1{\sqrt D}G_{\mathrm{in}}x\Big),
\qquad
W_{\mathrm{out},0,D}=\Pi_{2B}^{\mathrm{HS}}\Big(\frac1{\sqrt D} G_{\mathrm{out}}Q^{1/2}\Big).
\end{align}

where $\Pi_B^S$ denotes the projection on the ball of radius $B$ in $S$ and is introduced to ensure that Assumption~\ref{ass:data-loss} is met with probability $1$ (rather than just with high probability). These projections only apply to the initial embeddings.

The same initialization can be defined for infinite dimensional input/output spaces as follows. Let $g_1,\ldots,g_D$ be independent unit-variance isonormal Gaussian maps
$g_1,\ldots,g_D$ defined on a probability space $\Omega_{\rm emb}$, and set
\[
 S_Dh=\frac1{\sqrt D}(g_1(h),\ldots,g_D(h)),\qquad
 \EE_{\rm emb}[g_i(h)g_j(h')]
   =\mathbf1_{\{i=j\}}\langle h,h'\rangle_\Hh.
\]
and
\begin{equation}\label{eq:bounded-random-map}
 {W_{\tin,0,D}}(x)=\Pi_{2B}^{\Hh_D}(S_D{W_{\tin,0}}x),\qquad
 {W_{\tout,0,D}}=\Pi_{2B}^{\mathrm{HS}}\bigl(S_D{W_{\tout,0}}\bigr),
\end{equation}
We recall that we do not need linearity of the initial input embedding $W_{\tin,0,D}$,
only the learned increment $\Delta\widehat W_{\tin,k,D}$ is required to be linear.
\item \textbf{Principal-components embeddings.} One can also consider deterministic embeddings that are aligned with the principal directions of $C$. More precisely, let $P_r=P_r^\tin\oplus P_r^\tout$ be the orthogonal projection onto the first $r$ principal directions of $C$, choosing eigenvectors
within the two summands in case of ties. Taking the retained subspace $P_D\mathcal H$ as the hidden space, initialize
\begin{align}\label{eq:emb-principal}
W_{\mathrm{in},0,D}x=P_D(x,0),&&
W_{\mathrm{out},0,D}
=P_DW_{\mathrm{out},0}.
\end{align}
 When $Q=0$, this is simply PCA of the inputs fed as input embeddings and a zero output embedding. The projection is used only to build the initialization; the gradient updates of the embedding are not explicitly projected.
\end{itemize}

\subsection{Main theorem}\label{sec:main-result}
Consider the ResNet forward pass~\eqref{eq:resnet}.  Write
$\widehat\theta=(\widehat\theta_0,\Delta\widehat\theta)$ and let the increments follow GD and the loss~\eqref{eq:loss-resnet}
\begin{equation}\label{eq:resnet-gd}
 \Delta\widehat\theta_{0,D}=0,\qquad
 \Delta\widehat\theta_{k+1,D}=\Delta\widehat\theta_{k,D}
 -\eta\nabla\widehat\Ll(\widehat\theta_0,\Delta\widehat\theta_{k,D}),
\end{equation}
where the gradient is defined with respect to the Hilbert--Schmidt geometry for the
embeddings and the squared norm
$(LM)^{-1}\sum_{\ell,j}(\|\Delta\widehat u^{j,\ell}\|_{2}^2+
\|\Delta\widehat v^{j,\ell}\|_{2}^2)$ for the internal weights (the explicit
updates are given in Appendix~\ref{app:finite-comparison}).

For the initialization of internal weights, we take
$\widehat a_0^{j,\ell}=\sigma_a\xi^{a,j,\ell}$ in  $\RR^D$, for $a\in\{u,v\}$ where $\xi^{a,j,\ell}$ are
independent, centered, have variance one, and have sub-Gaussian norm at
most $B$ (each neuron and layer uses a copy of the same product law).
They are independent of the initial random embeddings given by~\eqref{eq:bounded-random-map} .

 In the case of Gaussian initialization,
this recovers the initialization~\eqref{eq:gaussian-init}. For non-Gaussian initial internal weights, assume additionally
$\rho\in C^4$ with bounded derivatives of orders one through four (these derivative bounds are included in $B$).

Our main theorem is the following.

\begin{theorem}[Quantitative convergence to Gaussian Mean ODE]
\label{thm:main}\label{thm:nongaussian}\label{cor:main-finite}
Under Assumption~\ref{ass:data-loss} and the initialization conditions above, let $\hat f_{k,D}$ and $f_k$ be the output after $k$ GD steps on the ResNet~\eqref{eq:resnet} with random embeddings~\eqref{eq:bounded-random-map} and the Gaussian Mean ODE with embeddings~\eqref{eq:reference-boundaries} respectively, with the same learning rate $\eta$.

For $0<\delta<1/2$ and $r\in [1:D]$ let
\begin{equation}\label{eq:finite-log}
 t=\log\frac{C_{B,K}(K+1)L^2M}{\delta}.
\end{equation}
If $L\ge C_{B,K}$, $LM\ge C_{B,K}(D+t)$ and
$D\ge C(r+\log(C/\delta))$, then with probability at least $1-\delta$,
\begin{equation}\label{eq:main-finite-rate}
 \begin{aligned}
 \max_{k\le K}\|\widehat f_{k,D}-f_k\|_{L^2(\mu;\Yy)}
 &\le C_{B,K}\bigg[
 \frac1L+t^{(K+2)/2}\sqrt{\frac D{LM}}
 +\left(\sum_{j>r}\lambda_j\right)^{1/2}+\sqrt{\frac{r+\log(CD/\delta)}D} \bigg].
 \end{aligned}
\end{equation}

The constants are independent of $L,M,D$, the input and output dimensions,
and the size of the dataset. One can minimize the bound over $r$, which is a proof parameter.
\end{theorem}

We can make the following remarks:
\begin{itemize}
\item For a fixed $K$, the condition for the error to vanish is $L,D\to \infty$ and, up to a polylog factor, $D/(LM)\to 0$. There is no separate lower bound on $M$, so $M=1$ is allowed. This is a fixed-horizon statement, the constants may grow exponentially or faster with $K$. The last term contains both the random embedding error and the Gaussianization error when the initial internal weights are non-Gaussian (see next Section for a breakdown of these errors).
\item For the principal component embeddings in Section~\ref{sec:main-theorem}, with Gaussian
internal initialization, replace the last two terms in
\eqref{eq:main-finite-rate} by
$\big(\sum_{j>D}\lambda_j\big)^{1/2}$, see Proposition~\ref{prop:embedding-error}. For Gaussian initialization, the last term in \eqref{eq:main-finite-rate} can be replaced by the sharper bound~\eqref{eq:main-random-population} given below.
\end{itemize}

\subsection{Proof ideas}\label{sec:proof-idea}
As previously discussed, the argument to obtain Theorem~\ref{thm:main} follows two main stages: the large $L$ limit and the large $D$ limit. In case of non-Gaussian initial internal weights, there is another third stage to control the Gaussianization error. We now detail these three stages separately.

\subsubsection{From ResNets to Mean ODEs}\label{sec:finite-limit}
For the first stage, we use the following result which refines the analysis in~\cite{chizat2025hidden}.
\begin{proposition}[ResNet to Mean ODE]
\label{thm:finite-mean-ode}\label{thm:finite-network}\label{prop:finite-mean-ode}
Assume Assumption~\ref{ass:data-loss} and independent centered
sub-Gaussian internal initial weights as above. Let the finite ResNet and
the Mean ODE in $\Hh_D$ have the same initial embeddings and same internal
initialization law, and follow $K$ steps with the same learning rate $\eta$.
With $t$ as in~\eqref{eq:finite-log}, if $L\ge C_{B,K}$ and
$LM\ge C_{B,K}(D+t)$, then with probability at least $1-\delta$,
\begin{equation}\label{eq:finite-mean-ode-rate}
 \max_{k\le K}\|\widehat f_{k,D}-f_{k,D}\|_{L^2(\mu;\Yy)}
 \le C_{B,K}\left[\frac1L+t^{(K+2)/2}\sqrt{\frac D{LM}}\right].
\end{equation}
The same right-hand side bounds the input and output
embedding errors in Hilbert--Schmidt norm and the forward and backward
state errors in $L^2(\mu)$, maximized over the depth.
\end{proposition}

The first term is the ODE discretization error and the second term is the error of sampling a total of $LM$ neurons rather than an infinite number of them throughout the residual part of the architecture. The argument, based on a stochastic approximation viewpoint, does not require each layer to converge to its mean-field limit, but instead relies on averaging through depth.

The key improvement compared to~\cite{chizat2025hidden} is the removal of the condition $M\gtrsim D$. Without this condition, say with $M=1$, the operator norm of each layer grows as $\sqrt{D}$, which makes it challenging to control how errors propagate through training\footnote{Indeed, in the regime $M\ll D$, the proper initialization scaling makes the operator norm of individual layers diverge. In particular, the ``spectral criterion'' for feature learning proposed in~\cite{yang2023spectral} is not a good heuristic in this context.}. This difficulty is bypassed by grouping adjacent layers (to restaure a uniform operator norm bound) and a careful analysis of the dependencies to track the errors within groups. The proof is in Appendix~\ref{app:finite-comparison}.

Numerical experiments in~\cite{chizat2025hidden} support the sharpness of the bound for hidden states. For the error on the outputs, one expects that the leading order fluctuation, which is centered, gets averaged out by going through the output embedding and looses a factor $\sqrt{D}$. In this argument, the higher-order fluctuations (due to self-interaction of the error terms) remain. This leads to the following conjecture.

\begin{conjecture}[Prediction error at finite $D$]\label{conj:prediction}
Under the assumptions of Proposition~\ref{prop:finite-mean-ode}, which include
$\|W_{\tout,0,D}\|_{\HS}\le B$, one expects, up to logarithmic factors,
\begin{equation}\label{eq:conjectured-output}
 \max_{k\le K}\|\widehat f_{k,D}-f_{k,D}\|_{L^2(\mu;\Yy)}
 \lesssim_{B,K}\frac1L+\frac1{\sqrt{LM}}+\frac D{LM}.
\end{equation}
\end{conjecture}
%Numerical experiments illustrating this rate can be found in~\cite{chaintron2026resnets}.
We note that it is possible to build simple examples where these terms are present, ruling out the possibility of replacing the factor $D$ by a smaller quantity.

\subsubsection{The embedding error}\label{sec:embedding-error}

With Gaussian initial internal weights, the large-dimensional limit follows from stability of the Mean ODE under perturbations of the initial embeddings, after comparison up to isometries.

\begin{proposition}[Approximation of the initial embeddings]
\label{prop:embedding-error}\label{prop:output-stability}
For two Mean ODEs satisfying Assumption~\ref{ass:data-loss}, with the same
data law, loss and activation, common internal initializations
$u_0,v_0$ having moments of every finite order and zero initial
increments $\Delta \theta=\Delta \tilde\theta=0$, GD with the same learning rate satisfies
\begin{equation}\label{eq:population-output-stability}
 \max_{k\le K}\|f_k-\widetilde f_k\|_{L^2(\mu)}
 \le C_{B,K}\|\theta_0-\widetilde\theta_0\|_2.
\end{equation}
Here the distance is~\eqref{eq:full-parameter-distance} with
$\Delta \theta=\Delta \tilde\theta=0$. Consequently, for Gaussian Mean ODEs with the embeddings of Section~\ref{sec:main-theorem} it holds:
\begin{itemize}
\item  For principal component embeddings~\eqref{eq:emb-principal} ,
\begin{equation}\label{eq:main-principal-population}
 \max_{k\le K}\|f_{k,D}-f_k\|_{L^2(\mu;\Yy)}
 \le C_{B,K}\left(\sum_{j>D}\lambda_j\right)^{1/2}.
\end{equation}
\item 
For random embeddings~\eqref{eq:bounded-random-map}, every $D\ge1$ and $0<\delta<1/2$, with probability at least $1-\delta$,
\begin{equation}\label{eq:main-random-population}
 \max_{k\le K}\|f_{k,D}-f_k\|_{L^2(\mu;\Yy)}
 \le C_{B,K}\left[
 \frac1D\sum_{j\le D}j\lambda_j+\sum_{j>D}\lambda_j
 +\frac{\lambda_1\log(1/\delta)}D\right]^{1/2}.
\end{equation}
\end{itemize}
These two bounds also control the hidden states error in $L^2(\mu)$, uniformly over depth and $k\le K$, up to a fixed isometry.
\end{proposition}

The simpler rank-$r$ estimate used in Theorem~\ref{thm:main} follows directly from~\eqref{eq:main-random-population}, since for every $1\le r\le D$,
\begin{equation}\label{eq:weighted-to-rank-bound}
 \frac1D\sum_{j\le D}j\lambda_j+\sum_{j>D}\lambda_j
 \le \frac rD\operatorname{tr}C+\sum_{j>r}\lambda_j.
\end{equation}
Using $\lambda_1\le\operatorname{tr}C\le2B^2$ in~\eqref{eq:main-random-population} therefore gives, on the same event,
\begin{equation}\label{eq:main-random-population-rank}
 \max_{k\le K}\|f_{k,D}-f_k\|_{L^2(\mu;\Yy)}
 \le C_{B,K}\left[\left(\sum_{j>r}\lambda_j\right)^{1/2}
 +\sqrt{\frac{r+\log(1/\delta)}D}\right].
\end{equation}

The first estimate~\eqref{eq:population-output-stability} follows by the stability of the gradient. Although in Proposition~\ref{prop:gradient-regularity} the gradient is not exactly Lipschitz, one can recover a Lipschitz-like estimate by using moments of increasing order at each GD step.  Gaussian invariance then permits comparison up to isometries, giving the principal-components bound~\eqref{eq:main-principal-population}. 

%For random embeddings, one needs an additional (standard) probabilistic argument to control the error it induces on the show that it approximates the identity on a $r$ dimensional subspace. These arguments are in Appendices~\ref{app:moment-stability} and~\ref{app:geometry-main}.

 For random embeddings, applying Gram-Schmidt orthogonalization to the Gaussian vectors in decreasing covariance order yields an expected squared error $O(j/D)$ for the $j$th direction, for $j\le D$, and $O(1)$ for the remaining directions. Summing with weights $\lambda_j$ gives the spectral term in~\eqref{eq:main-random-population}, and a concentration argument gives the tail bound. The proofs are in Appendices~\ref{app:moment-stability} and~\ref{app:geometry-main}.

\subsubsection{Gaussianization error}\label{sec:non-gaussian}
We now consider ResNets with non-Gaussian internal weights.
Since Proposition~\ref{prop:finite-mean-ode} holds in the non-Gaussian case as well, what remains to prove is that the high-dimensional limit of a non-Gaussian Mean ODE is again the Gaussian Mean ODE. The next proposition shows that this holds under a delocalization condition, which is in particular satisfied by random embeddings (see below). This condition requires that the directions in $\RR^D$ that carry the hidden states are sufficiently spread across coordinates, allowing a central-limit theorem-like behavior that ``Gaussianizes'' the distribution.

\begin{proposition}[Gaussianization on a spread initial subspace]
\label{prop:gaussianization}
Under Assumption~\ref{ass:data-loss}, let $f_k^{\mathrm{G}}$ and $f_k^{\mathrm{NG}}$ be respectively the Gaussian and non-Gaussian
Mean ODE's output after $k$ GD steps, with common initial
embeddings in $\Hh_D$, and under the subgaussianity assumption stated
before Theorem~\ref{thm:main}. Assume $\rho\in C^4$ with bounded derivatives of orders one through four, and furthermore $\rho\in C^5$ with bounded fifth derivative in order to benefit from the improved rate when the initial internal distribution has zero third moments (these derivative bounds are included in $B$).

Suppose that the initial embeddings and
$\operatorname{ran}W_{\tout,0}$ lie in a rank-$r$ subspace represented by
$A\in\RR^{D\times r}$ with $A^*A=I_r$. Write $a_i^*$ for its rows and
$\alpha=\max_i\|a_i\|_2$. Then
\begin{equation}\label{eq:rank-gaussianization}
 \max_{k\le K}\|f_k^{\mathrm{NG}}-f_k^{\mathrm G}\|_{L^2(\mu;\Yy)}
 \le C_{B,K}\begin{cases}
 \alpha,&\text{in general},\\
 \alpha^2,&\text{if zero third moments}.
 \end{cases}
\end{equation}
The same bound controls the forward and backward states in
$L^2(\mu;C([0,1];\Hh_D))$, including the components perpendicular
to $\operatorname{ran}A$.
\end{proposition}

The general proof structure follows the idea of \emph{skeleton maps} in~\cite{chaintron2026resnets}. These maps express the weights updates or related quantities as functions of the initial internal randomness, while freezing forward and backward states to the value they take on the Gaussian dynamics. Evaluating these maps with non-Gaussian inputs builds an approximate trajectory that lies between the Gaussian and the non-Gaussian ones. A Lindeberg replacement argument then controls the errors induced on the averaged forward and backward velocity fields. One then propagate these errors through GD using stability.

Finally, to prove Theorem~\ref{thm:main}, we retain only the first $r$ principal directions in the Gaussianization result. With random embeddings, we then have $\alpha\le C\sqrt{(r+\log(CD/\delta))/D}$ with high probability. Combining the three propositions, and allocating a fixed fraction of $\delta$ to each probabilistic estimate, gives the main bound. The Gaussian replacement argument
and the final assembly of all the estimates can be found in Appendix~\ref{app:gaussianization}.

\subsection{Numerical verification of scaling laws}\label{sec:scaling}

We study how the shape of a ResNet affects its approximation of the large-scale limit on a task with an approximately power-law covariance spectrum. The training horizon is fixed, and logarithmic factors are omitted from the scaling predictions.

\paragraph{Experimental setting.}
We use CIFAR-10 classification with flattened images, ten classes, cross-entropy loss and $\tanh$ activation. ResNets have Gaussian embeddings and internal initialization, and are trained for ten full-batch GD updates with LR $\eta=0.9$, with five initialization seeds per architecture. For the limit dynamics we use as a proxy a much larger ResNet. We measure discrepancies in $L^2(\mu)$ for outputs and last hidden states; for the latter, the initial embeddings are aligned by an isometry (PCA alignment), which is kept fixed during training. Results are averaged over random seeds.

\begin{figure}[!htbp]
 \centering
 \begin{subfigure}{0.49\linewidth}
  \centering
  \includegraphics[width=\linewidth]{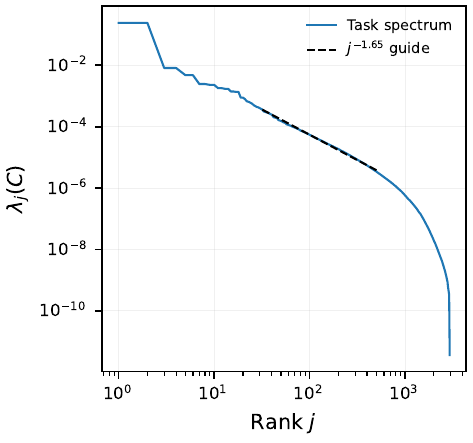}
  \caption{Task spectrum}
  \label{fig:cifar-spectrum}
 \end{subfigure}\hfill
 \begin{subfigure}{0.49\linewidth}
  \centering
  \includegraphics[width=\linewidth]{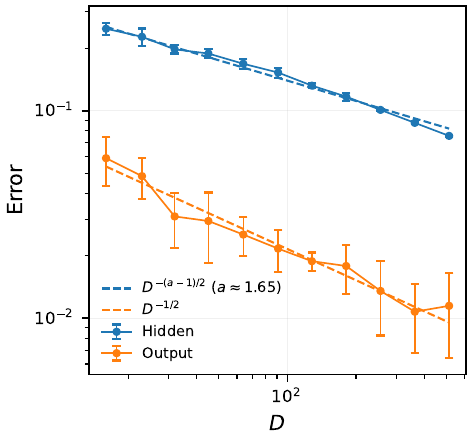}
  \caption{Dimension dependence}
  \label{fig:scaling-polar-dimension}
 \end{subfigure}
 \caption{CIFAR-10 spectrum and approximation errors. Left: eigenvalues of the task covariance $C$ in~\eqref{eq:task-covariance}). Right: hidden-state and output errors, with guides $D^{-(a-1)/2}$ at $a=1.65$ and $D^{-1/2}$; error bars show variation across initialization random seeds.}
 \label{fig:cifar-dimension}
\end{figure}

Figure~\ref{fig:cifar-spectrum} shows the task spectrum, which displays a polynomial scaling $\lambda_j\propto j^{-a}$ with $a=1.65$ (we used an output embedding covariance $Q$ with the same spectral decay so as to obtain a clean scaling law).

\paragraph{Dimension dependence.}
For $\lambda_j\lesssim j^{-a}$, Proposition~\ref{prop:embedding-error} bounds the embedding contribution by
\begin{equation}\label{eq:scaling-refined-alignment}
 \left(\frac1D\sum_{j\le D}j\lambda_j+\sum_{j>D}\lambda_j\right)^{1/2}
 \lesssim
 \begin{cases}
 D^{-(a-1)/2},&1<a<2,\\
 \sqrt{\log(1+D)/D},&a=2,\\
 D^{-1/2},&a>2.
 \end{cases}
\end{equation}
At $a=1.65$, the hidden-state guide is therefore $D^{-0.325}$. The measured slopes in Figure~\ref{fig:scaling-polar-dimension} are $-0.341$ for hidden states and $-0.473$ for outputs. The faster decay, close to $D^{-1/2}$, for the output error   can be explained as follows: over a fixed training horizon, the output only effectively depends on a bounded number of directions as it take time to learn the signal carried by weak directions (this phenomenon can be quantified in linear regression for instance). Therefore the approximation error for the learnt predictor is insensitive to the discarded covariance tail, over such horizons. We hence expect a rate $D^{-1/2}$ in general for the error on the outputs. We use this conjectural rate for the output scaling law below.

\paragraph{Scaling law in hidden space.}
For Gaussian internal initialization, Propositions~\ref{prop:finite-mean-ode} and~\ref{prop:embedding-error} give for the error in hidden space
\begin{equation}\label{eq:scaling-principal}
 E_h\lesssim\frac1L+\sqrt{\frac D{LM}}
 +\left(\frac1D\sum_{j\le D}j\lambda_j+\sum_{j>D}\lambda_j\right)^{1/2}.
\end{equation}
Fix the non-embedding budget $\mathcal C=LMD$. Balancing $\sqrt{D/(LM)}=D/\sqrt{\mathcal C}$ with the embedding term gives, for $1<a<2$,
\begin{equation}\label{eq:scaling-hidden-budget}
 D\asymp\mathcal C^{1/(a+1)},
 \qquad E_h\lesssim\mathcal C^{-(a-1)/(2(a+1))},
 \qquad L\gtrsim\mathcal C^{(a-1)/(2(a+1))}.
\end{equation}
Taking equality in the depth condition gives $M\asymp\mathcal C^{1/2}$. For $a\ge2$, the dimension and error exponents saturate at $1/3$ and $-1/6$, up to logarithmic factors when $a=2$.

\paragraph{Scaling law in output space.}
Combining the averaging effect in Conjecture~\ref{conj:prediction} with the conjectural $D^{-1/2}$ embedding error  discussed above gives
\begin{equation}\label{eq:scaling-numerical-guides}
 E_f\lesssim\frac1L+\frac1{\sqrt{LM}}+\frac D{LM}+\frac1{\sqrt D}
 =\frac1L+\sqrt{\frac D{\mathcal C}}+\frac{D^2}{\mathcal C}+\frac1{\sqrt D}.
\end{equation}
Balancing $D^2/\mathcal C$ and $D^{-1/2}$ yields
\begin{equation}\label{eq:scaling-output-budget}
 D\asymp\mathcal C^{2/5},
 \qquad E_f\lesssim\mathcal C^{-1/5},
 \qquad L\gtrsim\mathcal C^{1/5}.
\end{equation}
The remaining term is $\sqrt{D/\mathcal C}=\mathcal C^{-3/10}$. Thus $L\asymp\mathcal C^{1/5}$ and $M\asymp\mathcal C^{2/5}$ achieve the predicted rate, independently of $a$.

\begin{figure}[!htbp]
 \centering
 \includegraphics[width=\linewidth]{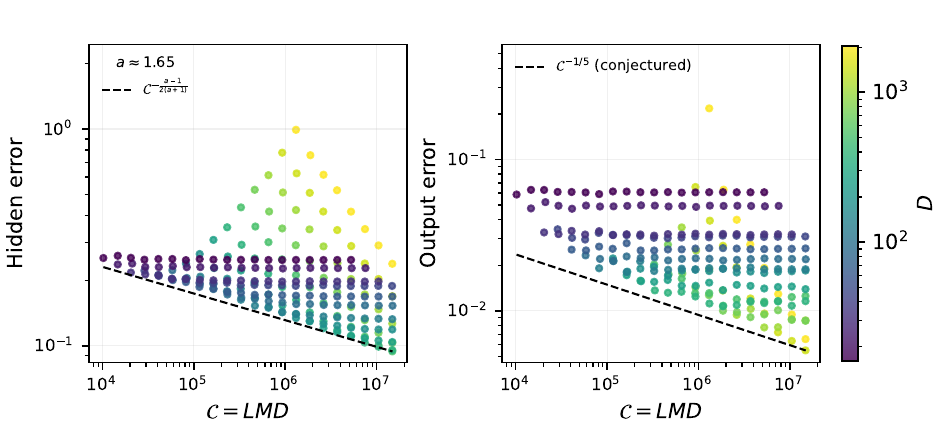}
 \caption{CIFAR-10 approximation errors versus non-embedding parameters budget $\mathcal C=LMD$: hidden states (left) and outputs (right). Each dot represents one architecture, colored by $D$. The dashed guides are $\mathcal C^{-(a-1)/(2(a+1))}$ with $a=1.65$ and the conjectured $\mathcal C^{-1/5}$.}
 \label{fig:scaling-polar-budget}
\end{figure}

Figure~\ref{fig:scaling-polar-budget} compares 219 architectures with non-embedding parameter budget $\mathcal C\le2\times10^7$ and various shapes. The predicted (proved) error in hidden space has exponent $-(a-1)/(2(a+1))\simeq-0.122$ at $a=1.65$, while the predicted (conjectured) error in output space has exponent $-1/5$. Empirical fits on these plots give slopes $-0.133$ and $-0.273$, respectively: this closely follows the prediction for the hidden error and is slightly conservative (yet well within fluctuation margins) for the output error.

\section{Mixture of Experts}\label{sec:experts}
In order to illustrate the flexibility of the Gaussian Mean ODE viewpoint, we now formulate a candidate limit model for a ResNet with Mixture of Experts layers. Such models were previously studied in the large-scale limit with the DMFT viewpoint in~\cite{jiang2026moe, dhir2026routers}. Here our goal is modest and only to hint at the fact that the limiting mechanisms are similar to those for the plain ResNet, leading to a simple nested Mean ODE limit. For the sake of brevity we limit ourselves to rapid justifications and we fix $d_\tin,d_\tout<+\infty$ and treat them as $\Theta(1)$ parameters.

\paragraph{MoE ResNet architecture} A Mixture of Experts (MoE) layer contains $N$ experts $F^{\alpha,\ell}$ indexed by $\alpha$, each with $M$
neurons, and a gating weight $g^{\alpha,\ell}$:
\[
 F^{\alpha,\ell}(h)=\frac1M\sum_{j=1}^M v^{\alpha j,\ell}
 \rho\big((u^{\alpha j,\ell})^\top h\big),\qquad
 g^{\alpha,\ell}(h)=\gamma\big((w^{\alpha,\ell})^\top h\big).
\]
Here $\rho$ and $\gamma$ are smooth nonlinear activation functions with $\gamma$ valued in $[0,1]$ (this is a simplifying model as generally $\gamma$ also involves a discrete switch in the forward pass). The forward pass of the MoE ResNet we consider is
\[
 \widehat h^0=\widehat W_{\tin}x,\qquad
 \widehat h^\ell=\widehat h^{\ell-1}
 +\frac1{LN}\sum_{\alpha=1}^N
 g^{\alpha,\ell}(\widehat h^{\ell-1})
 F^{\alpha,\ell}(\widehat h^{\ell-1}),\qquad
 \widehat f=\widehat W_{\tout}^*\widehat h^L.
\]

We retain the convention that the hidden states have $\ell_2$ norm of order $1$.
We initialize the neuron weights and routers independently with
covariances $\sigma_a^2I_D$, for $a\in\{u,v,w\}$ and consider GD with LRs $LMN\eta$
for the neurons, $LN\eta$ for the router vectors and $\eta$ for the embeddings.

\paragraph{The large scale limit} We know describe the large-scale limit of this architecture.
We write the limit in its reduced form in $\RR^{r}$, with $r=d_\tin+d_\tout$.

Take independent initial vectors
$a_0\sim\Nn(0,\sigma_a^2I_{r})$ for $a\in\{u,v,w\}$.
Write $\xi=w_0$ for the router's source of randomness and $\omega=(u_0,v_0)$ for a neuron's source of
randomness. The learned weights are $a(s,\xi,\omega)=a_0+\Delta a(s,\xi,\omega)$
for $a\in\{u,v\}$ and $w(s,\xi)=\xi+\Delta w(s,\xi)$.
Replacing the two averages by expectations gives the nested Gaussian Mean ODE
\begin{align}
 F^\xi(s,h)&=\EE_\omega\!\left[v(s,\xi,\omega)
                     \rho\big(u(s,\xi,\omega)^\top h\big)\right],
 &g^\xi(s,h)&=\gamma\big(w(s,\xi)^\top h\big),
 \label{eq:moe-gaussian-expert}\\
 \partial_s h(s,x)&=\EE_\xi\!\left[g^\xi(s,h(s,x))F^\xi(s,h(s,x))\right],
 &h(0,x)&=W_{\tin}x,\quad f_\theta(x)=W_{\tout}^*h(1,x).
 \label{eq:moe-gaussian-forward}
\end{align}
The inner expectation averages over neurons with the router randomness held fixed and the outer expectation averages over experts. The initial embeddings are, as in~\eqref{eq:reference-boundaries}, $W_{\tin,0}x=(x,0)$ and
$W_{\tout,0}y=(0,Q^{1/2}y)$. As in the vanilla Mean ODE, the hidden state is deterministic.  The
gradient is taken in the geometry that averages squared $\ell_2$ norms over the units and
depth and uses HS norms for embeddings. 

\paragraph{Convergence conditions} Our informal claim is that the large-scale limit  of the MoE ResNet training dynamics is described by GD on this limit model. This holds under shape scalings that allow for the suitable concentration mechanisms to take place:
\begin{itemize}
\item We need $L\to \infty$ for the ODE limit to arise and $D\to \infty$ for the embeddings to concentrate to isometric inclusions;
\item We need $D/(LN)\to 0$ for the stochastic approximation argument to obtain concentration of the $LN$ experts to their expectation $\EE_{\xi}[\cdot]$;
\item We need $M\to \infty$ for each expert to converge to its own mean-field limit $\EE_{\omega}[\cdot]$.
\end{itemize}
Notably, no separate condition $D/M\to0$ should be required: although the expert’s vector-valued sampling fluctuations have squared norm of order $D/M$ at initialization, their contribution to the hidden states is averaged across experts and depth. The router update, which acts separately within each expert, instead involves a scalar quantity $\langle b,F^{\alpha,\ell}(h)\rangle_2$, whose sampling fluctuations are of order $M^{-1/2}$. Training the router feeds these fluctuations back into the dynamics, so this remaining error requires $M\to\infty$. 

We refer to~\cite{jiang2026moe, dhir2026routers} for dedicated analysis of the scaling regimes of MoE. Our main aim here is to emphasize the simple expression~\eqref{eq:moe-gaussian-forward} available for the limit model in the regime $D/(LN)\to 0$, in the line of our analysis.

\section{Conclusion}
By exploiting the geometry of the embedding space, our analysis yields a simple description of the large-scale limit, and refined quantitative convergence bounds compared to those of~\cite{chaintron2026resnets}. It would be interesting to see how much of this viewpoint extends to the SDE depth scaling under the same shape condition $D\ll LM$. At the critical scaling $D\asymp LM$, however, the SDE scaling and the MLU regime coincide (see~\cite{chizat2025hidden}), and we do not expect the limiting dynamics to admit a similarly simple gradient descent formulation.

\begingroup\footnotesize

\bibliographystyle{plain}
\bibliography{references}

\endgroup

%\clearpage
\appendix

\section{Mean ODE: auxiliary results}\label{app:population}

We use the parameter norms~\eqref{eq:parameter-norm} and write $L^p$
for $L^p(\Omega)$. Suprema in depth and over the data are essential
suprema. We absorb
$B_2,B_4$ into $B$; a constant indexed by $p$ may also depend on the
finitely many higher moment bounds used in its proof. Constants for
$K$ updates may depend on the fixed $\eta\ge0$, but never on dimensions
or sample size. The standing bound on an initialization internal map is
\begin{equation}\label{eq:seed-moments}
 \|a_0(w)\|_{L^p}\le B_p\|w\|_\Hh,\qquad a\in\{u,v\},
\end{equation}
whenever the order $p$ is used. We may choose $B_p$ nondecreasing.

For the auxiliary estimates that require uniform control in depth, write
\[
 \|\Delta\theta\|_{\Thetainf{p}}^2
 :=\|\Delta W_{\tin}\|_{\HS}^2+\|\Delta W_{\tout}\|_{\HS}^2
 +\operatorname*{ess\,sup}_s\|\Delta u(s)\|_{L^p}^2
 +\operatorname*{ess\,sup}_s\|\Delta v(s)\|_{L^p}^2.
\]
This is stronger than $\Theta_p$, which always means $L^p$ in depth.
The uniform norm is used for the fixed-horizon bounds and discretization;
Section 2.4 and Appendix~\ref{app:gradient-regularity} use the integrated norm.

\subsection{Well-posedness and gradient}
\label{ssec:gradient_expression_proof}
\begin{proof}[Proof of Lemma~\ref{lem:gradient}]
Fix the initialization $\theta_0$ and increments $\Delta\theta\in\Theta_2$.
The argument differentiates only the increments. The distinction from a
uniform-in-depth norm is that the ODE Lipschitz coefficients below are
integrable functions of depth.

\paragraph{Step 1. Existence and state bounds.}
For $F(s,w):=A_v(s)^*\rho(A_u(s)w)$,
\[
 \begin{aligned}\|F(s,w)-F(s,\widetilde w)\|
 &\le B(B+\|\Delta v(s)\|_{L^2})(B+\|\Delta u(s)\|_{L^2})
       \|w-\widetilde w\|,\\
 \|F(s,0)\|&\le B(B+\|\Delta v(s)\|_{L^2}).\end{aligned}
\]
Both coefficients are integrable by Cauchy--Schwarz in depth. Strong
measurability follows as in the usual Carath\'eodory theorem
\cite[Theorem 2.4]{aulbach1996integral}. That theorem and Gronwall give a
unique absolutely continuous forward solution, with
$\sup_{s,x}\|h(s,x)\|\le C_{B,\|\Delta\theta\|_{\Theta_2}}$.
Consequently the prediction and residual are bounded, and
\begin{equation}\label{eq:loss-growth-bound}
 |\loss_y(a)|\le |\loss_y(0)|+\|\nabla\loss_y(0)\|\|a\|+\tfrac B2\|a\|^2
\end{equation}
proves finiteness of the risk. The linearized operator
$J(s,x)w:=A_v(s)^*[\rho'(p(s,x))A_u(s)w]$ has the same integrable
operator bound. Reversed-time Carath\'eodory theory therefore gives a
unique bounded backward solution with terminal value $W_{\tout}r$.
All integrals are Bochner integrals; measurability follows from
separability and continuity in the data variables where required
\cite[Sections 1.1--1.2]{hytonen2016analysis}.

\paragraph{Step 2. First-order differences.}
Let $\xi:=(\delta W_{\tin},\delta W_{\tout},\delta u,\delta v)$ and
$\varepsilon:=\|\xi\|_{\Theta_2}\le1$. A superscript $+$ denotes
$\Delta\theta+\xi$, with $\delta h:=h^+-h$ and $\delta p:=p^+-p$.
The exact identities are
\begin{align}\label{eq:projection-perturbation}
 \delta p&=A_u\delta h+\langle\delta u,h^+\rangle,\notag\\
 \partial_s\delta h&=A_v^*[\rho(p^+)-\rho(p)]
                  +\EE[\delta v\rho(p^+)],
 &\delta h(0,x)&=\delta W_{\tin}x.
\end{align}
The forcing in the second equation is $O(\varepsilon)$ in $L^1$ of
depth, by Cauchy--Schwarz on the products of parameter norms.
Gronwall and the first identity give
\begin{equation}\label{eq:gradient-short-3}
 \sup_{s,x}\|\delta h(s,x)\|
 +\sup_x\|\delta p(\cdot,x)\|_{L^2(ds\,d\PP)}\le C\varepsilon.
\end{equation}

\paragraph{Step 3. The nonlinear remainder.}
The remainder $r_\rho:=\rho(p^+)-\rho(p)-\rho'(p)\delta p$ satisfies
$|r_\rho|\le C\min\{|\delta p|^2,|\delta p|\}\le C|\delta p|^{3/2}$.
Since $v_0:\Hh\to L^4$, duality and H\"older in depth give
\[
 \sup_x\int_0^1\|v_0^*r_\rho(s,x)\|\,ds
 \le C\sup_x\|\delta p(\cdot,x)\|_{L^2(ds\,d\PP)}^{3/2}
 \le C\varepsilon^{3/2}.
\]
For the learned part, truncate $\|\Delta v(s,\omega)\|$ at $T$ on the
joint depth--neuron probability space. Then
\[
 \sup_x\int_0^1\|\EE[\Delta v\,r_\rho]\|\,ds
 \le CT\varepsilon^2+C\varepsilon
 \|\Delta v\mathbf1_{\{\|\Delta v\|>T\}}\|_{L^2(ds\,d\PP)}.
\]
The last norm tends to zero at the fixed parameter as $T\to\infty$.
First sending $\varepsilon\to0$, then $T\to\infty$, proves
\begin{equation}\label{eq:gradient-short-1}
 \sup_x\int_0^1\|A_v^*r_\rho(s,x)\|\,ds=o(\varepsilon).
\end{equation}

\paragraph{Step 4. Linearization and the risk.}
Let $\dot h$ solve
\begin{equation}\label{eq:gradient-short-2}
 \dot h(0,x)=\delta W_{\tin}x,\qquad
 \partial_s\dot h=J\dot h
 +A_v^*[\rho'(p)\langle\delta u,h\rangle]+\EE[\delta v\rho(p)].
\end{equation}
Its forcing is $O(\varepsilon)$ in $L^1_s$, so $\sup_{s,x}\|\dot h\|\le
C\varepsilon$. The equation for $h^+-h-\dot h$ has forcing
\[
 A_v^*[\rho'(p)\langle\delta u,\delta h\rangle]
 +\EE[\delta v(\rho(p^+)-\rho(p))]+A_v^*r_\rho.
\]
The first two terms are $O(\varepsilon^2)$ in $L^1_s$, using
\eqref{eq:gradient-short-3} and Cauchy--Schwarz in $(s,\omega)$; the
last is $o(\varepsilon)$. Gronwall yields
\begin{equation}\label{eq:prediction-linearization}
 f^+-f=\delta W_{\tout}^*h(1)+W_{\tout}^*\dot h(1)+o(\varepsilon),
\end{equation}
uniformly over the data. The Lipschitz loss gradient therefore gives
\[
 D_\Delta\Ll(\theta)[\xi]
 =\int\langle r,\delta W_{\tout}^*h(1)+W_{\tout}^*\dot h(1)\rangle\,d\mu.
\]

\paragraph{Step 5. The adjoint calculation.}
The product rule and cancellation of $J,J^*$ give
\[
 \partial_s\langle b,\dot h\rangle
 =\EE[\rho'(p)q\langle h,\delta u\rangle
              +\rho(p)\langle b,\delta v\rangle].
\]
Integrate in depth and data and use $b(1)=W_{\tout}r$ and
$\dot h(0)=\delta W_{\tin}x$. The rank-one identity
$\langle a\otimes b,T\rangle_{\HS}=\langle a,Tb\rangle$ gives the four
components in~\eqref{eq:population-gradients}. Their embedding norms are
bounded, and
\[
 \|\nabla_u\Ll\|_{L^2(ds\,d\PP)}+
 \|\nabla_v\Ll\|_{L^2(ds\,d\PP)}
 \le C\bigl(1+\|\Delta u\|_{L^2(ds\,d\PP)}
               +\|\Delta v\|_{L^2(ds\,d\PP)}\bigr).
\]
Thus the derivative is represented in $\Theta_2$, the GD Hilbert space.
\end{proof}

\subsection{Moment bounds through training}\label{app:moment-stability}
\begin{lemma}[Bounds through a fixed number of updates]\label{lem:moment-propagation}
Under Assumption~\ref{ass:data-loss} and~\eqref{eq:seed-moments} at every
finite order, exact GD from zero is uniquely defined through every
$K\in\mathbb N_0$. For $k\le K$,
\begin{align}
 \sup_{s,z}(\|h_k(s,x)\|+\|b_k(s,z)\|)
 +\|\Delta {W_{\tin,k}}\|_{\HS}+\|\Delta {W_{\tout,k}}\|_{\HS}
 &\le C_{B,K},\label{eq:population-bounds}\\
 \sup_s(\|\Delta u_k(s)\|_{L^p}+\|\Delta v_k(s)\|_{L^p})
 +\sup_{s,z}(\|p_k(s,x)\|_{L^p}+\|q_k(s,z)\|_{L^p})
 &\le C_{B,K}(1+B_p).\label{eq:population-moments}
\end{align}
If $B_p\le B\sqrt p$, the latter bound is $C_{B,K}\sqrt p$.
The states are uniformly Lipschitz in depth, with constant $C_{B,K}$.
For every finite $p$, the internal increments and projections have
$L^p$-valued representatives that are Lipschitz in depth, with constant
$C_{B,K,p}$.
\end{lemma}

\begin{proof}
We first close an induction using second moments only. Higher moments
then satisfy a linear recursion whose coefficients are already bounded.
Finally we compare two depth values in the update formulas.

\paragraph{Step 1. Bounds for one forward and backward pass.}
Set
\[
 a_{k,p}:=\sup_s(\|\Delta u_k(s)\|_{L^p}+\|\Delta v_k(s)\|_{L^p}),\quad
 e_k:=\|\Delta {W_{\tin,k}}\|_{\HS},\quad t_k:=B+\|\Delta {W_{\tout,k}}\|_{\HS},
\]
and let $H_k:=\sup_{s,x}\|h_k(s,x)\|$, $J_k:=\sup_{s,z}\|b_k(s,z)\|$.
The operator bounds and Gronwall give
\begin{equation}\label{eq:forward-backward-gronwall-bounds}
 H_k\le C_B(1+e_k+a_{k,2})e^{C_B(1+a_{k,2})^2},\qquad
 J_k\le Bt_k(1+t_kH_k)e^{C_B(1+a_{k,2})^2}.
\end{equation}
For example, the forward velocity has norm at most
$C_B(1+a_{k,2})[1+(1+a_{k,2})\|h_k\|]$; the terminal backward state
has norm at most $Bt_k(1+t_kH_k)$. The projections satisfy
\[
 \|p_k(s,x)\|_{L^p}\le H_k(B_p+\|\Delta u_k(s)\|_{L^p}),\qquad
 \|q_k(s,z)\|_{L^p}\le J_k(B_p+\|\Delta v_k(s)\|_{L^p}).
\]

\paragraph{Step 2. Propagate every finite moment through GD.}
Applying Minkowski's inequality to the four gradients gives
\begin{align}\label{eq:moment-induction}
 a_{k+1,p}&\le a_{k,p}+\eta B J_k[1+H_k(2B_p+a_{k,p})],\notag\\
 e_{k+1}&\le e_k+\eta B J_k,
 &t_{k+1}&\le t_k+\eta B H_k(1+t_kH_k).
\end{align}
At $p=2$, these inequalities and~\eqref{eq:forward-backward-gronwall-bounds}
inductively bound $a_{k,2},e_k,t_k,H_k,J_k$ by $C_{B,K}$, starting from
$a_{0,2}=e_0=0$, $t_0=B$. Lemma~\ref{lem:gradient} therefore constructs
each next step in $\Theta_2$. At any $p\ge2$, the first inequality becomes
\[
 a_{k+1,p}\le(1+C_{B,K})a_{k,p}+C_{B,K}(1+B_p).
\]
Since $a_{0,p}=0$, summing the resulting geometric series proves
$a_{k,p}\le C_{B,K}(1+B_p)$. The projection estimates from Step 1 give
\eqref{eq:population-moments}, including the stated $\sqrt p$ consequence.

\paragraph{Step 3. Compare two depths.}
The state velocities are uniformly bounded by Step 1, hence
$\|h_k(s)-h_k(t)\|+\|b_k(s)-b_k(t)\|\le C_{B,K}|s-t|$.
Writing $U:=\Delta u_k$, the projection difference is
\[
 p_k(s)-p_k(t)=u_0(h_k(s)-h_k(t))
 +\langle U(s),h_k(s)-h_k(t)\rangle
 +\langle U(s)-U(t),h_k(t)\rangle.
\]
The analogous identity for $q_k$ gives, for $r\ge2$,
\begin{align*}
 &\sup_z(\|p_k(s)-p_k(t)\|_{L^r}+\|q_k(s)-q_k(t)\|_{L^r})\\
 &\qquad\le C_{B,K,r}\big(|s-t|+\|\Delta u_k(s)-\Delta u_k(t)\|_{L^r}
                       +\|\Delta v_k(s)-\Delta v_k(t)\|_{L^r}\big).
\end{align*}
In the $u$-gradient, subtract the integrands at $s,t$ as
\begin{align*}
 h_s\rho'(p_s)q_s-h_t\rho'(p_t)q_t
 ={}&(h_s-h_t)\rho'(p_s)q_s
       +h_t\rho'(p_s)(q_s-q_t)\\
    &+h_tq_t[\rho'(p_s)-\rho'(p_t)].
\end{align*}
The last term uses H\"older exponents $2,2$ in $L^p$; the $v$-gradient
has the two analogous terms with $b\rho(p)$. Thus
\begin{equation}\label{eq:propagation_regularity_internal_params}
 \sum_{a\in\{u,v\}}\|\Delta a_{k+1}(s)-\Delta a_{k+1}(t)\|_{L^p}
 \le C_{B,K,p}\left(|s-t|
       +\sum_{a\in\{u,v\}}\|\Delta a_k(s)-\Delta a_k(t)\|_{L^{2p}}\right).
\end{equation}
Induction from the constant zero increments proves the claimed depth
regularity at every finite order, and the preceding projection identity
then proves it for $p_k,q_k$.
\end{proof}

\subsection{State comparison and stability}\label{sec:stability}

For the uniform estimates in this subsection, use
\begin{equation}\label{eq:embedding-discrepancy}
 d_p:=\|\Delta\theta-\Delta\widetilde\theta\|_{\Thetainf{p}},\qquad
 \epsilon:=\|W_{\tin,0}-\widetilde W_{\tin,0}\|_{L^2(\mu;\Hh)}
       +\|W_{\tout,0}-\widetilde W_{\tout,0}\|_{\mathrm{op}}.
\end{equation}
The initialization maps are common. These auxiliary symbols separate
the fixed embedding discrepancy from the changing increments in the
induction; the main statements use the full parameter distance.

\begin{lemma}[State comparison]\label{lem:state-comparison}
Assume both initializations satisfy Assumption~\ref{ass:data-loss}, share $u_0,v_0$ with $B_4<\infty$, and have increment norms in $\Thetainf{4}$ bounded by $A$. Let the
first states solve the exact equations and the second states solve
\eqref{eq:stability-forced-equations} at a fixed parameter $\widetilde\theta$, with depth
errors $e^h,e^b\in L^1([0,1];L^2(\mu;\Hh))$. With $d_2,\epsilon$ as
in~\eqref{eq:embedding-discrepancy},
\begin{equation}\label{eq:states-close}
 \Big\|\sup_s(\|h-\widetilde h\|+\|b-\widetilde b\|)\Big\|_{L^2(\mu)}
 +\|f-\widetilde f\|_{L^2(\mu)}
 \le C_{B,A}(d_2+\epsilon+\|e^h\|_{L^1_sL^2_\mu}
                                  +\|e^b\|_{L^1_sL^2_\mu}).
\end{equation}
For any order $r$ for which $B_r$ and the two $\Thetainf{r}$ norms are
bounded, the projection differences satisfy
\begin{equation}\label{eq:projection-differences}
 \|p(s,z)-\widetilde p(s,z)\|_{L^r}
 +\|q(s,z)-\widetilde q(s,z)\|_{L^r}
 \le C_r(\|h(s,z)-\widetilde h(s,z)\|
             +\|b(s,z)-\widetilde b(s,z)\|+d_r).
\end{equation}
\end{lemma}

\begin{proof}
The forward comparison uses second moments. For the backward comparison,
we test the averaged product against a hidden vector before using
H\"older's inequality; this keeps the parameter discrepancy in $L^2$.
Write $\delta h:=h-\widetilde h$, $\delta b:=b-\widetilde b$,
$\delta p:=p-\widetilde p$, $\delta q:=q-\widetilde q$, and
$D_a:=\Delta a-\Delta\widetilde a$.

\paragraph{Step 1. Projections and forward velocity.}
The useful decompositions are
\[
 \delta p=\widetilde A_u\delta h+\langle D_u,h\rangle,
 \qquad\delta q=\widetilde A_v\delta b+\langle D_v,b\rangle.
\]
The exact states $h,b$ are uniformly bounded using Lemma~\ref{lem:moment-propagation}.
These identities immediately give~\eqref{eq:projection-differences}.
If $F:=A_v^*\rho(p)$ and $\widetilde F:=\widetilde A_v^*\rho(\widetilde p)$,
then
\[
 F-\widetilde F=\EE[D_v\rho(p)]
                   +\widetilde A_v^*[\rho(p)-\rho(\widetilde p)],
 \qquad\|F-\widetilde F\|\le C_{B,A}(d_2+\|\delta h\|).
\]
Consequently forward Gronwall gives, for almost every $z$,
\begin{equation}\label{eq:forward-state-comparison}
 \sup_s\|\delta h(s,x)\|
 \le C_{B,A}\left(\|({W_{\tin,0}}-\widetilde W_{\tin,0})(x)\|+d_2
                             +\int_0^1\|e^h(s,z)\|\,ds\right).
\end{equation}

\paragraph{Step 2. Backward velocity and terminal value.}
Set $H:=A_u^*[\rho'(p)q]$ and
$\widetilde H:=\widetilde A_u^*[\rho'(\widetilde p)\widetilde q]$.
Their difference is the sum of three explicit terms:
\begin{align*}
 H-\widetilde H={}&\EE[D_u\rho'(p)q]
 +\widetilde A_u^*[\rho'(\widetilde p)\delta q]
 +\widetilde A_u^*[q(\rho'(p)-\rho'(\widetilde p))].
\end{align*}
The first two have norms at most $C(d_2+\|\delta b\|)$.
For the last, duality and H\"older exponents $4,4,2$ give
\begin{align*}
 \|\widetilde A_u^*[q(\rho'(p)-\rho'(\widetilde p))]\|
 &\le B\sup_{\|w\|\le1}
    \|\widetilde A_uw\|_{L^4}\|q\|_{L^4}\|\delta p\|_{L^2}\\
 &\le C_{B,A}(d_2+\|\delta h\|).
\end{align*}
Thus $\|H-\widetilde H\|\le C(d_2+\|\delta h\|+\|\delta b\|)$.
With $\delta {W_{\tout}}:={W_{\tout}}-\widetilde W_{\tout}$, the endpoint differences obey
\[
 f-\widetilde f=\widetilde W_{\tout}^*\delta h(1)+\delta {W_{\tout}}^*\,h(1),\qquad
 b(1)-\widetilde b(1)=\widetilde W_{\tout}(r-\widetilde r)+\delta {W_{\tout}}r,
\]
so both are bounded by
$C(\|\delta h(1)\|+d_2+\|{W_{\tout,0}}-\widetilde W_{\tout,0}\|_{\mathrm{op}})$.
Backward Gronwall therefore yields
\begin{equation}\label{eq:backward-state-comparison}
 \sup_s\|\delta b(s,z)\|
 \le C\left(\sup_s\|\delta h(s,x)\|+d_2
   +\|{W_{\tout,0}}-\widetilde W_{\tout,0}\|_{\mathrm{op}}
   +\int_0^1\|e^b(s,z)\|\,ds\right).
\end{equation}
Take $L^2(\mu)$ norms in~\eqref{eq:forward-state-comparison} and
\eqref{eq:backward-state-comparison}, use Minkowski for the error
integrals, and use the prediction identity. This proves~\eqref{eq:states-close}.
\end{proof}

\paragraph{Stability through training.}
To cover discretization and averaging errors, we allow a second
trajectory whose forward and backward velocities have additive errors
$e_k^h,e_k^b$ and whose parameter update has error $e_k^\theta$:
\begin{align}\label{eq:stability-forced-equations}
 \partial_s\widetilde h_k
 &=\widetilde A_{v,k}^*\rho(\widetilde p_k)+e_k^h,
 &\widetilde h_k(0)&=\widetilde W_{\tin,0}(x)+\Delta\widetilde W_{\tin,k}x,\notag\\
 -\partial_s\widetilde b_k
 &=\widetilde A_{u,k}^*[\rho'(\widetilde p_k)\widetilde q_k]+e_k^b,
 &\widetilde b_k(1)&=\widetilde W_{\tout,k}\widetilde r_k,\notag\\
 \Delta\widetilde\theta_{k+1}
 &=\Delta\widetilde\theta_k-\eta\widetilde g_k+e_k^\theta,
 &\Delta\widetilde\theta_0&=0.
\end{align}
Here $\widetilde p_k:=\widetilde A_{u,k}\widetilde h_k$,
$\widetilde q_k:=\widetilde A_{v,k}\widetilde b_k$,
$\widetilde f_k:=\widetilde W_{\tout,k}^*\widetilde h_k(1)$ and
$\widetilde r_k:=\nabla\loss_y(\widetilde f_k)$.
The four components of $\widetilde g_k$ are the integrals
in~\eqref{eq:population-gradients} evaluated at these approximate states.
The depth errors are deterministic in $\omega$; $e_k^\theta$ is the
actual additive increment, including any learning-rate factor.
For $m:=\max\{4,2^{K+1}\}$ write
\begin{equation}\label{eq:stability-residual}
 \varepsilon_{\mathrm{res}}
 :=\max_{k\le K}(\|e_k^h\|_{L^1_sL^2_\mu}
                  +\|e_k^b\|_{L^1_sL^2_\mu})
   +\sum_{k<K}\|e_k^\theta\|_{\Thetainf{m}}.
\end{equation}
These errors measure how accurately an approximation satisfies the
population equations, so the next estimate converts equation errors
into prediction errors.

\paragraph{Stability through gradient descent.}
\begin{theorem}[Stability through a fixed number of updates]
\label{thm:population-stability}\label{prop:perturbed-stability}
Assume Assumption~\ref{ass:data-loss} for both initializations,
common $u_0,v_0$ with $B_m<\infty$, and common learning rate.
Compare exact GD from zero with~\eqref{eq:stability-forced-equations},
also starting from zero increments. Suppose both trajectories have
states bounded by $A$ uniformly in $k,s,z$ and increment norms in
$\Thetainf{m}$ bounded by $A_m$, where $m:=\max\{4,2^{K+1}\}$.
Then the exact and forced trajectories satisfy
\begin{equation}\label{eq:population-forced-error}
 \begin{aligned}
 \max_{k\le K}\Big[&
 \|\Delta\theta_k-\Delta\widetilde\theta_k\|_{\Thetainf{2}}
 +\Big\|\sup_s(\|h_k-\widetilde h_k\|+\|b_k-\widetilde b_k\|)\Big\|_{L^2(\mu)}
 +\|f_k-\widetilde f_k\|_{L^2(\mu)}\Big]\\
 &\le C_{B,A,A_m,K}(\epsilon+\varepsilon_{\mathrm{res}}).
 \end{aligned}
\end{equation}
For two exact trajectories with initialization moments of every finite
order, the same left-hand side satisfies
\begin{equation}\label{eq:population-error}
 \begin{aligned}
 \max_{k\le K}\Big[&
 \|\Delta\theta_k-\Delta\widetilde\theta_k\|_{\Thetainf{2}}
 +\Big\|\sup_s(\|h_k-\widetilde h_k\|+\|b_k-\widetilde b_k\|)\Big\|_{L^2(\mu)}
 +\|f_k-\widetilde f_k\|_{L^2(\mu)}\Big] \le C_{B,K}\epsilon.
 \end{aligned}
\end{equation}
\end{theorem}

\begin{proof}
We compare one update in $\Thetainf{p}$ using a parameter discrepancy in
$\Thetainf{2p}$. Repeatedly doubling the moment order then controls the
final $\Thetainf{2}$ discrepancy without weakening its linear dependence
on the equation errors. Set
$d_{k,p}:=\|\Delta\theta_k-\Delta\widetilde\theta_k\|_{\Thetainf{p}}$ and
$m:=\max\{4,2^{K+1}\}$.

\paragraph{Step 1. Compare one forward pass, backward pass and update.}
Lemma~\ref{lem:state-comparison} gives
\begin{equation}\label{eq:training-state-comparison}
 \Big\|\sup_s(\|h_k-\widetilde h_k\|+\|b_k-\widetilde b_k\|)\Big\|_{L^2(\mu)}
 +\|f_k-\widetilde f_k\|_{L^2(\mu)}\le C(d_{k,2}+\epsilon+\varepsilon_{\mathrm{res}}).
\end{equation}
The decomposition~\eqref{eq:u-gradient-difference} applies to the
approximate update directions as well. More explicitly, its three
integrands have $L^p(\Omega;\Hh)$ norms at most
\[
 C_p\|\delta h_k\|,\qquad
 C_p(\|\delta b_k\|+d_{k,p}),\qquad
 C_p\|\widetilde q_k\|_{L^{2p}}\|\delta p_k\|_{L^{2p}}
       \le C_p(\|\delta h_k\|+d_{k,2p}),
\]
respectively. The $v$ and embedding decompositions from the same proof
have only the first two types of terms. Integrating in $\mu$, taking the
depth supremum and using~\eqref{eq:training-state-comparison} yields
$\|g_k-\widetilde g_k\|_{\Thetainf{p}}\le C_p(d_{k,2p}+\epsilon+\varepsilon_{\mathrm{res}})$.
The additive update error then gives
\begin{equation}\label{eq:training-comparison}
 d_{k+1,p}\le d_{k,p}+\eta C_p(d_{k,2p}+\epsilon+\varepsilon_{\mathrm{res}})
                     +\|e_k^\theta\|_{\Thetainf{p}},\qquad2\le p\le m/2.
\end{equation}
All constants use only the assumed state and $m$th-moment bounds.

\paragraph{Step 2. Close the finite sequence of moment estimates.}
For $K\ge1$ define
$M_k:=\max_{1\le j\le K+1-k}d_{k,2^j}$, for $0\le k\le K$.
The index set is nonempty and every order is at most $m$.
For each order entering $M_{k+1}$, its doubled order enters $M_k$.
Since $d_{0,p}=0$ and $\|e_k^\theta\|_{\Thetainf{p}}\le
\varepsilon_{\mathrm{res}}$,~\eqref{eq:training-comparison} implies
\[
 M_0=0,\qquad M_{k+1}\le(1+C)M_k+C(\epsilon+\varepsilon_{\mathrm{res}}),
 \qquad M_k\le[(1+C)^k-1](\epsilon+\varepsilon_{\mathrm{res}}).
\]
In particular $d_{k,2}\le C_K(\epsilon+\varepsilon_{\mathrm{res}})$ for every $k\le K$.
Substitution into~\eqref{eq:training-state-comparison} proves
\eqref{eq:population-forced-error}. For $K=0$, the parameter discrepancy
is zero and the state lemma gives the result directly; the imposed
$m\ge4$ still supplies its needed fourth moments.

For exact trajectories all errors vanish, and
Lemma~\ref{lem:moment-propagation} supplies the uniform state and moment
bounds with constants determined by initialization. This proves~\eqref{eq:population-error}, which also yields the initialization-stability estimate stated in Proposition~\ref{prop:embedding-error}.
\end{proof}

\paragraph{Changing the initialization law.}
Let $(\bar h_k,\bar b_k)$ be a Gaussian reference trajectory, embedded
in the space carrying general maps $u_0,v_0$.
The superscript $\mathrm{rep}$ labels the replay of these fixed
state histories under this initialization. On their probability
space start from $\Delta\bar u_0^{\mathrm{rep}}=\Delta\bar v_0^{\mathrm{rep}}=0$ and set
\begin{align}\label{eq:changed-law-recursion}
 \bar p_k^{\mathrm{rep}}&:=u_0(\bar h_k)+\langle\Delta\bar u_k^{\mathrm{rep}},\bar h_k\rangle,
 &\bar q_k^{\mathrm{rep}}&:=v_0(\bar b_k)+\langle\Delta\bar v_k^{\mathrm{rep}},\bar b_k\rangle,\notag\\
 \Delta\bar u_{k+1}^{\mathrm{rep}}
 &=\Delta\bar u_k^{\mathrm{rep}}-\eta\int\bar h_k\rho'(\bar p_k^{\mathrm{rep}})\bar q_k^{\mathrm{rep}}\,d\mu,
 &\Delta\bar v_{k+1}^{\mathrm{rep}}
 &=\Delta\bar v_k^{\mathrm{rep}}-\eta\int\bar b_k\rho(\bar p_k^{\mathrm{rep}})\,d\mu.
\end{align}
Keep the reference embeddings $\bar W_{\tin,k},\bar W_{\tout,k}$, whose updates depend
only on the fixed states. With
$\bar A_{a,k}^{\mathrm{rep}}h:=a_0(h)+\langle\Delta\bar a_k^{\mathrm{rep}},h\rangle$, the errors are
\begin{equation}\label{eq:changed-law-errors}
 e_k^h:=\partial_s\bar h_k-(\bar A_{v,k}^{\mathrm{rep}})^*\rho(\bar p_k^{\mathrm{rep}}),\qquad
 e_k^b:=-\partial_s\bar b_k-(\bar A_{u,k}^{\mathrm{rep}})^*
       [\rho'(\bar p_k^{\mathrm{rep}})\bar q_k^{\mathrm{rep}}],\qquad e_k^\theta:=0.
\end{equation}
This is a trajectory of~\eqref{eq:stability-forced-equations} on the same
probability space as the actual dynamics initialized by $u_0,v_0$. The replacement estimates
in Appendix~\ref{app:gaussianization} bound these averaged velocity
errors, including components outside a retained subspace; the stability
proposition then propagates them through training.

\subsection{Proof of gradient regularity}
\label{app:gradient-regularity}\label{app:normalized-update}
\begin{proof}[Proof of Proposition~\ref{prop:gradient-regularity}]
Throughout this proof, the parameter norms integrate over depth.
Write $\delta$ for differences.
The common initialization maps are fixed. Define
$Q(s):=(\int|q(s,z)|^2\,d\mu(z))^{1/2}$ and $\widetilde Q$ likewise;
these are random variables on $\Omega$ at each depth.
Steps 1--3 use only fourth-moment bounds. The subgaussian assumption
enters in Step 4.

\paragraph{Step 1. State comparison with integrated parameter norms.}
The pointwise decompositions in Lemma~\ref{lem:state-comparison} remain
valid. Here their coefficients must be integrated before they are
bounded. In the forward comparison they are products of at most two
factors of the form $B+\|\Delta a(s)\|_{L^2}$, which are integrable;
coefficients multiplying a parameter difference are in $L^2_s$.
For the backward comparison, the only extra term is bounded by
\[
 B\sup_{\|w\|\le1}\|\widetilde A_u(s)w\|_{L^4}
       \|q(s,z)\|_{L^4}\|\delta p(s,z)\|_{L^2}.
\]
Using $\delta p=\widetilde A_u\delta h+
\langle\Delta u-\Delta\widetilde u,h\rangle$, its state coefficient is
a product of three $L^4_s$ factors and its parameter coefficient a
product of two $L^4_s$ factors. The former is in $L^1_s$ and the latter
in $L^2_s$. The bounds on $\Theta_4$, H\"older and Gronwall give
\begin{equation}\label{eq:stability-state-comparison}
 \Big\|\sup_s(\|\delta h\|+\|\delta b\|)\Big\|_{L^2(\mu)}
 +\|\delta f\|_{L^2(\mu)}\le C\|\theta-\widetilde\theta\|_2.
\end{equation}
The estimates use the exact states in parameter-difference terms, so
only their uniform bounds are needed before applying Gronwall.
The same argument allows additive velocity errors, with their
$L^1_sL^2_\mu$ norms added to the right-hand side.

\paragraph{Step 2. Embedding and $v$ gradients.}
The embedding differences are
\[
 \delta g_{W_{\tin}}:=\int\delta b(0,z)\otimes x\,d\mu,\qquad
 \delta g_{W_{\tout}}:=\int[h(1)\otimes\delta r+\delta h(1)\otimes\widetilde r]\,d\mu.
\]
They have Hilbert--Schmidt norm at most $C\|\theta-\widetilde\theta\|_2$. Likewise
$\delta g_v:=\int[\delta b\rho(p)+\widetilde b(\rho(p)-\rho(\widetilde p))]\,d\mu$.
Minkowski, the projection identities, and Step 1 give
\begin{equation}\label{eq:v-gradient-comparison}
 \|\delta g_v\|_{L^2(ds\,d\PP)}\le C\|\theta-\widetilde\theta\|_2.
\end{equation}
At order $p$, the same reasoning bounds this norm by
$C_p\|\theta-\widetilde\theta\|_p$ on bounded $\Theta_p$ sets.

\paragraph{Step 3. The random multiplier in the $u$ gradient.}
Subtracting the three factors gives
\begin{equation}\label{eq:u-gradient-difference}
 \delta g_u:=\int[\delta h\rho'(p)q+\widetilde h\rho'(p)\delta q
       +\widetilde h\widetilde q(\rho'(p)-\rho'(\widetilde p))]\,d\mu.
\end{equation}
The first two terms have joint $L^2(ds\,d\PP)$ norm at most $C\|\theta-\widetilde\theta\|_2$.
Split the last activation difference into the change of state at fixed
$A_u$ and the change from $A_u$ to $\widetilde A_u$ at fixed
$\widetilde h$. The state contribution is at most
$C\|\widetilde q(s,z)\|_{L^4}\|A_u(s)\delta h(s,x)\|_{L^4}$.
H\"older in depth with exponents $4,4$, followed by Step 1, bounds its
joint $L^2$ norm by $C\|\theta-\widetilde\theta\|_2$. Boundedness and Lipschitz continuity of $\rho'$
bound the remaining term pointwise by
$C\widetilde Q\min\{1,\|\Delta u-\Delta\widetilde u\|_\Hh\}$.
Its joint $L^2$ norm is at most
\begin{equation}\label{eq:gradient-tail-product}
 C\left[T\|\Delta u-\Delta\widetilde u\|_{L^2(ds\,d\PP)}
 +\|\widetilde Q\mathbf1_{\{\widetilde Q>T\}}\|_{L^2(ds\,d\PP)}\right].
\end{equation}
Define
\[
 \mathcal T(T):=\|Q\mathbf1_{\{Q>T\}}\|_{L^2(ds\,d\PP)}
 +\|\widetilde Q\mathbf1_{\{\widetilde Q>T\}}\|_{L^2(ds\,d\PP)}.
\]
Adding the preceding estimates and taking the infimum over $T$ gives
\begin{equation}\label{eq:gradient-tail-profile}
 \begin{aligned}
 \|\nabla\Ll(\theta)-\nabla\Ll(\widetilde\theta)\|_{\Theta_2}
 \le C\bigg[\|\theta-\widetilde\theta\|_2 +\inf_{T\ge1}\{T\|\theta-\widetilde\theta\|_2+\mathcal T(T)\}\bigg].
 \end{aligned}
\end{equation}
Alternatively, H\"older on the joint probability space gives
\begin{equation}\label{eq:higher-gradient}
 \|\nabla\Ll(\theta)-\nabla\Ll(\widetilde\theta)\|_{\Theta_p}
 \le C_{B,A_{2p},p}\|\theta-\widetilde\theta\|_{2p},\qquad p\ge2,
\end{equation}
on bounded $\Theta_{2p}$ sets when $B_{2p}<\infty$.
Taking depth suprema instead gives the same estimate with
$\Theta_p,\Theta_{2p}$ replaced by $\Thetainf{p},\Thetainf{2p}$ and
with $d_{2p}+\epsilon$ on the right, in the notation of
\eqref{eq:embedding-discrepancy}. Both versions also apply to update
directions formed from bounded forced states, with the state error
included explicitly. This is the version used in Appendix~\ref{sec:stability}.

\paragraph{Step 4. Backward-projection tails.}\label{lem:feedback-tails}
Minkowski in $L^{p/2}$ and the projection identity imply
\begin{equation}\label{eq:feedback-moment-bound}
 \|Q\|_{L^p(ds\,d\PP)}
 \le (B_p+\|\Delta v\|_{L^p(ds\,d\PP)})\sup_s\|b(s,\cdot)\|_{L^2(\mu)}.
\end{equation}
The parameter assumption~\eqref{eq:parameter-subgaussian} gives
$B_p\le A\sqrt p$ and
$\|\Delta v\|_{L^p(ds\,d\PP)}+\|\Delta\widetilde v\|_{L^p(ds\,d\PP)}
\le A\sqrt p$. The uniform state bounds therefore imply
$\|Q\|_{L^p(ds\,d\PP)}+\|\widetilde Q\|_{L^p(ds\,d\PP)}
\le C_{B,A}\sqrt p$.
This yields a Gaussian tail \cite[Proposition 2.5.2]{vershynin2018high}, hence
\begin{equation}\label{eq:feedback-exponential-tail}
 \mathcal T(T)\le Ce^{-cT^2}\qquad(T\ge C),
\end{equation}
where $C,c>0$ depend only on $B,A$.
For $0<\|\theta-\widetilde\theta\|_2\le1$, choose $T$ to be a sufficiently
large constant multiple of
$\sqrt{\log(e/\|\theta-\widetilde\theta\|_2)}$ in
\eqref{eq:gradient-tail-profile}. The tail term is then bounded by
$C\|\theta-\widetilde\theta\|_2$, and the other terms are bounded by
$C\omega(\|\theta-\widetilde\theta\|_2)$, proving
\eqref{eq:feedback-gradient-moduli}. If $\|\theta-\widetilde\theta\|_2=0$,
let $T\to\infty$ in~\eqref{eq:gradient-tail-profile}.

\end{proof}

\begin{proof}[Proof of~\eqref{eq:normalized-lipschitz}]
Only the stated fourth-moment bounds are used here; Steps 1 and 2 of the
proposition's proof remain applicable under these bounds.
On $L^2(\mu)$, the map $N(q):=q/\sqrt{1+\|q\|_{L^2(\mu)}^2}$ has norm at
most one and is one-Lipschitz, since
\[
 DN(q)[v]=\frac{v}{\sqrt{1+\|q\|^2}}
 -\frac{q\langle q,v\rangle}{(1+\|q\|^2)^{3/2}},\qquad
 \|DN(q)\|_{\mathrm{op}}\le1.
\]
At each $(s,\omega)$, write $g_u^{\mathrm{cond}}:=\int h\rho'(p)N(q)\,d\mu$.
Its difference is the sum of
\[
 \int[\delta h\rho'(p)N(q)
 +\widetilde h(\rho'(p)-\rho'(\widetilde p))N(q)
 +\widetilde h\rho'(\widetilde p)(N(q)-N(\widetilde q))]\,d\mu.
\]
Cauchy--Schwarz in the data bounds its norm by
$C(\|\delta h\|_{L^2(\mu)}+\|\delta p\|_{L^2(\mu)}+
\|\delta q\|_{L^2(\mu)})$. Taking joint $L^2(ds\,d\PP)$ and using the
projection identities and Step 1 proves~\eqref{eq:normalized-lipschitz}.
The other three components were treated in Step 2.
\end{proof}

\section{Approximation of embeddings}\label{app:geometry-main}
\subsection{Spectral approximation}

\begin{corollary}[Comparison after isometric embeddings]\label{cor:alignment}
Consider two Gaussian models satisfying Assumption~\ref{ass:data-loss}
with the same constant $B$, data law, loss, activation, variances and
learning rate. For linear isometries $J:\Hh\to\Hh_*$ and
$\widetilde J:\widetilde\Hh\to\Hh_*$ into a common Hilbert space,
\begin{equation}\label{eq:alignment-stability}
 \max_{k\le K}\|f_k-\widetilde f_k\|_{L^2(\mu)}
 \le C_{B,K}\big(
 \|J{W_{\tin,0}}-\widetilde J\widetilde W_{\tin,0}\|_{L^2(\mu;\Hh_*)}
 +\|J{W_{\tout,0}}-\widetilde J\widetilde W_{\tout,0}\|_{\mathrm{op}}\big).
\end{equation}
Under the coupling below, the state and parameter estimate
\eqref{eq:population-error} also holds after applying the two isometries.
One may take the infimum over common isometric embeddings.
\end{corollary}

\begin{proof}
We realize both initializations using common Gaussian maps on $\Hh_*$,
verify that the embedded states and parameters satisfy the same
equations, then apply the already proved stability theorem.

\paragraph{Step 1. Couple the Gaussian maps and identify their adjoints.}
Let $G_{a,*}:\Hh_*\to L^2(\Omega)$ be independent Gaussian maps
with covariance $\sigma_a^2\langle\cdot,\cdot\rangle_{\Hh_*}$. Define
$G_a(w):=G_{a,*}(Jw)$ and $a_0(\widetilde w):=G_{a,*}(\widetilde J\widetilde w)$.
Their covariances are the required ones because both embeddings are
isometries. In particular,
\[
 \EE[G_a(w)G_a(v)]=\sigma_a^2\langle Jw,Jv\rangle_{\Hh_*}
                          =\sigma_a^2\langle w,v\rangle_\Hh.
\]
For $\phi$ measurable with respect to Gaussian projections on $J\Hh$,
Lemma~\ref{lem:subspace} places $G_{a,*}^*\phi$ in $J\Hh$, and testing
against $Jw$ gives
\[
 \langle G_{a,*}^*\phi,Jw\rangle
 =\EE[\phi G_a(w)]=\langle JG_a^*\phi,Jw\rangle.
\]
Therefore $G_{a,*}^*\phi=JG_a^*\phi$. The same identity holds for the
learned adjoints when the increments are transformed by $J$.

\paragraph{Step 2. Transport the dynamics and compare.}
Transform the first model by
\begin{equation}\label{eq:HI_isometry}
 (h,b,\Delta {W_{\tin}},\Delta {W_{\tout}},\Delta u,\Delta v)
 \mapsto(Jh,Jb,J\Delta {W_{\tin}},J\Delta {W_{\tout}},J\Delta u,J\Delta v).
\end{equation}
Its initial embeddings are $J{W_{\tin,0}}$ and $J{W_{\tout,0}}$.
In the common space,
$A_{a,*}(s)w:=G_{a,*}(w)+\langle J\Delta a(s),w\rangle$.
The scalar projections are unchanged since
$\langle J\Delta a,Jw\rangle=\langle\Delta a,w\rangle$.
The adjoint identity from Step 1 gives
\[
 \partial_s(Jh)=A_{v,*}^*\rho(p),\qquad
 -\partial_s(Jb)=A_{u,*}^*[\rho'(p)q],\qquad
 (J{W_{\tout}})^*Jh(1)={W_{\tout}}^*h(1).
\]
The internal updates commute with $J$ because their scalar factors
$p,q$ are unchanged. For the embedding updates use
\[
 (Jb)\otimes x=J(b\otimes x),\qquad
 (Jh)\otimes r=J(h\otimes r).
\]
This verifies the four GD components in the common space. Also
$\|J\Delta {W_{\tin}}\|_{\HS}=\|\Delta {W_{\tin}}\|_{\HS}$,
$\|J\Delta {W_{\tout}}\|_{\HS}=\|\Delta {W_{\tout}}\|_{\HS}$, and
$\|J\Delta a\|_{L^p}=\|\Delta a\|_{L^p}$, so all assumptions and
parameter estimates are preserved. Perform the same construction with
$\widetilde J$, then apply~\eqref{eq:population-error} in $\Hh_*$.
The output-embedding discrepancy is
\[
 \|J{W_{\tout,0}}-\widetilde J\widetilde W_{\tout,0}\|_{\mathrm{op}}
 =\|(J{W_{\tout,0}})^*-(\widetilde J\widetilde W_{\tout,0})^*\|_{\mathrm{op}}.
\]
This proves~\eqref{eq:alignment-stability} and the stated stronger
comparison. Since the estimate holds for each pair of isometries,
taking an infimum requires no existence of an optimizing pair.
\end{proof}

Use the map $S_D$ defined in Section~\ref{sec:main-theorem}.
For $\psi\in L^2(\nu;\Hh)$, its finite-rank approximations converge in
$L^2(\Omega_{\rm emb}\times\nu;\Hh_D)$, since
$\EE_{\rm emb}\|S_D\psi\|_{L^2(\nu;\Hh_D)}^2
=\|\psi\|_{L^2(\nu;\Hh)}^2$.
This defines $S_D\psi$ even when a realization of $S_D$ is not bounded on $\Hh$.

\begin{lemma}[Spectral tails and Gaussian alignment]\label{lem:geometry}
Let $C$ be a positive trace-class operator on $\Hh$, with eigenvalues
$\lambda_1\ge\lambda_2\ge\cdots\ge0$, and let $P_r$ project onto
its first $r$ eigenvectors. If
$\psi\in L^2(\nu;\Hh)$ and $C=\int\psi\otimes_\Hh\psi\,d\nu$, then
\begin{equation}\label{eq:spectral-tail}
 \|(I-P_r)\psi\|_{L^2(\nu;\Hh)}^2
 =\operatorname{tr}((I-P_r)C)=\sum_{j>r}\lambda_j.
\end{equation}
There are absolute constants $C_0,C_1$ such that, for $t\ge1$ and
$D\ge C_1(r+t)$, with probability at least $1-2e^{-t}$ the restriction
$T:=S_D|_{P_r\Hh}$ has an isometric polar factor $J:=T(T^*T)^{-1/2}$ satisfying
\begin{equation}\label{eq:polar}
 \|T-J\|_{\mathrm{op}}\le C_0\sqrt{(r+t)/D},
 \qquad\|T\|_{\mathrm{op}}\le2.
\end{equation}
On the same event,
\begin{equation}\label{eq:polar-inverse}
 \|(T^*T)^{-1/2}\|_{\mathrm{op}}\le2.
\end{equation}
Moreover, if $\psi_\perp\in L^2(\nu;\Hh)$ has covariance
$(I-P_r)C(I-P_r)$, then, with probability at least $1-e^{-t}$,
\begin{equation}\label{eq:gaussian-tail}
 \begin{aligned}
 \|S_D\psi_\perp\|_{L^2(\nu;\Hh_D)}^2
 &\le\sum_{j>r}\lambda_j
 +2\sqrt{\frac tD\sum_{j>r}\lambda_j^2}
 +\frac{2t}D\lambda_{r+1}\\
 &\le(1+2\sqrt{t/D}+2t/D)\sum_{j>r}\lambda_j.
 \end{aligned}
\end{equation}

In particular, if $C=\int {W_{\tin}}(x)\otimes_\Hh {W_{\tin}}(x)\,d\mu+{W_{\tout}}{W_{\tout}}^*$ with
${W_{\tin}}\in L^2(\mu;\Hh)$ and ${W_{\tout}}\in\operatorname{HS}(\Yy,\Hh)$, then
\begin{equation}\label{eq:geometry-embeddings-tail}
 \|(I-P_r){W_{\tin}}\|_{L^2(\mu;\Hh)}^2+\|(I-P_r){W_{\tout}}\|_{\HS}^2
 =\sum_{j>r}\lambda_j.
\end{equation}
The sum
$\|S_D(I-P_r){W_{\tin}}\|_{L^2(\mu;\Hh_D)}^2+
\|S_D(I-P_r){W_{\tout}}\|_{\HS}^2$
satisfies the same upper bounds and probability estimate as
\eqref{eq:gaussian-tail}.

\end{lemma}

\begin{proof}
We compute the discarded squared norm, align the retained Gaussian
matrix with an isometry, and express the random discarded norm as
weighted Gaussian squares. The last calculation controls both embeddings
at once, including when their covariance is not block diagonal.

\paragraph{Step 1. Compute the spectral error.}
Let $(u_j)$ be an orthonormal eigenbasis on the support of $C$.
For every $w\in\ker C$,
\[
 0=\langle Cw,w\rangle_\Hh
 =\int|\langle\psi(z),w\rangle_\Hh|^2\,d\nu(z).
\]
Since $\Hh$ is separable, using a countable orthonormal basis of $\ker C$ shows that
$\psi(z)\in(\ker C)^\perp$ for $\nu$-almost every $z$.
Tonelli and Parseval give
\[
 \|(I-P_r)\psi\|_{L^2}^2
 =\sum_{j>r}\int|\langle\psi,u_j\rangle_\Hh|^2\,d\nu
 =\sum_{j>r}\langle Cu_j,u_j\rangle_\Hh
 =\sum_{j>r}\lambda_j.
\]
For the two embeddings, take an orthonormal basis $(e_\ell)$ of $\Yy$.
The coefficient identity
\[
 \int\langle {W_{\tin}},u_i\rangle_\Hh\langle {W_{\tin}},u_j\rangle_\Hh\,d\mu
 +\sum_\ell\langle {W_{\tout}}e_\ell,u_i\rangle_\Hh
              \langle {W_{\tout}}e_\ell,u_j\rangle_\Hh
 =\langle Cu_i,u_j\rangle_\Hh=\lambda_i\mathbf1_{\{i=j\}}
\]
implies~\eqref{eq:geometry-embeddings-tail} by summing over $i=j>r$.
The column series converges because
\[
 \sum_\ell\|{W_{\tout}}e_\ell\|_\Hh^2=\|{W_{\tout}}\|_{\HS}^2.
\]

\paragraph{Step 2. Apply the Gaussian singular-value bound.}
If $P_r\Hh=\{0\}$, all operators in this step act on the zero space
and the claims are immediate. Otherwise, in orthonormal coordinates,
$T:=G/\sqrt D$, where $G$ is a standard Gaussian matrix of size
$D\times\operatorname{rank}P_r$.
Corollary~5.35 of~\cite{vershynin2012introduction}, with deviation
$\sqrt{2t}$, yields, with failure probability at most $2e^{-t}$,
\[
 1-\gamma\le s_{\min}(T)\le s_{\max}(T)\le1+\gamma,
 \qquad \gamma:=\sqrt{r/D}+\sqrt{2t/D}\le\sqrt{3(r+t)/D}.
\]
For $D\ge12(r+t)$, $\gamma\le1/2$ and $J^*J=I$.
If $T=U\Sigma V^*$, then $J=UV^*$, hence
\[
 \|T-J\|_{\mathrm{op}}=\max_i|s_i(T)-1|\le\gamma,\quad
 \|T\|_{\mathrm{op}}\le1+\gamma\le2,\quad
 \|(T^*T)^{-1/2}\|_{\mathrm{op}}=s_{\min}(T)^{-1}\le2.
\]
This proves~\eqref{eq:polar}--\eqref{eq:polar-inverse}.

\paragraph{Step 3. Apply the weighted Gaussian-square bound.}
Set $\xi_{dj}:=g_d(u_j)$ and
$\psi_j:=\langle\psi_\perp,u_j\rangle_\Hh$, so
$\int\psi_i\psi_j\,d\nu=\lambda_i\mathbf1_{\{i=j\}}$.
For a finite expansion,
\[
 \frac1D\sum_{d=1}^D\int
       \left|\sum_{j=r+1}^n\xi_{dj}\psi_j\right|^2\,d\nu
 =\frac1D\sum_{d=1}^D\sum_{i,j=r+1}^n
       \xi_{di}\xi_{dj}\lambda_i\mathbf1_{\{i=j\}}
 =\frac1D\sum_{d=1}^D\sum_{j=r+1}^n\lambda_j\xi_{dj}^2.
\]
The infinite nonnegative sum has finite expectation $\sum_{j>r}\lambda_j$.
Thus the orthogonal series converges in $L^2(\nu)$ almost surely, and
\begin{equation}\label{eq:gaussian-tail-squares}
 X:=\|S_D\psi_\perp\|_{L^2(\nu;\Hh_D)}^2
   =\frac1D\sum_{d=1}^D\sum_{j>r}\lambda_j\xi_{dj}^2.
\end{equation}
For its partial sum $X_n$, Lemma~1 of~\cite{laurent2000adaptive}
applies with weights $w_{dj}:=\lambda_j/D$.
Since $\sum w_{dj}^2=D^{-1}\sum_{r<j\le n}\lambda_j^2$ and
$\max w_{dj}=\lambda_{r+1}/D$, it gives
\[
 \PP\{X_n>B_n\}\le e^{-t},\qquad
 B_n:=\sum_{r<j\le n}\lambda_j
   +2\sqrt{\frac tD\sum_{r<j\le n}\lambda_j^2}
   +\frac{2t}D\lambda_{r+1}.
\]
Here $X_n\uparrow X$ and $B_n\to B_\infty$, the first right-hand side
of~\eqref{eq:gaussian-tail}. On $\{X>B_\infty\}$, convergence of $X_n$ and $B_n$ implies
$X_n>B_n$ for every sufficiently large $n$. Thus Fatou gives $\PP\{X>B_\infty\}\le\liminf_n\PP\{X_n>B_n\}\le e^{-t}$.
Finally,
$\sum_{j>r}\lambda_j^2\le(\sum_{j>r}\lambda_j)^2$ and
$\lambda_{r+1}\le\sum_{j>r}\lambda_j$ give the second bound.

For the two embeddings, Step~1 shows that the coefficient of
$\xi_{di}\xi_{dj}$ in the sum of their squared norms is again
$\lambda_i\mathbf1_{\{i=j\}}$.
The same finite expansion and limit therefore give exactly $X$
in~\eqref{eq:gaussian-tail-squares}. The probability bound just proved
applies to their sum.
\end{proof}

\begin{lemma}[Covariance-weighted Gaussian alignment]\label{lem:weighted-gaussian-alignment}
Let $W_{\tin}\in L^2(\mu;\Hh)$ and $W_{\tout}\in\operatorname{HS}(\Yy,\Hh)$, with covariance
$C:=\int W_{\tin}(x)\otimes W_{\tin}(x)\,d\mu+W_{\tout}W_{\tout}^*$ and eigenvalues $(\lambda_j)_{j\ge1}$ in decreasing order. Enlarge $\Hh$ isometrically if needed, and define
\[
 E_{\rm emb}:=\inf_{J:\Hh_D\hookrightarrow\Hh}
 \left(\|JS_DW_{\tin}-W_{\tin}\|_{L^2(\mu;\Hh)}^2
       +\|JS_DW_{\tout}-W_{\tout}\|_{\HS}^2\right)^{1/2},
\]
where the infimum is over linear isometries. Then
\begin{equation}\label{eq:weighted-alignment-mean}
 \EE_{\rm emb}E_{\rm emb}^2
 \le \frac2D\sum_{j\le D}j\lambda_j+2\sum_{j>D}\lambda_j.
\end{equation}
For every $t>0$, with probability at least $1-e^{-t}$,
\begin{equation}\label{eq:weighted-alignment-tail}
 E_{\rm emb}\le
 \left(\frac2D\sum_{j\le D}j\lambda_j+2\sum_{j>D}\lambda_j\right)^{1/2}
 +\sqrt{\frac{2\lambda_1t}D}.
\end{equation}
\end{lemma}

\begin{proof}
Choose covariance eigenvectors $(e_j)$, padding with zero-eigenvalue directions in a larger space when necessary. Extend the isonormal maps independently to the added directions. Put $z_j:=S_De_j$; these are independent $\Nn(0,I_D/D)$ vectors. The covariance identity used in Lemma~\ref{lem:geometry} gives, for every isometry $J$,
\begin{equation}\label{eq:weighted-column-discrepancy}
 \|JS_DW_{\tin}-W_{\tin}\|_{L^2}^2
 +\|JS_DW_{\tout}-W_{\tout}\|_{\HS}^2
 =\sum_{j\ge1}\lambda_j\|Jz_j-e_j\|_\Hh^2.
\end{equation}

\paragraph{Step 1. Orthogonalize the Gaussian columns.}
Apply Gram--Schmidt to $z_1,\ldots,z_D$, obtaining an orthonormal basis $q_1,\ldots,q_D$ of $\Hh_D$ almost surely, and choose $Jq_j=e_j$. For $j\le D$, write $P_{j-1}$ for the orthogonal projection onto the preceding $q_i$ and let $R_j:=\|(I-P_{j-1})z_j\|_2$. Conditionally on those preceding columns, $DR_j^2$ has the $\chi^2_{D-j+1}$ distribution and $\EE\|P_{j-1}z_j\|_2^2=(j-1)/D$. Since $(R_j-1)^2\le(R_j^2-1)^2$,
\[
 \EE\|Jz_j-e_j\|_\Hh^2
 =\EE\big[\|P_{j-1}z_j\|_2^2+(R_j-1)^2\big]
 \le\frac{j-1}D+\frac{2(D-j+1)+(j-1)^2}{D^2}
 \le\frac{2j}D.
\]
For $j>D$, $e_j$ is orthogonal to $J\Hh_D$, so $\EE\|Jz_j-e_j\|_\Hh^2=2$. Since $E_{\rm emb}^2$ is bounded by the cost of this particular isometry, taking expectations in~\eqref{eq:weighted-column-discrepancy} proves~\eqref{eq:weighted-alignment-mean}.

\paragraph{Step 2. Concentrate the optimal discrepancy.}
First retain finitely many eigenvectors, and write $G:=\sqrt D\,[z_1,\ldots,z_n]$. For each fixed isometry, the triangle inequality shows that its discrepancy changes by at most
\[
 \left(\frac1D\sum_{j\le n}\lambda_j
       \|G_{:,j}-G'_{:,j}\|_2^2\right)^{1/2}
 \le\sqrt{\frac{\lambda_1}D}\,\|G-G'\|_{\mathrm F}.
\]
The infimum over isometries has the same Lipschitz constant. The Gaussian concentration inequality for Lipschitz functions therefore bounds its deviation above its mean by $\sqrt{2\lambda_1t/D}$ with failure probability at most $e^{-t}$. This extends to trace-class $C$ by finite-rank approximation: uniformly over isometries, the change on omitting the columns beyond $n$ is at most
\[
 \left(\sum_{j>n}\lambda_j\|z_j\|_2^2\right)^{1/2}
 +\left(\sum_{j>n}\lambda_j\right)^{1/2},
\]
which tends to zero almost surely and in $L^2(\Omega_{\rm emb})$. Jensen's inequality and~\eqref{eq:weighted-alignment-mean} now give~\eqref{eq:weighted-alignment-tail}. If $C=0$, all discrepancies vanish and the conclusion is immediate.
\end{proof}

\subsection{Proof of Proposition~\ref{prop:embedding-error}}

\begin{proof}

The initialization-stability assertion~\eqref{eq:population-output-stability} was proved in~\eqref{eq:population-error}. We combine it with Gaussian invariance and the preceding alignment estimate.

\paragraph{Step 1. Compare the reference with its principal projection.}
For $P:=P_D$, the covariance identity~\eqref{eq:task-covariance} and
Lemma~\ref{lem:geometry} give
\begin{equation}\label{eq:principal-auxiliary-error}
 \|(I-P){W_{\tin,0}}\|_{L^2(\mu;\Hh)}+\|(I-P){W_{\tout,0}}\|_{\mathrm{op}}
 \le\sqrt{2}\bigl(\operatorname{tr}((I-P)C)\bigr)^{1/2}
 =\left(2\sum_{j>D}\lambda_j\right)^{1/2}.
\end{equation}
Here $\|\cdot\|_{\mathrm{op}}\le\|\cdot\|_{\HS}$ and
$a+b\le\sqrt{2(a^2+b^2)}$.
Let $f_k^P$ start from $P{W_{\tin,0}},P{W_{\tout,0}}$ in $\Hh$, using the same
Gaussian maps and learning rate as the reference. Equation~\eqref{eq:population-error}
bounds $\max_{k\le K}\|f_k^P-f_k\|_{L^2}$ by the right-hand side of
\eqref{eq:main-principal-population}.
Lemma~\ref{lem:subspace} keeps the projected trajectory in $P\Hh$,
and Corollary~\ref{cor:alignment} identifies it with the model initialized by
$JP{W_{\tin,0}},JP{W_{\tout,0}}$ in $\Hh_D$. Their predictions coincide,
proving the principal-embedding bound. The same argument with $P_r$ gives, for every
isometry $J:P_r\Hh\to\Hh_D$,
\begin{equation}\label{eq:rank-r-principal-comparison}
 \max_{k\le K}\|f_{k,r}^J-f_k\|_{L^2(\mu;\Yy)}
 \le C_{B,K}\left(\sum_{j>r}\lambda_j\right)^{1/2}.
\end{equation}

\paragraph{Step 2. Align the full random embeddings and apply the caps.}
Apply Lemma~\ref{lem:weighted-gaussian-alignment} to $W_{\tin,0}$ and $W_{\tout,0}$, with $t:=\log(1/\delta)$. For every comparison isometry $J$, radial projection onto the initialization balls commutes with $J$. The reference embeddings lie in those balls, since $\|W_{\tin,0}(x)\|_\Hh\le B$ almost everywhere and $\|W_{\tout,0}\|_{\HS}\le B$. Nonexpansiveness of Hilbert-space projection, pointwise for the input and in Hilbert--Schmidt norm for the output, gives
\begin{align*}
 \|JW_{\tin,0,D}-W_{\tin,0}\|_{L^2}
 &\le\|JS_DW_{\tin,0}-W_{\tin,0}\|_{L^2},\\
 \|JW_{\tout,0,D}-W_{\tout,0}\|_{\HS}
 &\le\|JS_DW_{\tout,0}-W_{\tout,0}\|_{\HS}.
\end{align*}
The capped model and the reference satisfy the initialization bounds with constant $2B$. Condition on the embeddings and apply Corollary~\ref{cor:alignment}, using $\|\cdot\|_{\mathrm{op}}\le\|\cdot\|_{\HS}$ and $a+b\le\sqrt{2(a^2+b^2)}$. Taking the infimum over $J$ bounds the output error by $C_{B,K}E_{\rm emb}$. Equation~\eqref{eq:weighted-alignment-tail}, followed by $(a+b)^2\le2(a^2+b^2)$, proves~\eqref{eq:main-random-population}. The state estimate follows from the stronger comparison in Corollary~\ref{cor:alignment}. Finally, splitting the weighted sum at $r$ gives~\eqref{eq:weighted-to-rank-bound} and hence~\eqref{eq:main-random-population-rank}.

\end{proof}

To control the Gaussianization error, we also need the following comparison with a rank-$r$ initialization in the original coordinates of $\Hh_D$.

\begin{lemma}[Rank-$r$ comparison in the original coordinates]\label{lem:rank-r-embedding}
Under the embedding assumptions of Proposition~\ref{prop:embedding-error}, let $1\le r\le D$, $0<\delta<1/2$, and $\tau:=\log(3/\delta)$. If $D\ge C(r+\tau)$, then with probability at least $1-\delta$, the polar isometry $J$ of $S_D|_{P_r\Hh}$ satisfies
\begin{equation}\label{eq:capped-gaussian-embedding-error}
 \|W_{\tin,0,D}-JP_rW_{\tin,0}\|_{L^2}
 +\|W_{\tout,0,D}-JP_rW_{\tout,0}\|_{\HS}
 \le C_B\left[\left(\sum_{j>r}\lambda_j\right)^{1/2}
 +\sqrt{\frac{r+\tau}D}\right].
\end{equation}
\end{lemma}

\begin{proof}
Put $\tau:=\log(3/\delta)>1$. Intersect the polar event
\eqref{eq:polar} with the event bounding
\[
 \|S_D(I-P_r){W_{\tin,0}}\|_{L^2}^2+\|S_D(I-P_r){W_{\tout,0}}\|_{\HS}^2
 \le(1+2\sqrt{\tau/D}+2\tau/D)\sum_{j>r}\lambda_j.
\]
Their failure probabilities sum to $2e^{-\tau}+e^{-\tau}=\delta$.
On the intersection, write $T:=S_D|_{P_r\Hh}$ and use its polar isometry $J$.
The two errors split as
\begin{align*}
 S_D{W_{\tin,0}}-JP_r{W_{\tin,0}}&=(T-J)P_r{W_{\tin,0}}+S_D(I-P_r){W_{\tin,0}},\\
 S_D{W_{\tout,0}}-JP_r{W_{\tout,0}}&=(T-J)P_r{W_{\tout,0}}+S_D(I-P_r){W_{\tout,0}}.
\end{align*}
Their retained parts satisfy
\[
 \|(T-J)P_r{W_{\tin,0}}\|_{L^2}^2+\|(T-J)P_r{W_{\tout,0}}\|_{\HS}^2
 \le\|T-J\|_{\mathrm{op}}^2\operatorname{tr}(P_rC)
 \le C_B(r+\tau)/D.
\]
The discarded parts are bounded by the preceding event.
The triangle inequality in the direct sum of the two Hilbert spaces,
followed by $a+b\le\sqrt{2(a^2+b^2)}$, yields
\begin{equation}\label{eq:full-gaussian-embedding-error}
 \|S_D{W_{\tin,0}}-JP_r{W_{\tin,0}}\|_{L^2}
 +\|S_D{W_{\tout,0}}-JP_r{W_{\tout,0}}\|_{\HS}
 \le C_B\left[\left(\sum_{j>r}\lambda_j\right)^{1/2}
                     +\sqrt{\frac{r+\tau}D}\right].
\end{equation}

\paragraph{Apply the caps.}
The comparison embeddings lie in the cap balls:
$\|JP_r{W_{\tin,0}}(x)\|_{2}\le B$ and $\|JP_r{W_{\tout,0}}\|_{\HS}\le B$.
Metric projection onto a closed Hilbert ball is nonexpansive.
Apply it pointwise for the input and in Hilbert--Schmidt norm for the output embedding. This proves~\eqref{eq:capped-gaussian-embedding-error}.
\end{proof}

If $\operatorname{tr}((I-P_D)C)=0$, the initial discrepancies in
\eqref{eq:principal-auxiliary-error} vanish, proving exactness at every
fixed horizon. For empirical $Q=C_\tout$,
$\operatorname{rank}C\le\operatorname{rank}C_\tin+
\operatorname{rank}C_\tout\le2n$.

\subsection{Exact realization}\label{app:exact-horizon}

\begin{lemma}[Exact realization in finite dimension]\label{lem:subspace}\label{lem:finite-horizon}
Consider Gaussian Mean ODE GD from zero increments, and fix $K\ge0$.
Let $S\subset\Hh$ be a closed subspace containing $W_{\tin,0}(x)$ and
$W_{\tout,0}r_k(z)$ for $\mu$-almost every $z$ and every $k<K$,
where $r_k$ are the reference residuals. Restrict the Gaussian maps to
$S$ and initialize the restricted model with
$P_SW_{\tin,0}$ and $P_SW_{\tout,0}$. Then
\[
 h_k^S=h_k,\quad f_k^S=f_k\quad(0\le k\le K),\qquad
 b_k^S=b_k\quad(0\le k<K).
\]
All increments through step $K$ agree after inclusion into $\Hh$;
the internal increments are $S$-valued and measurable with respect to
the Gaussian projections on $S$, and
$\operatorname{ran}\Delta W_{\tin,k}+\operatorname{ran}\Delta W_{\tout,k}
\subset S$.
For $n$ data points one may take~\eqref{eq:exact-rank}, with
$\dim S\le(K+1)n$; if the initial input embedding is linear, then
\[
 \dim S\le\min\{n,d_\tin\}+\min\{Kn,d_\tout\}.
\]
If $S$ contains $\operatorname{ran}W_{\tout,0}$, then
$W_{\tout,0}r_k(z)\in S$ for every $k\ge0$; the same fixed subspace
therefore reproduces all forward and backward passes, without restricting $k<K$.
\end{lemma}

\begin{proof}[Proof of Lemma~\ref{lem:finite-horizon}]
First consider an empirical law and take $S$ from~\eqref{eq:exact-rank}. We show by induction that restricting
the Gaussian maps to $S$ preserves every forward pass through $K$,
and every backward pass used by the first $K$ updates.
The terminal backward signal at step $K$ is not needed for $f_K$.

\paragraph{Step 1. Identify the adjoints on the subspace.}
Write $\mu:=\sum_{i=1}^n\pi_i\delta_{z_i}$, with
$\pi_i>0$ and $\sum_i\pi_i=1$. There are at most $(K+1)n$ vectors
in~\eqref{eq:exact-rank}, so $S$ is finite dimensional.
Its defining residuals are deterministic with respect to the internal
randomness. Let $\mathcal F_S$ be generated by
$\{G_u(w),G_v(w):w\in S\}$.
For $\mathcal F_S$-measurable $\phi\in L^2(\Omega)$ and $w\in S^\perp$,
Gaussian independence gives
\begin{equation}\label{restricted_subspace_HI}
 \langle G_a^*\phi,w\rangle_\Hh
 =\EE[\phi G_a(w)]=\EE[\phi]\EE[G_a(w)]=0,
 \qquad
 G_a^*\phi=(G_a|_S)^*\phi\in S.
\end{equation}
We identify vectors in $S$ with their inclusions in $\Hh$.
If $\Delta a$ is also $S$-valued, then
$A_a^*\phi=G_a^*\phi+\EE[\phi\Delta a]\in S$, with the same
formula for the full and restricted maps.

\paragraph{Step 2. Identify the forward and backward passes.}
At $k=0$ all increments vanish.
Assume they agree between the two models at step $k$, the internal
increments are $S$-valued and $\mathcal F_S$-measurable, and
\[
 \operatorname{ran}\Delta {W_{\tin,k}}\subseteq S,\qquad
 \operatorname{ran}\Delta {W_{\tout,k}}\subseteq S.
\]
The full and restricted forward initial values both equal
${W_{\tin,0}}(x_i)+\Delta {W_{\tin,k}}x_i\in S$.
Along an $S$-valued solution, $p_k$ and $\rho(p_k)$ are
$\mathcal F_S$-measurable, so Step~1 identifies their velocities.
Uniqueness gives
\[
 h_k^S(s,x_i)=h_k(s,x_i)\in S,\quad p_k^S=p_k,\quad
 f_k^S(x_i)=(P_S{W_{\tout,0}}+\Delta {W_{\tout,k}})^*h_k(1,x_i)=f_k(x_i).
\]
Thus $r_k^S(z_i)=r_k(z_i)$. For $k<K$, the definition of $S$ also gives
\begin{equation}\label{eq:finite-horizon-terminal}
 b_k(1,z_i)={W_{\tout,0}}r_k(z_i)+\Delta {W_{\tout,k}}r_k(z_i)\in S.
\end{equation}
The restricted terminal value is the same.
Its $q_k$ and $\rho'(p_k)q_k$ are $\mathcal F_S$-measurable, so the
same adjoint calculation and backward uniqueness yield
$b_k^S(s,z_i)=b_k(s,z_i)\in S$ and $q_k^S=q_k$.

\paragraph{Step 3. Check the four updates.}
The gradient formulas and the increment update~\eqref{eq:exact-population} give
\begin{align*}
 \Delta u_{k+1}(s)&=\Delta u_k(s)-\eta\sum_i\pi_i
          \rho'(p_k(s,x_i))q_k(s,z_i)h_k(s,x_i),\\
 \Delta v_{k+1}(s)&=\Delta v_k(s)-\eta\sum_i\pi_i
          \rho(p_k(s,x_i))b_k(s,z_i),\\
 \Delta {W_{\tin,k+1}}&=\Delta {W_{\tin,k}}-\eta\sum_i\pi_i b_k(0,z_i)\otimes_\Xx x_i,\\
 \Delta {W_{\tout,k+1}}&=\Delta {W_{\tout,k}}-\eta\sum_i\pi_i
                     h_k(1,x_i)\otimes_\Yy r_k(z_i).
\end{align*}
Every right-hand side agrees between the two models.
The first two remain $S$-valued and $\mathcal F_S$-measurable; the last
two retain their stated ranges. This closes the induction for $k<K$.
Step~2 at $k=K$ proves equality of the final predictions.

If ${W_{\tin,0}}$ is linear and the input and output dimensions are finite,
$S\subseteq\operatorname{ran}{W_{\tin,0}}+\operatorname{ran}{W_{\tout,0}}$ gives the
additional dimension bound in the main text.
For $D\ge\dim S$, Corollary~\ref{cor:alignment} transports the restricted
trajectory isometrically to $\Hh_D$, completing the claimed realization.
For a general data law and a closed $S$ satisfying the lemma,
replace the finite sums by data integrals and the finite-dimensional
span by the given closed subspace. A countable orthonormal basis of $S$
generates the completed Gaussian sigma-field. The adjoint identity,
uniqueness and update induction are unchanged. If the entire range of
$W_{\tout,0}$ lies in $S$, apply the statement with any horizon,
and with horizon $K+1$ to include the backward pass at step $K$.
\end{proof}

\section{From finite ResNets to Mean ODEs}\label{sec:resnets}\label{app:finite-comparison}

We condition on the common initial input and output embeddings, whose
bounds are uniform and which are independent of the internal randomness.
We first replace neuron averages by expectations on the same depth
grid, then compare this grid model with the Mean ODE. For the sampling
comparison, groups of neighboring layers supply enough independent
initial rows even when $M=1$.

\subsection{Finite training dynamics}
As throughout the paper, $\Hh_D=\RR^D$ carries its standard $\ell_2$ geometry. All matrix calculations below use its standard orthonormal basis.
Adjoints and tensors use the Hilbert inner products, so
$(a\otimes b)c=\langle b,c\rangle a$.
Write $\widehat\theta:=(\widehat\theta_0,\Delta\widehat\theta)$.
The increment array belongs to
\begin{equation}\label{eq:finite-parameter-space}
 \Theta_{2,L,M,D}:=\HS(\Xx,\Hh_D)\oplus\HS(\Yy,\Hh_D)
                  \oplus(\Hh_D\oplus\Hh_D)^{LM},
\end{equation}
with the Hilbert pairing
\begin{equation}\label{eq:finite-gradient-pairing}
 \langle\Delta\theta,\Delta\vartheta\rangle_{2,L,M,D}
 :=\langle\Delta W_{\tin},\Delta W_{\tin}'\rangle_{\HS}
 +\langle\Delta W_{\tout},\Delta W_{\tout}'\rangle_{\HS}
 +\frac1{LM}\sum_{\ell,j}\sum_{a\in\{u,v\}}
       \langle\Delta a^{j,\ell},\Delta a'^{j,\ell}\rangle_{2}.
\end{equation}
Its induced norm is the discrete counterpart of $\Theta_2$.

A hat denotes the network and $\ell\in\{1,\ldots,L\}$ the block from
node $\ell-1$ to node $\ell$. Initially $\Delta\widehat\theta_{0,D}=0$.
The input embedding may be nonlinear:
$\widehat W_{\tin,k,D}(x):={W_{\tin,0,D}}(x)+\Delta\widehat W_{\tin,k,D}x$;
$\widehat W_{\tout,k,D}:={W_{\tout,0,D}}+\Delta\widehat W_{\tout,k,D}$. The forward equations are
\begin{align}\label{eq:finite-forward}
 \widehat h_{k,D}^0&=\widehat W_{\tin,k,D}(x),&
 \widehat p_k^{j,\ell}&:=\langle\widehat u_k^{j,\ell},
                                     \widehat h_{k,D}^{\ell-1}\rangle_{2},
 &\widehat f_{k,D}&:=\widehat W_{\tout,k,D}^*\widehat h_{k,D}^L,\notag\\
 \widehat h_{k,D}^{\ell}
 &=\widehat h_{k,D}^{\ell-1}
   +\frac1{LM}\sum_{j=1}^M\widehat v_k^{j,\ell}
                                  \rho(\widehat p_k^{j,\ell}).
\end{align}
With $\widehat r_k(z):=\nabla\loss_y(\widehat f_{k,D}(x))$, the adjoint is
\begin{align}\label{eq:finite-backward}
 \widehat b_{k,D}^L&=\widehat W_{\tout,k,D}\widehat r_k,&
 \widehat q_k^{j,\ell}&:=\langle\widehat v_k^{j,\ell},
                                      \widehat b_{k,D}^{\ell}\rangle_{2},
 &
 \widehat b_{k,D}^{\ell-1}
 &=\widehat b_{k,D}^{\ell}
 +\frac1{LM}\sum_{j=1}^M\widehat u_k^{j,\ell}
                \rho'(\widehat p_k^{j,\ell})\widehat q_k^{j,\ell}.
\end{align}
%The outgoing adjoint $\widehat b^\ell$ is required because $(D_h\widehat h^\ell)^*b=b+(LM)^{-1} \sum_j\widehat u^{j,\ell}\rho'(\widehat p^{j,\ell}) \langle\widehat v^{j,\ell},b\rangle_{2}$.
All updates use the same fixed learning rate $\eta$:
\begin{align}\label{eq:finite-updates}
 (\widehat u_{k+1}^{j,\ell},\widehat v_{k+1}^{j,\ell})
 &=(\widehat u_k^{j,\ell},\widehat v_k^{j,\ell})
 -\eta\int\!\big(\widehat h_{k,D}^{\ell-1}
 \rho'(\widehat p_k^{j,\ell})\widehat q_k^{j,\ell},\,
 \widehat b_{k,D}^{\ell}\rho(\widehat p_k^{j,\ell})\big)\,d\mu,\notag\\
 \Delta\widehat W_{\tin,k+1,D}&=\Delta\widehat W_{\tin,k,D}
 -\eta\int\widehat b_{k,D}^0\otimes x\,d\mu,\notag\\
 \Delta\widehat W_{\tout,k+1,D}&=\Delta\widehat W_{\tout,k,D}
 -\eta\int\widehat h_{k,D}^L\otimes\widehat r_k\,d\mu.
\end{align}
Indeed, the $u^{j,\ell}$ variation of the risk in direction $w$ is
$(LM)^{-1}\langle\int\widehat h^{\ell-1}\rho'(\widehat p^{j,\ell})
\widehat q^{j,\ell}\,d\mu,w\rangle_{2}$.
The factor $(LM)^{-1}$ is already in the pairing, so it is absent
from the resulting neuron update.

\subsection{Probability estimates}
For a scalar or Hilbert-valued variable, write
$\|Z\|_{\psi_\alpha}:=\inf\{c>0:\EE\exp((\|Z\|/c)^\alpha)\le2\}$
for $\alpha=1,2$. In orthonormal coordinates the initial rows have
independent centered Gaussian entries with variances $\sigma_u^2,\sigma_v^2$.
The estimates below also hold for independent centered entries with
uniformly bounded $\psi_2$ norms.

\begin{lemma}[Matrices on a group of layers]\label{lem:group-matrices}
Let $U_0,V_0$ be independent $m\times D$ matrices whose entries are
independent, centered, and have $\psi_2$ norm at most $B$.
For deterministic $S\in[-1,1]^{m\times m}$, with probability at least
$1-Ce^{-t}$,
\begin{align}\label{eq:group-matrix}
 \|U_0\|_{\mathrm{op}}+\|V_0\|_{\mathrm{op}}
 &\le C_B(\sqrt m+\sqrt D+\sqrt t),\notag\\
 \|S\odot(U_0V_0^*)\|_{\mathrm{op}}
 &\le C_B[\sqrt{D(m+t)}+m+t],
\end{align}
where $\odot$ means entrywise multiplication. If $m=qM\ge C_B(D+t)$, then, with probability
at least $1-Ce^{-t}$
the matrices divided by $\sqrt m$ have bounded operator norm, and if
\begin{equation}\label{eq:triangular-matrix}
 T_I[(\ell,j),(r,i)]
 :=\frac{\mathbf1_{r<\ell}}{LM}
       \langle u_0^{j,\ell},v_0^{i,r}\rangle_{2}, \quad \text{then}
 \quad \|T_I\|_{\mathrm{op}}\le C_Bq/L.
\end{equation}
The transpose gives the corresponding sum over strictly later layers.
\end{lemma}
\begin{proof}
The rectangular estimate is \cite[Theorem~4.4.5]{vershynin2018high}.
We prove the second bound in \eqref{eq:group-matrix}
using Bernstein's inequality.

\paragraph{Step 1. Bound the scalar products.}
For unit $x,y\in\RR^m$, set $A:=\operatorname{diag}(x)S\operatorname{diag}(y)$.
Then $\|A\|_{\HS}^2=\sum_{i,j}x_i^2S_{ij}^2y_j^2\le1$ and
$x^*[S\odot(U_0V_0^*)]y=\sum_{d=1}^D Z_d$, where
$Z_d:=U_{0,:,d}^*AV_{0,:,d}$ are independent and centered.
The linear-form bound \cite[Proposition~2.6.1]{vershynin2018high},
conditional on $U_{0,:,d}$, gives for $p\ge2$
\[
 \|Z_d\|_{L^p}
 \le C_B\sqrt p\left(\sum_j
     \Big\|\sum_i A_{ij}U_{0,id}\Big\|_{L^p}^2\right)^{1/2}
 \le C_Bp\|A\|_{\HS}\le C_Bp.
\]
Thus $\|Z_d\|_{\psi_1}\le C_B$, and
\cite[Theorem~2.8.1]{vershynin2018high} yields
$\PP(|\sum_dZ_d|>C_B(\sqrt{Dv}+v))\le2e^{-v}$.

\paragraph{Step 2. Operator norms and layer order.}
Two $1/4$-nets of the sphere in $\RR^m$ have at most $9^{2m}$ pairs,
and $\|A'\|_{\mathrm{op}}\le2\max_{x,y\text{ in the nets}}|x^*A'y|$.
Use $v=t+2m\log9+\log2$ in Step 1. This proves the second line
of~\eqref{eq:group-matrix}. With $S_{(\ell,j),(r,i)}:=\mathbf1_{r<\ell}$,
its right side is at most $C_Bm$; division by $LM$ gives
\eqref{eq:triangular-matrix}. Restrictions to any rows preserve the
rectangular bounds because coordinate projections have norm at most one.
\end{proof}

% \begin{lemma}[Hilbert sums and independent-row projections]
% \label{lem:functional-concentration}
% For independent centered variables $Z_i$ in a separable Hilbert space,
% with $\|Z_i\|_{\psi_1}\le v$ and $p\ge2$,
% \begin{equation}\label{eq:hilbert-maximal}
%  \left\|\max_{r\le N}\left\|\sum_{i=1}^rZ_i\right\|\right\|_{L^p}
%  \le Cv(\sqrt{Np}+p).
% \end{equation}
% If the hypotheses hold pointwise in the data with uniform $v$, the
% same bound holds after placing $L^2(\mu)$ inside $L^p(\Omega)$.
% If $F^\ell\in L^2(\mu;\Hh_D)$ depends only on the randomness strictly before
% layer $\ell$ and $\|F^\ell\|_{L^2(\mu)}\le\beta$ almost surely, then,
% for a group $I$ of $q$ layers and $m=qM$, with probability at least $1-e^{-t}$,
% \begin{equation}\label{eq:predictable-projections}
%  \left[\frac1m\sum_{\ell\in I,j}\int
%    |\langle u_0^{j,\ell},F^\ell(z)\rangle_{2}|^2\,d\mu(z)\right]^{1/2}
%  \le C_B\beta\sqrt{1+t/m}.
% \end{equation}
% The reverse-order version uses $v_0$ and functions of strictly later layers.
% \end{lemma}

\begin{lemma}[Hilbert sums and independent-row projections]
\label{lem:functional-concentration}
For independent centered variables $Z_i$ in a separable Hilbert space,
with $\|Z_i\|_{\psi_1}\le v$ and $p\ge2$,
\begin{equation}\label{eq:hilbert-maximal}
 \left\|\max_{r\le N}\left\|\sum_{i=1}^rZ_i\right\|\right\|_{L^p}
 \le Cv(\sqrt{Np}+p).
\end{equation}
If $Z_i=Z_i(z)$ also depends on the data, and the
preceding assumptions hold for $\mu$-almost every $z$
with the same constant $v$, then Minkowski's inequality
gives
\[
\left\|
\left(
\int \max_{r\le N}
\left\|\sum_{i=1}^r Z_i(z)\right\|^2\,d\mu(z)
\right)^{1/2}
\right\|_{L^p(\Omega)}
\le Cv(\sqrt{Np}+p).
\]
Next, let $F^\ell\in L^2(\mu;\Hh_D)$ be random
vectors determined by the initialization of layers
$1,\ldots,\ell-1$, and assume
$\|F^\ell\|_{L^2(\mu)}\le\beta$ almost surely.
In particular, $F^\ell$ is independent of the
initial weights $u_0^{j,\ell}$ of layer $\ell$.
For a group $I$ of $q$ layers and $m=qM$,
with probability at least $1-e^{-t}$,
\begin{equation}\label{eq:predictable-projections}
 \left[\frac1m\sum_{\ell\in I,j}\int
   |\langle u_0^{j,\ell},F^\ell(z)\rangle_{2}|^2
   \,d\mu(z)\right]^{1/2}
 \le C_B\beta\sqrt{1+t/m}.
\end{equation}
The same estimate holds with $v_0^{j,\ell}$
when $F^\ell$ depends only on the initialization
of layers strictly after $\ell$.
\end{lemma}

\begin{proof}
We apply a standard Hilbert martingale inequality to the sums and
condition successively on rows for the projections.

\paragraph{Step 1. Maximal sums and the data norm.}
The bound $\EE\|Z_i\|^j\le2j!v^j$ and
\cite[Theorem~3.1]{pinelis1994optimum}, with Hilbert smoothness constant one,
give, for $0<\lambda v<1$,
\[
 \PP\!\left(\max_{r\le N}\Big\|\sum_{i\le r}Z_i\Big\|\ge u\right)
 \le2\exp\!\left(-\lambda u+\frac{2N\lambda^2v^2}{1-\lambda v}\right).
\]
Choosing $\lambda=cv^{-1}\min(\sqrt{t/N},1)$ and integrating the
resulting tail proves~\eqref{eq:hilbert-maximal}. For
$X(z):=\max_r\|\sum_{i\le r}Z_i(z)\|$, Minkowski gives
$\|\|X\|_{L^2(\mu)}\|_{L^p(\Omega)}
\le(\int\|X(z)\|_{L^p(\Omega)}^2\,d\mu(z))^{1/2}$.
%This proves the data assertion without a covering of the data space.

% \paragraph{Step 2. Projections and conditioning.}
% For $\beta>0$ and $F$ independent of a current row, Jensen with weights
% $\|F(z)\|^2/\beta^2$ and remaining mass at zero gives
% \[
%  \EE\!\left[\left.
%   \exp\!\left(\frac{\int|\langle u_0,F(z)\rangle_{2}|^2\,d\mu}
%                     {C_B\beta^2}\right)\right|F\right]
%  \le1-\frac{\|F\|_{L^2}^2}{\beta^2}
%        +2\frac{\|F\|_{L^2}^2}{\beta^2}\le2.
% \]
% Here the linear-form $\psi_2$ bound controls each unit direction
% $F(z)/\|F(z)\|$. Condition on successive rows and layers; each $F^\ell$
% is already determined when its rows are revealed. The exponential
% moment of the sum of the $m$ squared projections, divided by
% $C_B\beta^2$, is at most $2^m$. Markov at $m\log2+t$ proves
% \eqref{eq:predictable-projections}; $\beta=0$ is immediate.
% If the bound on $\|F^\ell\|_{L^2}$ holds only on another event, apply the
% argument to $F^\ell\mathbf1_{\{\|F^\ell\|_{L^2}\le\beta\}}$ and then
% intersect events. This preserves the dependence on past layers;
% conditioning on the global event would not do so.

\paragraph{Step 2. Projections onto the current layer.}
Fix $\ell$ and condition on the initialization of all
preceding layers. Then $F^\ell$ is fixed and independent
of the rows $u_0^{j,\ell}$.
For $\beta>0$, Jensen's inequality and the sub-Gaussian
bound on the rows give
\[
\begin{aligned}
&\EE\left[
\exp\left(
\frac{1}{C_B\beta^2}
\int|\langle u_0^{j,\ell},F^\ell(z)\rangle_2|^2
\,d\mu(z)\right)
\,\middle|\,F^\ell\right]\\
&\le 1-\frac{\|F^\ell\|_{L^2(\mu)}^2}{\beta^2}
+\frac1{\beta^2}\int
\|F^\ell(z)\|_2^2
\EE\left[
\exp\left(
\frac{|\langle u_0^{j,\ell},
F^\ell(z)/\|F^\ell(z)\|_2\rangle_2|^2}{C_B}
\right)\,\middle|\,F^\ell\right]d\mu(z)\\
&\le 1+\frac{\|F^\ell\|_{L^2(\mu)}^2}{\beta^2}
\le 2.
\end{aligned}
\]
where the integrand is taken to be zero when
$F^\ell(z)=0$.
Reveal the rows successively, in increasing layer order.
Since $F^\ell$ depends only on preceding layers,
the conditional estimate applies at each row.
Iterating conditional expectations gives
\[
\EE\exp\left(
\frac1{C_B\beta^2}
\sum_{\ell\in I}\sum_{j=1}^M
\int|\langle u_0^{j,\ell},F^\ell(z)\rangle_2|^2
\,d\mu(z)
\right)\le 2^m.
\]
Consequently, Markov's inequality yields
\[
\PP\left(
\sum_{\ell\in I}\sum_{j=1}^M
\int|\langle u_0^{j,\ell},F^\ell(z)\rangle_2|^2
\,d\mu(z)
>C_B\beta^2(m\log 2+t)
\right)\le e^{-t},
\]
which proves \eqref{eq:predictable-projections}.

For the reverse-order estimate, use $v_0^{j,\ell}$
and reveal the layers in decreasing order.
If $\|F^\ell\|_{L^2(\mu)}\le\beta$ holds only
on another event, apply the argument to
$F^\ell\mathbf1_{\{\|F^\ell\|_{L^2(\mu)}\le\beta\}}$
and intersect the resulting events.
\end{proof}

\subsection{Depth discretization of the population Mean ODE}

Suppress $D$ from now on. With $s_\ell:=\ell/L$, replace neuron averages
in~\eqref{eq:finite-forward}--\eqref{eq:finite-updates} by expectations
over a single seed. The grid states $\bar h_k^\ell,\bar b_k^\ell$
are deterministic, and its weights $\bar u_k^\ell,\bar v_k^\ell$ are random:
\begin{align}\label{eq:grid-population}
 \bar p_k^\ell&:=\langle\bar u_k^\ell,\bar h_k^{\ell-1}\rangle_{2},&
 \bar q_k^\ell&:=\langle\bar v_k^\ell,\bar b_k^\ell\rangle_{2},\notag\\
 \bar h_k^\ell&=\bar h_k^{\ell-1}
       +L^{-1}\EE[\bar v_k^\ell\rho(\bar p_k^\ell)],&
 \bar h_k^0&={W_{\tin,0}}(x)+\Delta\bar W_{\tin,k}x,\notag\\
 \bar b_k^{\ell-1}&=\bar b_k^\ell
       +L^{-1}\EE[\bar u_k^\ell\rho'(\bar p_k^\ell)\bar q_k^\ell],&
 \bar b_k^L&=\bar W_{\tout,k}\nabla\loss_y(\bar W_{\tout,k}^*\bar h_k^L).
\end{align}
Its updates, starting from zero increments, are
\begin{align}\label{eq:grid-updates}
 \bar u_{k+1}^\ell&=\bar u_k^\ell
       -\eta\int\bar h_k^{\ell-1}\rho'(\bar p_k^\ell)\bar q_k^\ell\,d\mu,\notag\\
 \bar v_{k+1}^\ell&=\bar v_k^\ell
       -\eta\int\bar b_k^\ell\rho(\bar p_k^\ell)\,d\mu,\notag\\
 \Delta\bar W_{\tin,k+1}&=\Delta\bar W_{\tin,k}-\eta\int\bar b_k^0\otimes x\,d\mu,\notag\\
 \Delta\bar W_{\tout,k+1}&=\Delta\bar W_{\tout,k}
       -\eta\int\bar h_k^L\otimes\nabla\loss_y(\bar W_{\tout,k}^*\bar h_k^L)\,d\mu.
\end{align}
Only the marginal law enters these expectations. Thus the same randomness
may be used at every layer when coupling the grid and continuous models.

Consider the population model on the depth grid
defined by \eqref{eq:grid-population}--\eqref{eq:grid-updates}.
To compare its random parameters with those of the
continuous Mean ODE, use the same initial random variables
$u_0,v_0$ in both models at each depth $s_{\ell-1}$.

\begin{lemma}[Depth discretization] \label{lem:depth-euler}
For $L\ge C_{B,K}$, the following hold uniformly
over $k\le K$.
The grid states and embedding increments are bounded
independently of $L,D$ and the data. Moreover, for $p\ge2$,
\[
\begin{aligned}
\max_\ell\bigl(
\|\Delta\bar u_k^\ell\|_{L^p}
+\|\Delta\bar v_k^\ell\|_{L^p}\bigr) +
\operatorname*{ess\,sup}_z\max_\ell
\bigl(\|\bar p_k^\ell(z)\|_{L^p}
+\|\bar q_k^\ell(z)\|_{L^p}\bigr)
\le C_{B,K}\sqrt p.
\end{aligned}
\]
The grid states, predictions and embedding increments
differ from their continuous counterparts by at most
$C_{B,K}/L$, uniformly over the depth grid.
In addition,
\[
\begin{aligned}
\max_\ell\bigl(
\|\Delta\bar u_k^\ell-
       \Delta u_k(s_{\ell-1})\|_{L^p}+ \|\Delta\bar v_k^\ell-
       \Delta v_k(s_{\ell-1})\|_{L^p}
\bigr)
\le \frac{C_{B,K,p}}{L}.
\end{aligned}
\]
\end{lemma}

% \begin{lemma}[Depth discretization]\label{lem:depth-euler}
% For $L\ge C_{B,K}$, the grid states and embedding increments are bounded
% uniformly in $L,D$ and almost every data point. For $p\ge2$,
% \[
%  \max_\ell(\|\Delta\bar u_k^\ell\|_{L^p}
%            +\|\Delta\bar v_k^\ell\|_{L^p})
%  +\mathop{\mathrm{ess\,sup}}_z\max_\ell
%        (\|\bar p_k^\ell(z)\|_{L^p}+\|\bar q_k^\ell(z)\|_{L^p})
%  \le C_{B,K}\sqrt p.
% \]
% The grid states, predictions and embedding increments differ from the
% continuous Mean ODE by at most $C_{B,K}/L$, including the maximum over
% nodes inside $L^2(\mu)$. With the same-seed coupling, the neuron-increment
% error at layer $\ell$ versus depth $s_{\ell-1}$ is at most
% $C_{B,K,p}/L$ in $\max_\ell L^p(\Omega;\Hh_D)$.
% \end{lemma}

\begin{proof}
We first establish bounds independent of the number
of depth steps $L$. We then compare the discrete and
continuous depth equations, whose local difference is
of order $L^{-2}$. Finally, we compare their gradient
descent updates and iterate the resulting estimates
over the $K$ training steps

\paragraph{Step 1. Uniform bounds.}
Put $a_{k,p}:=\max_\ell(\|\Delta\bar u_k^\ell\|_{L^p}
+\|\Delta\bar v_k^\ell\|_{L^p})$, and let $H_k,J_k$ be the maxima over
nodes and data of $\|\bar h_k^\ell\|,\|\bar b_k^\ell\|$.
For $a\in\{u,v\}$, the initial projection $a_0(h):=\langle a_0,h\rangle_{2}$
satisfies $\|a_0(h)\|_{L^p}\le C_B\sqrt p\|h\|$. Thus the full projection
operator has $\Hh_D\to L^2$ norm at most $C_B+a_{k,2}$, and
\[
 \|\bar p_k^\ell\|_{L^p}\le H_k(C_B\sqrt p+a_{k,p}),\qquad
 \|\bar q_k^\ell\|_{L^p}\le J_k(C_B\sqrt p+a_{k,p}).
\]
The state recursions consequently give, with
$c_k:=C_B(1+a_{k,2})^2$,
\[
 \|\bar h_k^\ell\|\le(1+c_k/L)\|\bar h_k^{\ell-1}\|+c_k/L,\qquad
 \|\bar b_k^{\ell-1}\|\le(1+c_k/L)\|\bar b_k^\ell\|.
\]
Their coefficient products are at most $e^{c_k}$. The endpoints are
bounded in terms of the embedding norms, and~\eqref{eq:grid-updates} gives
\[
 a_{k+1,p}\le a_{k,p}+C_{B,\eta}J_k[1+H_k(\sqrt p+a_{k,p})],\qquad
 \|\Delta\bar W_{\tin,k+1}-\Delta\bar W_{\tin,k}\|_{\HS}\le\eta BJ_k,
\]
and $\|\Delta\bar W_{\tout,k+1}-\Delta\bar W_{\tout,k}\|_{\HS}
\le\eta BH_k(1+\|\bar W_{\tout,k}\|_{\mathrm{op}}H_k)$.
Induction first at $p=2$ bounds $H_k,J_k$ and the embeddings.
Then $a_{k+1,p}\le(1+C_{B,K})a_{k,p}+C_{B,K}\sqrt p$ proves the
asserted moments for every $p$.

\paragraph{Step 2. Depth residual and state comparison.}
The depth regularity in Lemma~\ref{lem:moment-propagation} bounds the
forward and backward velocities' Lipschitz constants by $C_{B,K}$.
For the backward coefficient this uses precisely
\[
 \|\rho'(p(s))q(s)-\rho'(p(t))q(t)\|_{L^2}
 \le B\|q(s)-q(t)\|_{L^2}
       +B\|q(t)\|_{L^4}\|p(s)-p(t)\|_{L^4}.
\]
Hence the residual of a continuous trajectory in a grid cell is
$\int_{s_{\ell-1}}^{s_\ell}C_{B,K}(s-s_{\ell-1})\,ds\le C_{B,K}L^{-2}$.
In the adjoint, replacing $b(s_{\ell-1})$ by $b(s_\ell)$ changes its
velocity by at most $C_{B,K}/L$, so the outgoing-adjoint convention
has the same $L^{-2}$ cell residual.

For fixed $k$ and $p\ge2$, set
\[
\begin{aligned}
d_{k,p}
&:=\sum_{a\in\{\tin,\tout\}}
\|\Delta\bar W_{a,k}-\Delta W_{a,k}\|_{\HS}
+\max_{1\le\ell\le L}\sum_{a\in\{u,v\}}
\|\Delta\bar a_k^\ell-\Delta a_k(s_{\ell-1})\|_{L^p},\\
e_h^\ell
&:=\operatorname*{ess\,sup}_z
\|\bar h_k^\ell(z)-h_k(s_\ell,z)\|,\qquad e_b^\ell
:=\operatorname*{ess\,sup}_z
\|\bar b_k^\ell(z)-b_k(s_\ell,z)\|.
\end{aligned}
\]
Subtracting the grid equations, using the preceding product split,
gives
\[
 e_h^\ell\le(1+C/L)e_h^{\ell-1}+Cd_{k,2}/L+C/L^2,\qquad e_h^0\le Bd_{k,2},
\]
\[
 e_b^{\ell-1}\le(1+C/L)e_b^\ell
      +C(d_{k,4}+\max_r e_h^r)/L+C/L^2,\qquad
 e_b^L\le C(d_{k,2}+e_h^L).
\]
Forward and reverse iteration, with $(1+C/L)^L\le e^C$, yield
\begin{equation}\label{eq:grid-state-error}
 \max_\ell(e_h^\ell+e_b^\ell)\le C_{B,K}(d_{k,4}+L^{-1}).
\end{equation}

\paragraph{Step 3. Training comparison.}
For the neuron updates use the exact product differences
\[
 \bar h\rho'(\bar p)\bar q-h\rho'(p)q
 =(\bar h-h)\rho'(\bar p)\bar q
       +h\rho'(\bar p)(\bar q-q)+hq[\rho'(\bar p)-\rho'(p)],
\]
and $\bar b\rho(\bar p)-b\rho(p)
=(\bar b-b)\rho(\bar p)+b[\rho(\bar p)-\rho(p)]$.
Hölder bounds the final $u$ term in $L^p$ by
$C\|q\|_{L^{2p}}\|\bar p-p\|_{L^{2p}}$; all other terms use $L^p$.
The grid updates use $\bar b_k^\ell$, whereas the
continuous updates are evaluated at $s_{\ell-1}$.
By the depth regularity of the backward state, $\|b_k(s_\ell)-b_k(s_{\ell-1})\|
\le C_{B,K} / L$,
and hence
\[
\|\langle v_k(s_{\ell-1}),
b_k(s_\ell)-b_k(s_{\ell-1})\rangle_2\|_{L^p}
\le \frac{C_{B,K,p}}{L}.
\]
Thus this difference contributes at most
$C_{B,K,p}/L$ to the neuron-update estimates.
The embedding updates use $\|a\otimes b\|_{\HS}=\|a\|\|b\|$.
Together with~\eqref{eq:grid-state-error}, these estimates give
\begin{equation}\label{eq:grid-training-error}
 d_{k+1,p}\le C_{B,K,p}(L^{-1}+d_{k,2p}),\qquad d_{0,p}=0.
\end{equation}
Iteration at $p,2p,\ldots,2^Kp$ gives $d_{k,p}\le C_{B,K,p}/L$.
Equation~\eqref{eq:grid-state-error} controls the states uniformly in data,
and $\bar W_{\tout}^*\bar h^L-{W_{\tout}}^*h(1)=(\bar W_{\tout}-{W_{\tout}})^*\bar h^L+{W_{\tout}}^*(\bar h^L-h(1))$
controls the prediction.
\end{proof}

For each $(j,\ell)$, define
$(\bar u_k^{j,\ell},\bar v_k^{j,\ell})$ by the
population parameter updates on the depth grid,
starting from the same initial weights as
$(\widehat u_k^{j,\ell},\widehat v_k^{j,\ell})$.
Since these updates involve only deterministic
population quantities, the pairs
$(\bar u_k^{j,\ell},\bar v_k^{j,\ell})$ are
independent across $(j,\ell)$.
In contrast, the finite-network parameters become
dependent through their shared forward and backward states.

For compactness in the following computations set
$a_k^{j,\ell}:=\rho(p_k^{j,\ell})$ and
$c_k^{j,\ell}:=\rho'(p_k^{j,\ell})q_k^{j,\ell}$, in either system.
For one row, suppress $(j,\ell)$ and let $h_t$ mean $h_t^{\ell-1}$ and
$b_t$ mean $b_t^\ell$. Summing the updates gives the exact identities
\begin{align}\label{eq:finite-histories}
 u_k&=u_0-\eta\sum_{t=0}^{k-1}\int h_t c_t\,d\mu,
 &v_k&=v_0-\eta\sum_{t=0}^{k-1}\int b_t a_t\,d\mu,\notag\\
 v_ka_k(z)&=v_0a_k(z)-\eta\sum_{t=0}^{k-1}
                   \int b_t(z')a_t(z')a_k(z)\,d\mu(z'),\notag\\
 u_kc_k(z)&=u_0c_k(z)-\eta\sum_{t=0}^{k-1}
                   \int h_t(z')c_t(z')c_k(z)\,d\mu(z').
\end{align}
These identities separate an initial matrix from learned terms that can
be estimated by sums of products of scalar coefficients.
Define, at nodes $n=0,\ldots,L$, the centered reference sums
\begin{align}\label{eq:reference-sums}
 F_k^{h,n}&:=\frac1{LM}\sum_{r=1}^{n}\sum_{j=1}^M
    (\bar v_k^{j,r}\bar a_k^{j,r}-\EE\bar v_k^{j,r}\bar a_k^{j,r}),\notag\\
 F_k^{b,n}&:=\frac1{LM}\sum_{r=n+1}^{L}\sum_{j=1}^M
    (\bar u_k^{j,r}\bar c_k^{j,r}-\EE\bar u_k^{j,r}\bar c_k^{j,r}).
\end{align}
In particular, $F_k^{h,\ell-1}$ uses only layers before $\ell$, and
$F_k^{b,\ell}$ uses only layers after $\ell$. They are independent of
the initial row at layer $\ell$ onto which they will be projected.
Choose
\begin{equation}\label{eq:group-length}
 q:=\left\lceil C_B(D+t)/M\right\rceil,\qquad m:=qM.
\end{equation}
Partition the layers into consecutive groups
$I:=\{a+1,\ldots,d\}$ of at most $q$ layers. If the last group has fewer
than $q$ layers, add unused independent rows to its matrices and give
them zero coefficients in every trajectory sum. For a scalar row array
use the common denominator $m$:
\[
 \|w(z)\|_I^2:=\frac1m\sum_{\ell\in I}\sum_{j=1}^M|w^{j,\ell}(z)|^2,
 \qquad
 \|w\|_{L^2(\mu;I)}^2:=\int\|w(z)\|_I^2\,d\mu(z).
\]
For vector arrays replace the absolute value by the hidden norm.
The size assumptions of Proposition~\ref{thm:finite-mean-ode} give
\[
 \frac qL\le\frac{C_B(D+t)}{LM}+\frac1L,
 \qquad \sum_I\frac qL\le1+\frac qL\le2.
\]
Increasing $C_{B,K}$ in those assumptions makes $q/L$ as small as needed
for the absorptions below, while $m\ge C_B(D+t)$ ensures the matrix
bounds on every group.

\begin{lemma}[A common probability event]\label{lem:finite-event}
With probability at least $1-\delta$, all group matrix bounds hold and,
for $k\le K$,
\begin{align}\label{eq:reference-errors}
 &\left\|\max_{0\le n\le L}
       (\|F_k^{h,n}\|+\|F_k^{b,n}\|)\right\|_{L^2(\mu)}
 \le\beta,\qquad \beta:=C_{B,K}\sqrt{Dt/(LM)},\notag\\
 &\left\|\bigl(\langle u_0^{j,\ell},F_k^{h,\ell-1}\rangle_{2}
                   \bigr)_{j,\ell\in I}\right\|_{L^2(\mu;I)}
 +\left\|\bigl(\langle v_0^{j,\ell},F_k^{b,\ell}\rangle_{2}
                   \bigr)_{j,\ell\in I}\right\|_{L^2(\mu;I)}
 \le C_B\beta.
\end{align}
Moreover,
\begin{equation}\label{eq:reference-tail}
 \max_{k,I}\|\bar q_k
 \mathbf1_{\{|\bar q_k|>C_{B,K}\sqrt t\}}\|_{L^2(\mu;I)}
 \le(LM)^{-1/2}\le\beta.
\end{equation}
\end{lemma}

\begin{proof}
We estimate the reference sums, their independent-row projections, and
the feedback tails before intersecting their events.

\paragraph{Step 1. Sums and projections.}
Lemma~\ref{lem:depth-euler} bounds the reference increments and scalar
projections in $\psi_2$ by $C_{B,K}$. In orthonormal coordinates,
$\|\|u_0\|\|_{L^p}^2\le\sum_{d=1}^D\|u_{0,d}\|_{L^p}^2\le C_BDp$,
and likewise for $v_0$. Since
$\|XY\|_{\psi_1}\le C\|X\|_{\psi_2}\|Y\|_{\psi_2}$ without independence,
each centered summand in~\eqref{eq:reference-sums} has $\psi_1$ norm
at most $C_{B,K}\sqrt D$, uniformly in data.
Apply~\eqref{eq:hilbert-maximal} with $N=LM$, $p=t\ge2$, in forward
and reverse layer order, and use Markov at $e$ times the $L^t$ norm:
\[
 \left\|\max_n(\|F_k^{h,n}\|+\|F_k^{b,n}\|)\right\|_{L^2(\mu)}
 \le C_{B,K}\sqrt D\left(\sqrt{\frac t{LM}}+\frac t{LM}\right)
 \le C_{B,K}\sqrt{\frac{Dt}{LM}}
\]
except with probability $Ce^{-t}$; the last inequality uses $t\le LM$.
For the projections, $F_k^{h,\ell-1}$ uses only earlier reference rows,
and $F_k^{b,\ell}$ only later rows. Apply the final truncation argument
of Lemma~\ref{lem:functional-concentration}, then
\eqref{eq:predictable-projections}. Since $m\ge C_Bt$, the threshold is
$C_B\beta$.

\paragraph{Step 2. Tails and the common event.}
The $\psi_2$ bound for $\bar q_k$ implies, for a sufficiently large
constant $C_{B,K}$,
\[
\EE\left[
|\bar q_k(z)|^2
\mathbf1_{\{|\bar q_k(z)|>C_{B,K}\sqrt t\}}
\right]
\le C_{B,K}e^{-2t}.
\]
Averaging in rows and data, then using Markov, gives
\[
 \PP\!\left(
 \|\bar q_k\mathbf1_{\{|\bar q_k|>C_{B,K}\sqrt t\}}
 \|_{L^2(\mu;I)}^2>(LM)^{-1}
 \right)
 \le C_{B,K}LM e^{-2t}.
\]
There are at most $C(K+1)L$ events, including the matrix estimates.
With
$t=\log(C_{B,K}(K+1)L^2M/\delta)$,
their failure probabilities sum to at most $\delta$.
\end{proof}

\subsection{Bounds for the finite trajectories}

\begin{lemma}[Bounds on groups of layers]\label{lem:finite-apriori}
Assume that the matrix bounds
\eqref{eq:group-matrix}--\eqref{eq:triangular-matrix}
hold for every group $I$. Then, for $k\le K$,
the finite forward and backward states are bounded uniformly over
depth and almost every data point, the embedding increments
are bounded in Hilbert--Schmidt norm, and
\begin{equation}\label{eq:group-apriori}
\begin{aligned}
\max_I\Bigg\{&
\operatorname*{ess\,sup}_z
\left(\frac1m\sum_{\ell\in I,j}
\sum_{w\in\{p,q,a,c\}}|\widehat w_k^{j,\ell}(z)|^2\right)^{1/2}\\
&+\left(\frac1m\sum_{\ell\in I,j}
\|\Delta\widehat u_k^{j,\ell}\|_2^2\right)^{1/2}
+\left(\frac1m\sum_{\ell\in I,j}
\|\Delta\widehat v_k^{j,\ell}\|_2^2\right)^{1/2}
\Bigg\}\le C_{B,K},
\end{aligned}
\end{equation}
where $m=qM$. The same estimates hold for the reference
quantities, with hats replaced by bars.
\end{lemma}

\begin{proof}
We proceed by induction over the training steps.
The preceding updates first control the trained parameters.
We then bound the forward and backward states using the matrix
estimates on each group of layers. Finally, we establish
the same bounds for the reference parameters.

\paragraph{Step 1. Bounds on the trained parameters.}
Under the induction hypothesis, the parameter updates give
\[
\begin{aligned}
&\left(\frac1m\sum_{\ell\in I,j}
\|\Delta\widehat u_k^{j,\ell}\|_2^2\right)^{1/2}
\le \eta\sum_{t<k}\int
\max_\ell\|\widehat h_t^{\ell-1}(z)\|
\left(\frac1m\sum_{\ell\in I,j}
|\widehat c_t^{j,\ell}(z)|^2\right)^{1/2}d\mu(z)
\le C_{B,k}.
\end{aligned}
\]
The same argument, replacing $u,h,c$ by $v,b,a$,
bounds the increments of $v$.

By \eqref{eq:group-matrix}, the initial weight matrices
$U_0$ and $V_0$ divided by $\sqrt m$ have uniformly bounded
operator norm when restricted to any group $I$.
For the trained increments, Cauchy--Schwarz gives
\[
\sup_{\|x\|_2=1}
\left(\frac1m\sum_{\ell\in I,j}
|\langle\Delta\widehat u_k^{j,\ell},x\rangle_2|^2\right)^{1/2}
\le\left(\frac1m\sum_{\ell\in I,j}
\|\Delta\widehat u_k^{j,\ell}\|_2^2\right)^{1/2}
\le C_{B,k},
\]
and likewise for $v$. The embedding increments at step $k$
are bounded by the preceding updates. These estimates hold
at $k=0$, when the increments vanish.

\paragraph{Step 2. Forward states and projections.}
Fix $I=\{a+1,\ldots,d\}$ and a data point $z$, and set
\[
A_I:=\left(\frac1m\sum_{\ell\in I,j}
|\widehat a_k^{j,\ell}(z)|^2\right)^{1/2}.
\]
We first bound the variation of the forward state within $I$.
This allows us to control the neuron projections by the state
entering the group, and then propagate the bound to the next group.
The initial matrix bound and Step 1 give, for $a\le n\le d$,
\begin{equation}\label{eq:finite-local-forward}
\|\widehat h_k^n-\widehat h_k^a\|
\le\frac qL\left(C_B+
\left(\frac1m\sum_{\ell\in I,j}
\|\Delta\widehat v_k^{j,\ell}\|_2^2\right)^{1/2}\right)A_I
\le C_{B,k}\frac qL A_I.
\end{equation}
For $r\in I$, define
\[
d^r:=\frac1M\sum_{i=1}^M
\Delta\widehat v_k^{i,r}\widehat a_k^{i,r}.
\]
By Cauchy--Schwarz,
\[
\sum_{r\in I}\|d^r\|
\le q\left(\frac1m\sum_{\ell\in I,j}
\|\Delta\widehat v_k^{j,\ell}\|_2^2\right)^{1/2}A_I
\le C_{B,k}qA_I.
\]
Applying \eqref{eq:group-matrix} to the rows with
$\ell>r$, followed by the triangle inequality, gives
\begin{equation}\label{eq:restricted-drift}
\frac1{L\sqrt m}
\left\|\left(\sum_{r=a+1}^{\ell-1}
\langle u_0^{j,\ell},d^r\rangle_2
\right)_{j,\ell\in I}\right\|_2
\le C_{B,k}\frac qL A_I.
\end{equation}
Expanding the incoming state gives
\[
\langle u_0^{j,\ell},\widehat h_k^{\ell-1}\rangle_2
=\langle u_0^{j,\ell},\widehat h_k^a\rangle_2
+(T_I\widehat a_k)_{j,\ell}
+\frac1L\sum_{a<r<\ell}\langle u_0^{j,\ell},d^r\rangle_2.
\]
Using the bounds for $U_0$, $T_I$, and
\eqref{eq:restricted-drift}, we obtain
\[
\left(\frac1m\sum_{\ell\in I,j}
|\langle u_0^{j,\ell},\widehat h_k^{\ell-1}\rangle_2|^2\right)^{1/2}
\le C_{B,k}\left(\|\widehat h_k^a\|+\frac qL A_I\right).
\]
For the trained increments, Step 1 gives
\[
\left(\frac1m\sum_{\ell\in I,j}
|\langle\Delta\widehat u_k^{j,\ell},
\widehat h_k^{\ell-1}\rangle_2|^2\right)^{1/2}
\le C_{B,k}\max_{a\le n\le d}\|\widehat h_k^n\|.
\]
Combining these estimates with
\eqref{eq:finite-local-forward} yields
\[
\left(\frac1m\sum_{\ell\in I,j}
|\widehat p_k^{j,\ell}(z)|^2\right)^{1/2}
\le C_{B,k}\left(\|\widehat h_k^a(z)\|+\frac qL A_I\right).
\]
Since $|\rho(x)|\le B(1+|x|)$,
\[
A_I\le B+B\left(\frac1m\sum_{\ell\in I,j}
|\widehat p_k^{j,\ell}(z)|^2\right)^{1/2}.
\]
For the chosen group length, $C_{B,k}q/L\le1/2$,
so the term involving $A_I$ can be absorbed. Hence
\[
A_I+\left(\frac1m\sum_{\ell\in I,j}
|\widehat p_k^{j,\ell}(z)|^2\right)^{1/2}
\le C_{B,k}(1+\|\widehat h_k^a(z)\|).
\]
Inserting this into \eqref{eq:finite-local-forward} gives
\[
1+\|\widehat h_k^d(z)\|
\le\left(1+C_{B,k}\frac qL\right)
(1+\|\widehat h_k^a(z)\|).
\]
Iterating over the groups, starting from the input embedding,
proves the forward bounds, since
\[
\prod_I\left(1+C_{B,k}\frac qL\right)
\le\exp\left(C_{B,k}\sum_I\frac qL\right)
\le e^{2C_{B,k}}.
\]

\paragraph{Step 3. Backward states and projections.}
Fix the same group and data point, and set
\[
Q_I:=\left(\frac1m\sum_{\ell\in I,j}
|\widehat q_k^{j,\ell}(z)|^2\right)^{1/2}.
\]
Since $\rho'$ is bounded,
\[
\left(\frac1m\sum_{\ell\in I,j}
|\widehat c_k^{j,\ell}(z)|^2\right)^{1/2}
\le BQ_I.
\]
The backward equation and the matrix bounds give
\[
\max_{a\le n\le d}\|\widehat b_k^n-\widehat b_k^d\|
\le C_{B,k}\frac qL Q_I.
\]
More explicitly,
\[
\widehat b_k^\ell
=\widehat b_k^d
+\frac1{LM}\sum_{\substack{r\in I\\r>\ell},i}
(u_0^{i,r}+\Delta\widehat u_k^{i,r})\widehat c_k^{i,r}.
\]
Projecting onto $v_0^{j,\ell}$ and using $T_I^*$
together with the reverse version of
\eqref{eq:restricted-drift} gives
\[
Q_I\le C_{B,k}\|\widehat b_k^d\|
+C_{B,k}\frac qL Q_I,
\]
where the projection onto $\Delta\widehat v_k^{j,\ell}$
is controlled by Step 1.
Absorbing the last term and iterating over the groups
in reverse order gives the backward bounds.
The terminal condition
$\widehat b_k^L=\widehat W_{\tout,k}\widehat r_k$
is bounded by the forward estimates and the loss assumption.
Finally, the embedding updates satisfy
\[
\begin{aligned}
\|\Delta\widehat W_{\tin,k+1}
-\Delta\widehat W_{\tin,k}\|_{\HS}
&\le\eta B\operatorname*{ess\,sup}_z\|\widehat b_k^0(z)\|,\\
\|\Delta\widehat W_{\tout,k+1}
-\Delta\widehat W_{\tout,k}\|_{\HS}
&\le\eta\operatorname*{ess\,sup}_z\|\widehat r_k(z)\|
\operatorname*{ess\,sup}_x\|\widehat h_k^L(x)\|.
\end{aligned}
\]
This closes the induction over training steps.

\paragraph{Step 4. Reference parameters.}
The depth-grid states $\bar h_k^\ell,\bar b_k^\ell$
are deterministic and have depth increments bounded by
$C_{B,K}/L$. Since
\[
\bar h_k^{\ell-1}
=\bar h_k^a+\sum_{a<r<\ell}
(\bar h_k^r-\bar h_k^{r-1}),
\]
the matrix bounds \eqref{eq:group-matrix}, applied to
the rows with $\ell>r$, give
\[
\left(\frac1m\sum_{\ell\in I,j}
|\langle u_0^{j,\ell},\bar h_k^{\ell-1}\rangle_2|^2\right)^{1/2}
\le C_B\left(\|\bar h_k^a\|
+\sum_{r\in I}\|\bar h_k^r-\bar h_k^{r-1}\|\right)
\le C_{B,K}(1+q/L).
\]
Expanding $\bar b_k^\ell$ from $\bar b_k^d$ in
reverse order gives the analogous bound for
$\langle v_0^{j,\ell},\bar b_k^\ell\rangle_2$.
The reference updates then give
\[
\left(\frac1m\sum_{\ell\in I,j}
\|\Delta\bar u_{k+1}^{j,\ell}\|_2^2\right)^{1/2} \le\eta\sum_{t\le k}\int
\max_\ell\|\bar h_t^{\ell-1}(z)\|
\left(\frac1m\sum_{\ell\in I,j}
|\bar c_t^{j,\ell}(z)|^2\right)^{1/2}d\mu(z).
\]
The same inequality holds with $u,h,c$ replaced
by $v,b,a$. The bounded grid states and the preceding
projection estimates therefore control the reference
increments and the scalar quantities
$\bar p,\bar q,\bar a,\bar c$ by induction over $k$.
This proves \eqref{eq:group-apriori} with hats
replaced by bars.
\end{proof}

\subsection{Forward and backward error estimates}
Work on the event of Lemma~\ref{lem:finite-event}. A difference
$\delta$ means the finite quantity minus its coupled reference quantity.
For states, the reference is the deterministic population state on the
grid; for row variables it is the sampled reference history. Define
\begin{align*}
 F_k:={}&\left\|\max_n\|\delta h_k^n\|\right\|_{L^2(\mu)}
       +\max_I\|\delta p_k\|_{L^2(\mu;I)}
       +\|\Delta\widehat W_{\tin,k}-\Delta\bar W_{\tin,k}\|_{\HS}
       +\|\Delta\widehat W_{\tout,k}-\Delta\bar W_{\tout,k}\|_{\HS},\\
 B_k:={}&\left\|\max_n\|\delta b_k^n\|\right\|_{L^2(\mu)}
       +\max_I\|\delta q_k\|_{L^2(\mu;I)}.
\end{align*}
These are nonnegative numbers. The quantities $F_k^{h,n},F_k^{b,n}$ remain
the centered reference sums~\eqref{eq:reference-sums}. We include the
projection errors in $F_k,B_k$ because the neuron updates depend on
$p_k,q_k$; state errors alone do not directly control those updates.
The maximum over nodes is inside the data norm, whereas the maximum
over groups is outside its corresponding norm. %The proof below respectsthis order throughout.

The only nonlinear coefficient requiring a tail estimate is
$c_k:=\rho'(p_k)q_k$. Subtract it as
\[
 \widehat c_k-\bar c_k
 =\rho'(\widehat p_k)\delta q_k
                   +\bar q_k[\rho'(\widehat p_k)-\rho'(\bar p_k)].
\]
On $|\bar q_k|\le\tau$, use the Lipschitz bound for $\rho'$; on its
complement, use $|\rho'|\le B$. Thus
\begin{equation}\label{eq:coefficient-tail-split}
 |\widehat c_k-\bar c_k|
 \le B|\delta q_k|+B\tau|\delta p_k|
                  +2B|\bar q_k|\mathbf1_{\{|\bar q_k|>\tau\}}.
\end{equation}
In particular,
\[
 \max_I\|\delta c_k\|_{L^2(\mu;I)}
                      \le C_B(B_k+\tau F_k+\beta),
 \qquad
 \max_I\|\delta a_k\|_{L^2(\mu;I)}\le BF_k.
\]
This is the reason why \eqref{eq:reference-tail} was included
in Lemma~\ref{lem:finite-event}.

\begin{lemma}[Comparison through the training steps]\label{lem:finite-comparison}
With the preceding notation and $\tau\ge1$,
\begin{align}\label{eq:triangular-training}
 F_k&\le C_{B,K}\left[\beta+\sum_{j<k}(\tau F_j+B_j)\right],\notag\\
 B_k&\le C_{B,K}\left[\beta+\tau F_k+\sum_{j<k}(\tau F_j+B_j)\right].
\end{align}
\end{lemma}
\begin{proof}
We first control the embedding updates and the current forward states,
then the forward projections. With the forward error known, the backward
comparison follows in reverse depth order. This order avoids a factor
$\tau$ in the depth Gronwall coefficients. Write
\[
 E_{<k}:=\beta+\sum_{j<k}(\tau F_j+B_j).
\]
Constants may depend on the fixed number $K$ of training steps.

\paragraph{Step 1. Embeddings and predictions.}
By the output equation and the uniform state bounds,
\[
 \delta f_j=(\widehat W_{\tout,j}-\bar W_{\tout,j})^*\widehat h_j^L
   +\bar W_{\tout,j}^*\delta h_j^L,\quad
 \|\delta r_j\|_{L^2(\mu)}\le B\|\delta f_j\|_{L^2(\mu)}\le CF_j.
\]
The embeddings coincide initially. Their updates give
\begin{align*}
 \|\Delta\widehat W_{\tin,k}-\Delta\bar W_{\tin,k}\|_{\HS}
 &\le\eta\sum_{j<k}\int\|\delta b_j^0(z)\|\|x\|\,d\mu(z)
 \le C\sum_{j<k}B_j,\\
 \|\Delta\widehat W_{\tout,k}-\Delta\bar W_{\tout,k}\|_{\HS}
 &\le\eta\sum_{j<k}\int
 (\|\delta r_j\|\|\widehat h_j^L\|
 +\|\bar r_j\|\|\delta h_j^L\|)\,d\mu
 \le C\sum_{j<k}F_j.
\end{align*}
Here we used
$\widehat h_j^L\otimes\widehat r_j-\bar h_j^L\otimes\bar r_j
=\widehat h_j^L\otimes\delta r_j+\delta h_j^L\otimes\bar r_j$.
Consequently, both embedding errors and $\|\delta h_k^0\|_{L^2(\mu)}$
are bounded by $CE_{<k}$.

\paragraph{Step 2. Difference of the learned forward terms.}
Fix $I=\{a+1,\ldots,d\}$ and $z$, and set
\[
 A_I(z):=\frac1{\sqrt m}\| (\delta p_k^{j,\ell}(z))_{j,\ell\in I}\|_2.
\]
Subtract the two identities \eqref{eq:finite-histories} and isolate
the initial-weight contribution $v_0\delta a_k$. The remaining term
in layer $r$ is
\begin{align*}
 d_k^{h,r}(z):=-\frac\eta M\sum_{i=1}^M\sum_{t<k}\int
 \bigl[\widehat b_t^r(z')\widehat a_t^{i,r}(z')\widehat a_k^{i,r}(z)-\bar b_t^r(z')\bar a_t^{i,r}(z')\bar a_k^{i,r}(z)\bigr]d\mu(z').
\end{align*}
The bracket is the sum of
\[
 \delta b_t\,\widehat a_t\widehat a_k
 +\bar b_t\,\delta a_t\widehat a_k
 +\bar b_t\,\bar a_t\delta a_k.
\]
Cauchy--Schwarz over the $m=qM$ rows and the uniform bounds
of Lemma~\ref{lem:finite-apriori} control the three contributions,
after division by $q$, by $CB_t$, $CF_t$, and $CA_I(z)$.
Summing over $t<k$ yields
\begin{equation}\label{eq:forward-history-error}
 \frac1q\sum_{r\in I}\|d_k^{h,r}(z)\|
 \le C(E_{<k}+A_I(z)).
\end{equation}

\paragraph{Step 3. Forward states within a group.}
Subtract the centered reference sum, defining
$\varepsilon_k^{h,n}:=\delta h_k^n-F_k^{h,n}$. The finite and
population depth equations then give, for $a\le n\le d$,
\[
 \varepsilon_k^{h,n}=\varepsilon_k^{h,a}
 +\frac1{LM}\sum_{a<r\le n,j}v_0^{j,r}\delta a_k^{j,r}
 +\frac1L\sum_{r=a+1}^n d_k^{h,r}.
\]
Since $|\delta a_k^{j,\ell}|\le B|\delta p_k^{j,\ell}|$,
the matrix bound and \eqref{eq:forward-history-error} imply
\begin{equation}\label{eq:forward-group-error}
 \max_{a\le n\le d}\|\delta h_k^n(z)\|
 \le\|\varepsilon_k^{h,a}(z)\|+\max_n\|F_k^{h,n}(z)\|
 +C\frac qL(E_{<k}+A_I(z)).
\end{equation}

\paragraph{Step 4. Forward projections.}
Set
\[
 R_I^h(z):=\frac1{\sqrt m}\| (\langle u_0^{j,\ell},
 F_k^{h,\ell-1}(z)\rangle_2)_{j,\ell\in I}\|_2.
\]
By \eqref{eq:reference-errors}, $\|R_I^h\|_{L^2(\mu)}\le C\beta$.
Project the equation of Step 3 at node $\ell-1$.
The contributions of the initial weights are
\[
 U_0\varepsilon_k^{h,a},\quad
 (\langle u_0^{j,\ell},F_k^{h,\ell-1}\rangle_2)_{j,\ell},\quad
 T_I\delta a_k,\quad
 \left(\frac1L\sum_{a<r<\ell}\langle u_0^{j,\ell},d_k^{h,r}\rangle_2\right)_{j,\ell}.
\]
Their Euclidean norms divided by $\sqrt m$ are at most, respectively,
$C\|\varepsilon_k^{h,a}\|$, $R_I^h$, $C(q/L)A_I$,
and $C(q/L)(E_{<k}+A_I)$.

By \eqref{eq:finite-histories}, the learned part of each forward
projection equals
$-\eta\sum_{t<k}\int c_t(z')\langle h_t(z'),h_k(z)\rangle\,d\mu(z')$.
Its difference is the integral of
\[
 \delta c_t\langle\widehat h_t,\widehat h_k\rangle
 +\bar c_t\langle\delta h_t,\widehat h_k\rangle
 +\bar c_t\langle\bar h_t,\delta h_k\rangle.
\]
Minkowski, the preceding bound on $\delta c_t$, and the uniform state
bounds show that the normalized Euclidean norm of these row-wise
differences is at most
$C(E_{<k}+\max_{a\le n\le d}\|\delta h_k^n(z)\|)$.
Insert \eqref{eq:forward-group-error} and absorb $C(q/L)A_I$ to get
\begin{align}\label{eq:forward-endpoint}
 A_I(z)&\le C\bigl(\|\varepsilon_k^{h,a}(z)\|+\max_n\|F_k^{h,n}(z)\|
       +R_I^h(z)+E_{<k}\bigr),\notag\\
 \|\varepsilon_k^{h,d}(z)\|
 &\le\|\varepsilon_k^{h,a}(z)\|+C(q/L)(A_I(z)+E_{<k}).
\end{align}

\paragraph{Step 5. Forward iteration over groups.}
Let $0=n_0<\cdots<n_J=L$ be the group endpoints.
Iteration of \eqref{eq:forward-endpoint} gives
\[
 \max_{\nu\le J}\|\varepsilon_k^{h,n_\nu}(z)\|
 \le e^{2C}\left[\|\delta h_k^0(z)\|
 +2C(\max_n\|F_k^{h,n}(z)\|+E_{<k})
 +C\sum_I(q/L)R_I^h(z)\right].
\]
Indeed, $F_k^{h,0}=0$ and
$\prod_I(1+Cq/L)\le e^{C\sum_Iq/L}\le e^{2C}$.
Since $\sum_Iq/L\le2$ and $\|R_I^h\|_{L^2(\mu)}\le C\beta$,
the data norm of this maximum is at most $CE_{<k}$.
The first line of \eqref{eq:forward-endpoint} then gives
$\max_I\|A_I\|_{L^2(\mu)}\le CE_{<k}$.
Finally, \eqref{eq:forward-group-error} controls all intermediate
nodes, using
$\|\sum_I(q/L)A_I\|_{L^2(\mu)}\le2\max_I\|A_I\|_{L^2(\mu)}$.
Together with Step 1, this proves $F_k\le CE_{<k}$.

\paragraph{Step 6. Difference of the learned backward terms.}
Set
\[
 Q_I(z):=\frac1{\sqrt m}\| (\delta q_k^{j,\ell}(z))_{j,\ell\in I}\|_2,
 \quad T_I^q(z):=\frac1{\sqrt m}\|
 (\bar q_k^{j,\ell}(z)\mathbf1_{\{|\bar q_k^{j,\ell}(z)|>\tau\}})_{j,\ell\in I}\|_2.
\]
By \eqref{eq:reference-tail}, $\|T_I^q\|_{L^2(\mu)}\le\beta$,
and \eqref{eq:coefficient-tail-split} gives
\[
 \frac1{\sqrt m}\|(\delta c_k^{j,\ell}(z))_{j,\ell\in I}\|_2
 \le C(Q_I(z)+\tau A_I(z)+T_I^q(z)).
\]
The learned backward contribution from layer $r$ is
\begin{align*}
 d_k^{b,r}(z):=-\frac\eta M\sum_{i=1}^M\sum_{t<k}\int
 \bigl[\widehat h_t^{r-1}(z')\widehat c_t^{i,r}(z')\widehat c_k^{i,r}(z)-\bar h_t^{r-1}(z')\bar c_t^{i,r}(z')\bar c_k^{i,r}(z)\bigr]d\mu(z').
\end{align*}
The bracket is the sum of
$\delta h_t\widehat c_t\widehat c_k
 +\bar h_t\delta c_t\widehat c_k
 +\bar h_t\bar c_t\delta c_k$.
Cauchy--Schwarz over rows bounds their contributions by
$CF_t$, $C(B_t+\tau F_t+\beta)$, and
$C(Q_I(z)+\tau A_I(z)+T_I^q(z))$, respectively.
Therefore,
\begin{equation}\label{eq:backward-history-error}
 \frac1q\sum_{r\in I}\|d_k^{b,r}(z)\|
 \le C(E_{<k}+Q_I(z)+\tau A_I(z)+T_I^q(z)).
\end{equation}
The first two terms use the bound on the current $\widehat c_k$
from Lemma~\ref{lem:finite-apriori}, without an extra factor $\tau$.

\paragraph{Step 7. Backward states and projections within a group.}
Define $\varepsilon_k^{b,n}:=\delta b_k^n-F_k^{b,n}$.
For $a\le n\le d$,
\[
 \varepsilon_k^{b,n}=\varepsilon_k^{b,d}
 +\frac1{LM}\sum_{n<r\le d,j}u_0^{j,r}\delta c_k^{j,r}
 +\frac1L\sum_{r=n+1}^d d_k^{b,r}.
\]
Thus
\[
 \max_{a\le n\le d}\|\delta b_k^n(z)\|
 \le\|\varepsilon_k^{b,d}(z)\|+\max_n\|F_k^{b,n}(z)\|
 +C(q/L)(E_{<k}+Q_I(z)+\tau A_I(z)+T_I^q(z)).
\]
For the projection onto $b_k^\ell$, set
\[
 R_I^b(z):=\frac1{\sqrt m}\| (\langle v_0^{j,\ell},
 F_k^{b,\ell}(z)\rangle_2)_{j,\ell\in I}\|_2.
\]
Then $\|R_I^b\|_{L^2(\mu)}\le C\beta$ by \eqref{eq:reference-errors}.
Projecting the displayed recursion at $\ell$ gives the terms
$V_0\varepsilon_k^{b,d}$, the reference projection,
$T_I^*\delta c_k$, and the reverse learned drift. Their normalized
Euclidean norms are bounded by
\[
 C\|\varepsilon_k^{b,d}\|,\quad R_I^b,\quad
 C(q/L)(Q_I+\tau A_I+T_I^q),\quad
 C(q/L)(E_{<k}+Q_I+\tau A_I+T_I^q).
\]
The learned part of the backward projection is
$-\eta\sum_{t<k}\int a_t(z')\langle b_t(z'),b_k(z)\rangle\,d\mu(z')$.
Its difference equals the integral of
\[
 \delta a_t\langle\widehat b_t,\widehat b_k\rangle
 +\bar a_t\langle\delta b_t,\widehat b_k\rangle
 +\bar a_t\langle\bar b_t,\delta b_k\rangle.
\]
As in Step 4, its normalized Euclidean norm is at most
$C(E_{<k}+\max_{a\le n\le d}\|\delta b_k^n(z)\|)$.
Insert the local state estimate and absorb $C(q/L)Q_I$ to obtain
\begin{align}\label{eq:backward-endpoint}
 Q_I(z)&\le C\bigl(\|\varepsilon_k^{b,d}(z)\|+\max_n\|F_k^{b,n}(z)\|
  +R_I^b(z)+E_{<k}+\tau A_I(z)+T_I^q(z)\bigr),\notag\\
 \|\varepsilon_k^{b,a}(z)\|&\le\|\varepsilon_k^{b,d}(z)\|
 +C(q/L)(E_{<k}+Q_I(z)+\tau A_I(z)+T_I^q(z)).
\end{align}
Only the previously controlled forward error is multiplied by $\tau$;
the coefficient absorbed with $Q_I$ is independent of $\tau$.

\paragraph{Step 8. Backward iteration over groups.}
Since $F_k^{b,L}=0$, the terminal condition gives
\[
 \varepsilon_k^{b,L}=\delta b_k^L
 =(\widehat W_{\tout,k}-\bar W_{\tout,k})\widehat r_k
 +\bar W_{\tout,k}\delta r_k,\quad
 \|\varepsilon_k^{b,L}\|_{L^2(\mu)}\le CF_k.
\]
Iterating \eqref{eq:backward-endpoint} from right to left and using
$\prod_I(1+Cq/L)\le e^{2C}$ gives
\[
 \left\|\sum_I(q/L)(R_I^b+T_I^q+\tau A_I)\right\|_{L^2(\mu)}
 \le C\beta+2\tau\max_I\|A_I\|_{L^2(\mu)}\le C\beta+2\tau F_k.
\]
Since $\|\max_n\|F_k^{b,n}\|\|_{L^2(\mu)}\le\beta$,
we obtain
\[
 \left\|\max_{\nu\le J}\|\varepsilon_k^{b,n_\nu}\|\right\|_{L^2(\mu)}
 \le C(E_{<k}+\tau F_k),\quad
 \max_I\|Q_I\|_{L^2(\mu)}\le C(E_{<k}+\tau F_k).
\]
As in Step 5, the local estimate in Step 7 then controls all nodes.
Thus $B_k\le C(E_{<k}+\tau F_k)$, which completes
\eqref{eq:triangular-training}.
\end{proof}

\begin{proof}[Proof of Proposition~\ref{thm:finite-mean-ode}]
We solve the training recursion and add the independent depth error.

\paragraph{Step 1. Sampling error.}
At $k=0$,~\eqref{eq:triangular-training} gives $F_0\le C\beta$ and
$B_0\le C\tau\beta$. If $F_j\le C_j\tau^j\beta$ and
$B_j\le C_j\tau^{j+1}\beta$ for $j<k$, then $\tau\ge1$ gives
\[
 \beta+\sum_{j<k}(\tau F_j+B_j)
 \le\left(1+2\sum_{j<k}C_j\right)\tau^k\beta.
\]
The forward inequality followed by the backward inequality in
\eqref{eq:triangular-training} proves the induction. Thus, on the common event,
\[
 \max_{k\le K}(F_k+B_k)\le C_{B,K}\tau^{K+1}\beta
 =C_{B,K}t^{(K+2)/2}\sqrt{D/(LM)}.
\]
Prediction errors are at most $CF_k$. The grouping needed only
$m\ge C_B(D+t)$ and small $q/L$, both implied by the theorem's
conditions; $M=1$ remains admissible.

\paragraph{Step 2. Depth error.}
Pointwise in data,
$\max_n\|\widehat h_{k,D}^n-h_{k,D}(n/L)\|
\le\max_n\|\widehat h_{k,D}^n-\bar h_{k,D}^n\|
+\max_n\|\bar h_{k,D}^n-h_{k,D}(n/L)\|$.
Taking $L^2(\mu)$ and using Lemma~\ref{lem:depth-euler} adds $C_{B,K}/L$.
The same triangle inequality applies to adjoints, embeddings and predictions.
This proves~\eqref{eq:finite-mean-ode-rate}.
The proof also permits independent centered orthonormal coordinates with
bounded $\psi_2$ norms, with their corresponding maps $u_0,v_0$ in the
reference Mean ODE.
\end{proof}

\section{Gaussianization}\label{app:gaussianization}
We first prove~\eqref{eq:rank-gaussianization} for initial embeddings in a
fixed subspace. We then apply this estimate to the two embedding constructions in Theorem~\ref{thm:main} and its principal-component variant. Throughout this appendix,
we use the standard $\ell_2$ coordinates of $\Hh_D=\RR^D$, in which the initial weights are independent. Put $m:=3$ in general and $m:=4$ when
all third moments vanish.

\paragraph{The matrix representing the initial subspace.}
Let $S\subset\RR^D$ contain $W_{\tin,0}(x)$ for $\mu$-almost every $x$
and $\operatorname{ran}W_{\tout,0}$. Choose an orthonormal basis
$s_1,\ldots,s_r$ of $S$ and define
\[
 A:=(s_1\ \cdots\ s_r)\in\RR^{D\times r},\qquad
 Av=\sum_{j=1}^rv_js_j,\qquad A^*A=I_r,\qquad P:=AA^*.
\]
Thus $A$ embeds the coordinates in $S$ into the full hidden space,
$A^*$ gives the coordinates of the orthogonal projection onto $S$,
and $P$ is that projection. In particular,
\[
 PW_{\tin,0}(x)=W_{\tin,0}(x),\qquad PW_{\tout,0}=W_{\tout,0}.
\]
For the standard coordinate vector $e_i\in\RR^D$, write
$a_i:=A^*e_i\in\RR^r$; its transpose is the $i$th row of $A$. Then
\begin{equation}\label{eq:row-identities}
 (Av)_i=\langle a_i,v\rangle_2,\qquad
 \sum_{i=1}^Da_ia_i^*=I_r,\qquad
 \sum_{i=1}^D\|a_i\|_2^2=r,\qquad
 \alpha:=\max_i\|a_i\|_2\le1.
\end{equation}
The number $\alpha$ measures the largest projection of a coordinate
axis onto $S$, since $\|a_i\|_2=\|Pe_i\|_2$. It depends on the subspace,
not on its chosen orthonormal basis. The matrix $A$ is fixed during
training; we do not assume that the non-Gaussian states remain in $S$.
If the retained subspace has smaller rank, use its actual rank here.
If $S=\{0\}$, both initial embeddings, all states and all increments
vanish, so the comparison is immediate.

The initialization maps and their adjoints are
\[
 a_0(h):=\sigma_a\langle\xi^a,h\rangle_2,\qquad
 a_0^*\phi=\sigma_a\EE[\xi^a\phi],\qquad a\in\{u,v\}.
\]
By~\cite[Corollary~5.12]{vershynin2012introduction},
$\|a_0(h)\|_{L^p}\le C_B\sqrt p\|h\|_2$ for $p\ge2$, uniformly in $D$.
Thus~\eqref{eq:seed-moments} holds at every finite order.

\subsection{Derivatives of the parameter updates}

\paragraph{Two coordinate estimates.}
For $v_1,\ldots,v_j\in\RR^r$, use
$\|Av\|_2=\|v\|_2$ and $\|Av\|_\infty\le\alpha\|v\|_2$ to obtain
\begin{equation}\label{eq:coordinate-products}
 \sum_{i=1}^D\prod_{\ell=1}^j|(Av_\ell)_i|
 \le\|Av_1\|_2\|Av_2\|_2\prod_{\ell=3}^j\|Av_\ell\|_\infty
 \le\alpha^{j-2}\prod_{\ell=1}^j\|v_\ell\|_2,\qquad j\ge2.
\end{equation}
Likewise, for $w\in\RR^D$,
\begin{equation}\label{eq:weighted-coordinate-products}
 \sum_{i=1}^D|w_i|\prod_{\ell=1}^j|(Av_\ell)_i|
 \le\|w\|_2\|Av_1\|_2\prod_{\ell=2}^j\|Av_\ell\|_\infty
 \le\alpha^{j-1}\|w\|_2\prod_{\ell=1}^j\|v_\ell\|_2,\qquad j\ge1.
\end{equation}
We will keep these coordinate products when differentiating. Replacing
every $|(Av)_i|$ by $\|a_i\|_2\|v\|_2$ before summation would instead
introduce the unnecessary factor $r$ through
$\sum_i\|a_i\|_2^m\le r\alpha^{m-2}$.

\paragraph{Fixing the Gaussian states.}
Lemma~\ref{lem:subspace} gives $h_k^{\rm G},b_k^{\rm G}\in S$.
In this and the next subsection write
\[
 h_k:=A^*h_k^{\rm G},\qquad b_k:=A^*b_k^{\rm G},\qquad
 \sup_{k\le K,s,z}(\|h_k(s,x)\|_2+\|b_k(s,z)\|_2)\le C_{B,K}.
\]
The last bound follows from Lemma~\ref{lem:moment-propagation}.
These states are held fixed in all derivatives below. Only the initial
coordinate vector $\chi:=(\chi^u,\chi^v)\in\RR^D\times\RR^D$ varies.
For $z=(x,y)$ define $\Delta u_0:=0$, $\Delta v_0:=0$ and, successively,
\begin{align}\label{eq:seed-response}
 p_k(s,z;\chi)&:=\sigma_u\langle\chi^u,Ah_k(s,x)\rangle_2
                 +\langle\Delta u_k(s;\chi),h_k(s,x)\rangle_2,\notag\\
 q_k(s,z;\chi)&:=\sigma_v\langle\chi^v,Ab_k(s,z)\rangle_2
                 +\langle\Delta v_k(s;\chi),b_k(s,z)\rangle_2,\notag\\
 \Delta u_{k+1}(s;\chi)&=\Delta u_k(s;\chi)
       -\eta\int h_k(s,x)\rho'(p_k(s,z;\chi))q_k(s,z;\chi)\,d\mu(z),\notag\\
 \Delta v_{k+1}(s;\chi)&=\Delta v_k(s;\chi)
       -\eta\int b_k(s,z)\rho(p_k(s,z;\chi))\,d\mu(z).
\end{align}
Here $\Delta u_k,\Delta v_k$ take values in $\RR^r$ and depend on
$\chi$ only through $A^*\chi^u,A^*\chi^v$. Evaluating at independent
standard Gaussian vectors $g^u,g^v$ gives the Gaussian increments:
$A\Delta a_k(s;g)=\Delta a_k^{\rm G}(s)$, $a\in\{u,v\}$.
Evaluating at $\xi$ will instead define an approximate trajectory for
the non-Gaussian model.

\paragraph{Differentiation with respect to initialization.}
Let $e_i^u:=(e_i,0)$ and $e_i^v:=(0,e_i)$ in $\RR^{2D}$. We use the
ordinary scalar partial derivative (with $t\in\RR$)
\[
 \partial_{\chi_i^a}^{\,j}X(\chi)
 :=\left.\frac{d^j}{dt^j}X(\chi+t e_i^a)\right|_{t=0},
 \qquad a\in\{u,v\}.
\]
The superscript $a$ only selects one of the two coordinate vectors;
it is not a differentiation variable. Neither $A$ nor $h_k,b_k$
is differentiated.

For the replacement argument, let $\zeta$ have independent coordinates,
each with either the corresponding law of $\xi$ or the standard
Gaussian law. Set $\zeta^{-i,a}:=\zeta-\zeta_i^ae_i^a$.
For $t\in\RR$, the vector $\zeta^{-i,a}+t e_i^a$ has its selected
coordinate set to $t$. For deterministic $v\in\RR^r$,
\begin{equation}\label{eq:hybrid-projection}
 \left\|\sum_{j\ne i}\zeta_j^a(Av)_j+t(Av)_i\right\|_{L^p}
 \le C_B\sqrt p\|v\|_2+|t|\,|(Av)_i|
 \le(C_B\sqrt p+|t|)\|v\|_2.
\end{equation}
The other coordinate vector satisfies the same bound without the
$|t|$ term. Expectations here average over the remaining random
coordinates, with $t$ fixed.

Let $\mathcal V\subset\RR^r$ be the closure of the essential ranges
of $h_k,b_k$, $k\le K$, so $\sup_{v\in\mathcal V}\|v\|_2\le C_{B,K}$.
The next estimate expresses each derivative as a bound involving
exactly as many coordinate factors as its order. This will allow us
to sum over $i$ using~\eqref{eq:coordinate-products}.

\begin{lemma}[Bounds for coordinate derivatives]\label{lem:seed-responses}
Let
$X\in\{p_k,q_k,\Delta u_k,\Delta v_k,\rho(p_k),\rho'(p_k)q_k\}$,
$1\le j\le m$, $p\ge2$ finite, and $a\in\{u,v\}$.
There are an integer $N\ge0$ and a finite positive measure $\nu$ on
$\mathcal V^j$, of mass at most $C_{B,K,p,m}$, such that, for all
$t\in\RR$,
\begin{equation}\label{eq:coordinate-envelope}
 \|\partial_{\chi_i^a}^{\,j}X(\zeta^{-i,a}+t e_i^a)\|_{L^p}
 \le(1+|t|)^N\int_{\mathcal V^j}
             \prod_{\ell=1}^j|(Av_\ell)_i|\,d\nu(v_1,\ldots,v_j).
\end{equation}
For vector-valued $X$, the $L^p$ norm uses its Euclidean norm.
The measure may depend on the fixed states, $X,j,p,a,s,z$, but not
on $i,t$ or which coordinates of $\zeta$ are Gaussian. The mass bound
and $N$ are uniform in $s,z$ and independent of $r,D$.
\end{lemma}
\begin{proof}
We first bound the quantities without derivatives. We then differentiate
\eqref{eq:seed-response}: each derivative of an initial projection
contributes one coordinate factor, and the product rule preserves
the total number of these factors.

\paragraph{Step 1. Bound the moments before differentiating.}
All quantities in this step are evaluated at $\zeta^{-i,a}+t e_i^a$.
Bounded $\rho'$, linear growth of $\rho$, and~\eqref{eq:hybrid-projection}
give
\begin{align*}
 \sup_z(\|p_k\|_{L^p}+\|q_k\|_{L^p})
 &\le C_{B,K}(\sqrt p+|t|+\|\Delta u_k\|_{L^p}+\|\Delta v_k\|_{L^p}),\\
 \|\Delta u_{k+1}\|_{L^p}
 &\le\|\Delta u_k\|_{L^p}+C_{B,K}\sup_z\|q_k\|_{L^p},\\
 \|\Delta v_{k+1}\|_{L^p}
 &\le\|\Delta v_k\|_{L^p}+C_{B,K}(1+\sup_z\|p_k\|_{L^p}).
\end{align*}
Starting from zero increments and iterating this linear inequality yields
\begin{equation}\label{eq:coordinate-response-moments}
 \sup_{k\le K,s,z}
 (\|p_k\|_{L^p}+\|q_k\|_{L^p}
       +\|\Delta u_k\|_{L^p}+\|\Delta v_k\|_{L^p})
 \le C_{B,K}(\sqrt p+|t|).
\end{equation}
For a vector with no coordinate fixed, the same calculation gives
$C_{B,K}\sqrt p$. All constants in this step are independent of $p$.

\paragraph{Step 2. Differentiate the projections.}
Assume the derivative estimates hold for the increments at step $k$;
this is true at $k=0$ because the increments vanish. Since the states
are fixed,
\begin{align*}
 \partial_{\chi_i^a}^{\,j}p_k
 &=\mathbf1_{\{j=1,a=u\}}\sigma_u(Ah_k)_i
       +\langle\partial_{\chi_i^a}^{\,j}\Delta u_k,h_k\rangle_2,\\
 \partial_{\chi_i^a}^{\,j}q_k
 &=\mathbf1_{\{j=1,a=v\}}\sigma_v(Ab_k)_i
       +\langle\partial_{\chi_i^a}^{\,j}\Delta v_k,b_k\rangle_2.
\end{align*}
The first terms are bounded by Dirac measures at $h_k$ and $b_k$.
For the other terms use
$\|\langle Y,h_k\rangle_2\|_{L^p}\le\|h_k\|_2\|Y\|_{L^p}$,
and similarly for $b_k$. Thus the claimed form passes from the
increments to $p_k,q_k$.

\paragraph{Step 3. Apply the product and chain rules.}
If factors $Y_\ell$ have derivative orders $j_\ell\ge1$ and bounds
of the form~\eqref{eq:coordinate-envelope}, H\"older gives
\[
 \left\|\prod_{\ell=1}^qY_\ell\right\|_{L^p}
 \le\prod_{\ell=1}^q\|Y_\ell\|_{L^{pq}}
 \le(1+|t|)^{\sum_\ell N_\ell}
 \int_{\mathcal V^{\sum_\ell j_\ell}}
   \prod_{\ell=1}^{\sum_\ell j_\ell}|(Av_\ell)_i|
       \,d(\nu_1\otimes\cdots\otimes\nu_q).
\]
The measure has mass $\prod_\ell\nu_\ell(\mathcal V^{j_\ell})$.
Sums of terms correspond to sums of their measures. Factors with no
derivative are bounded using Step~1.

Writing $\mathcal P_j$ for the partitions of $\{1,\ldots,j\}$, the
chain rule and Leibniz's rule give
\begin{align*}
 \partial_{\chi_i^a}^{\,j}\rho(p_k)
 &=\sum_{\pi\in\mathcal P_j}\rho^{(|\pi|)}(p_k)
           \prod_{I\in\pi}\partial_{\chi_i^a}^{\,|I|}p_k,\\
 \partial_{\chi_i^a}^{\,j}[\rho'(p_k)q_k]
 &=\sum_{\ell=0}^j\binom j\ell
    \partial_{\chi_i^a}^{\,\ell}[\rho'(p_k)]
    \partial_{\chi_i^a}^{\,j-\ell}q_k.
\end{align*}
Apply the first identity also to $\rho'$. Each term has total
derivative order $j$, hence exactly $j$ coordinate factors.
Only $\rho^{(1)},\ldots,\rho^{(m+1)}$ occur; these are bounded by
the assumptions of Theorem~\ref{thm:main}.

\paragraph{Step 4. Differentiate the updates and close the induction.}
The two update derivatives are
\begin{align*}
 \partial_{\chi_i^a}^{\,j}\Delta u_{k+1}
 &=\partial_{\chi_i^a}^{\,j}\Delta u_k
       -\eta\int h_k\,\partial_{\chi_i^a}^{\,j}[\rho'(p_k)q_k]\,d\mu,\\
 \partial_{\chi_i^a}^{\,j}\Delta v_{k+1}
 &=\partial_{\chi_i^a}^{\,j}\Delta v_k
       -\eta\int b_k\,\partial_{\chi_i^a}^{\,j}\rho(p_k)\,d\mu.
\end{align*}
For either integral, Minkowski's inequality gives
\[
 \left\|\int h_k(z)Y(z)\,d\mu(z)\right\|_{L^p}
 \le\int\|h_k(z)\|_2\|Y(z)\|_{L^p}\,d\mu(z).
\]
If $\nu_z$ bounds $Y(z)$, integrate $\|h_k(z)\|_2\nu_z$ against
$\mu$. Its mass is at most $C_{B,K,p,m}$. Use $b_k$ for the other update.
Starting with Dirac measures, finite sums, tensor products and data
integrals give measurable choices independent of $i$ and the
coordinate laws.

For fixed $D$, bounded states and bounded positive-order activation
derivatives also give local polynomial bounds in $\chi$, uniformly
in $z$. Dominated differentiation therefore justifies the displayed
identities. Through $K$ updates, only finitely many products and higher
moment orders occur. The bounds on their masses remain independent
of $r,D$, completing the induction.
\end{proof}

\subsection{Lindeberg replacement}
Suppress $k,s,z$ in the formulas below and set
$\phi_v(\chi):=\rho(p_k(\chi))$ and
$\phi_u(\chi):=\rho'(p_k(\chi))q_k(\chi)$. The vectors
\begin{equation}\label{eq:gaussianization-tests}
 V_a(\chi):=(\sigma_a\chi^a+A\Delta a_k(\chi))\phi_a(\chi)
 \in\RR^D,\qquad a\in\{u,v\},
\end{equation}
are the integrands of the forward and backward velocities. We compare
their expectations in the full space, not only after projection onto $S$.

\begin{lemma}[Comparison of the mean velocities]\label{lem:replacement}
Let $g:=(g^u,g^v)$ have $2D$ independent standard Gaussian coordinates.
Uniformly in $k\le K,s,z$ and $a\in\{u,v\}$,
\begin{equation}\label{eq:replacement-bounds}
 \|\EE V_a(\xi)-\EE V_a(g)\|_2\le C_{B,K}\alpha^{m-2}.
\end{equation}
\end{lemma}
\begin{proof}
We test the vector difference against a unit vector, replace the
$2D$ scalar coordinates one at a time, and cancel the matching Taylor
coefficients.

\paragraph{Step 1. Bound the scalar derivatives.}
Fix $\|w\|_2=1$ and write
\[
 T_w(\chi):=\langle w,V_a(\chi)\rangle_2
 =\big(\sigma_a\langle w,\chi^a\rangle_2
          +\langle A^*w,\Delta a_k(\chi)\rangle_2\big)\phi_a(\chi).
\]
Since $\|A^*w\|_2\le1$, the parenthesized factor has $L^p$ norm at
most $C_{B,K,p}(1+|t|)$ at $\zeta^{-i,a'}+t e_i^{a'}$, $t\in\RR$.
For $a'\in\{u,v\}$, Leibniz's rule is explicitly
\begin{align*}
 \partial_{\chi_i^{a'}}^{\,m}T_w
 &=\sigma_a\langle w,\chi^a\rangle_2\,
                       \partial_{\chi_i^{a'}}^{\,m}\phi_a
   +m\mathbf1_{\{a'=a\}}\sigma_aw_i\,
                       \partial_{\chi_i^{a'}}^{\,m-1}\phi_a\\
 &\quad+\sum_{j=0}^m\binom mj
       \langle A^*w,\partial_{\chi_i^{a'}}^{\,j}\Delta a_k\rangle_2
                     \partial_{\chi_i^{a'}}^{\,m-j}\phi_a.
\end{align*}
Lemma~\ref{lem:seed-responses} and H\"older's inequality consequently give
\begin{align}\label{eq:taylor-moment}
 \EE\big|\partial_{\chi_i^{a'}}^{\,m}T_w(\zeta^{-i,a'}+t e_i^{a'})\big|
  \le(1+|t|)^N\left[
   \int_{\mathcal V^m}\prod_{\ell=1}^m|(Av_\ell)_i|\,d\nu
   +|w_i|\int_{\mathcal V^{m-1}}
                 \prod_{\ell=1}^{m-1}|(Av_\ell)_i|\,d\widetilde\nu\right].
\end{align}
Both measures have bounded mass independent of $i,w$ and the choices
of coordinate laws. In particular, pairing an increment derivative
with $A^*w$ costs at most its Euclidean norm.

\paragraph{Step 2. Replace one coordinate and cancel its moments.}
Take $g$ independent of $\xi$ and order the coordinates as
$u_1,\ldots,u_D,v_1,\ldots,v_D$. In this order,
put $\zeta^{(\ell)}:=(g_1,\ldots,g_\ell,\xi_{\ell+1},\ldots,\xi_{2D})$.
Then
\[
 \EE T_w(\xi)-\EE T_w(g)
 =\sum_{\ell=1}^{2D}
       \EE[T_w(\zeta^{(\ell-1)})-T_w(\zeta^{(\ell)})].
\]
For the coordinate $(i,a')$ being replaced, the two vectors agree
when that coordinate is set to zero; denote this common vector by
$\zeta^{-i,a'}$. For $t\in\RR$, Taylor's formula gives
\begin{align*}
 T_w(\zeta^{-i,a'}+t e_i^{a'})
 =\sum_{j=0}^{m-1}\frac{t^j}{j!}
                    \partial_{\chi_i^{a'}}^{\,j}T_w(\zeta^{-i,a'})+\frac{t^m}{(m-1)!}\int_0^1(1-\theta)^{m-1}
       \partial_{\chi_i^{a'}}^{\,m}T_w(\zeta^{-i,a'}+\theta t e_i^{a'})\,d\theta.
\end{align*}
The coordinate $t$ is independent of $\zeta^{-i,a'}$. The identities
$\EE(\xi_i^{a'})^j=\EE(g_i^{a'})^j$, $0\le j<m$, cancel the polynomial
terms. Moreover, for either choice of $t$,
$\EE[|t|^m(1+|t|)^N]\le C_{B,K,m}$. Thus the remainder is bounded
by the two coordinate integrals in~\eqref{eq:taylor-moment}.

\paragraph{Step 3. Sum the coordinate products.}
Equations~\eqref{eq:coordinate-products} and
\eqref{eq:weighted-coordinate-products} yield
\begin{align*}
 \sum_i\int_{\mathcal V^m}\prod_{\ell=1}^m|(Av_\ell)_i|\,d\nu
 &\le\alpha^{m-2}\int_{\mathcal V^m}\prod_{\ell=1}^m\|v_\ell\|_2\,d\nu
 \le C_{B,K}\alpha^{m-2},\\
 \sum_i|w_i|\int_{\mathcal V^{m-1}}
                  \prod_{\ell=1}^{m-1}|(Av_\ell)_i|\,d\widetilde\nu
 &\le\alpha^{m-2}\int_{\mathcal V^{m-1}}
                  \prod_{\ell=1}^{m-1}\|v_\ell\|_2\,d\widetilde\nu
 \le C_{B,K}\alpha^{m-2}.
\end{align*}
The same measures bound every replacement coordinate. Summing also
over $a'\in\{u,v\}$ and taking the supremum over $\|w\|_2=1$ proves
\eqref{eq:replacement-bounds}.

\end{proof}

\subsection{Stability through gradient descent}
\begin{proof}[Proof of~\eqref{eq:rank-gaussianization}]
We evaluate the fixed Gaussian update functions at the non-Gaussian
coordinates. This gives an approximate trajectory with exact parameter
updates; only its two depth equations have errors. We then apply the
stability estimate on the probability space carrying $\xi$.

\paragraph{Step 1. Define the approximate trajectory.}
Using the functions in~\eqref{eq:seed-response}, set
\[
 \widetilde h_k:=Ah_k=h_k^{\rm G},\qquad
 \widetilde b_k:=Ab_k=b_k^{\rm G},\qquad
 \Delta\widetilde a_k:=A\Delta a_k(\xi),\quad a\in\{u,v\},
\]
and retain $W_{\tin,k}^{\rm G},W_{\tout,k}^{\rm G}$ as the embeddings.
The scalar projections are $\widetilde p_k:=p_k(\xi)$ and
$\widetilde q_k:=q_k(\xi)$. Multiplying the last two equations of
\eqref{eq:seed-response} by $A$ gives
\begin{align*}
 \Delta\widetilde u_{k+1}
 =\Delta\widetilde u_k-\eta\int\widetilde h_k
                       \rho'(\widetilde p_k)\widetilde q_k\,d\mu, \qquad
 \Delta\widetilde v_{k+1}
 =\Delta\widetilde v_k-\eta\int\widetilde b_k
                       \rho(\widetilde p_k)\,d\mu.
\end{align*}
Also $\widetilde f_k:=(W_{\tout,k}^{\rm G})^*\widetilde h_k(1)=f_k^{\rm G}$,
so $\widetilde r_k:=\nabla\loss_y(\widetilde f_k)=r_k^{\rm G}$.
The Gaussian embedding updates can therefore be written
\[
 W_{\tin,k+1}^{\rm G}=W_{\tin,k}^{\rm G}
       -\eta\int\widetilde b_k(0)\otimes x\,d\mu,\qquad
 W_{\tout,k+1}^{\rm G}=W_{\tout,k}^{\rm G}
       -\eta\int\widetilde h_k(1)\otimes\widetilde r_k\,d\mu.
\]
Hence all four parameter updates are exact, corresponding to $e_k^\theta=0$ in~\eqref{eq:stability-forced-equations}.
The moment estimate~\eqref{eq:coordinate-response-moments}, now with
no coordinate fixed, gives
\[
 \sup_{k\le K,s}\big(\|\Delta\widetilde u_k\|_{L^p}
                  +\|\Delta\widetilde v_k\|_{L^p}\big)
 \le C_{B,K}\sqrt p.
\]

\paragraph{Step 2. Identify the two depth errors.}
Since $\widetilde h_k=h_k^{\rm G}$ and
$\widetilde b_k=b_k^{\rm G}$ satisfy the Gaussian depth equations,
\[
 \partial_s\widetilde h_k=\EE V_v(g),
 \qquad
 -\partial_s\widetilde b_k=\EE V_u(g).
\]
On the other hand, evaluating the right-hand sides of the
depth equations at the approximate parameters gives
$\EE V_v(\xi)$ and $\EE V_u(\xi)$, respectively.
Thus~\eqref{eq:stability-forced-equations} holds with
\[
 e_k^h:=\EE V_v(g)-\EE V_v(\xi),\qquad
 e_k^b:=\EE V_u(g)-\EE V_u(\xi),\qquad e_k^\theta:=0.
\]
Lemma~\ref{lem:replacement} gives
\begin{equation}\label{eq:ng-defect}
 \sup_{k\le K,s,z}(\|e_k^h\|_2+\|e_k^b\|_2)
 \le\epsilon:=C_{B,K}\alpha^{m-2}.
\end{equation}
The endpoints in both depth equations are exact because the states,
predictions and embeddings have been kept from the Gaussian model.

\paragraph{Step 3. Apply training stability.}
The actual non-Gaussian trajectory and the approximate one use the
same $u_0,v_0$, initial embeddings and zero initial increments.
Lemma~\ref{lem:moment-propagation} bounds the actual trajectory;
Step~1 bounds the approximate one. Theorem~\ref{thm:population-stability}
therefore applies. Its one-update estimate reads
\[
 d_{k+1,p}\le C_{B,K,p}(d_{k,2p}+\epsilon),\qquad
 d_{k,p}:=\|\Delta\theta_k^{\rm NG}-\Delta\widetilde\theta_k\|_{\Thetainf{p}},
 \qquad d_{0,p}=0.
\]
Iterating with the orders $2,4,\ldots,2^{K+1}$ gives
$d_{k,2}\le C_{B,K}\epsilon$. Its state estimate then yields
\[
 \max_{k\le K}\left[
 \left\|\sup_s(\|h_k^{\rm NG}-\widetilde h_k\|_2
                +\|b_k^{\rm NG}-\widetilde b_k\|_2)\right\|_{L^2(\mu)}
       +\|f_k^{\rm NG}-\widetilde f_k\|_{L^2(\mu)}\right]
 \le C_{B,K}\epsilon.
\]
Since $\widetilde h_k=h_k^{\rm G}$, $\widetilde b_k=b_k^{\rm G}$ and
$\widetilde f_k=f_k^{\rm G}$, this is~\eqref{eq:rank-gaussianization}.
Only the initial subspace was used to define $A$ and $\alpha$; no row
bound on the trained non-Gaussian states was needed.
\end{proof}

\subsection{Completion of the proof}

\begin{proof}[Proof of Theorem~\ref{thm:main}]
For Gaussian internal coordinates, combine
Propositions~\ref{prop:finite-mean-ode} and~\ref{prop:embedding-error},
using~\eqref{eq:main-random-population-rank}, conditioning on the embeddings and splitting the failure probability.
This proves the Gaussian case of Theorem~\ref{thm:main} and
the principal-component variant of Theorem~\ref{thm:main}. The rest of the proof treats non-Gaussian
coordinates and also establishes the principal-embedding remark.
We compare the full population model with a rank-$r$ model, replace
its internal coordinates by Gaussians, and compare it with the Hilbert
reference. We finally add the finite-network error. The actual
initializations are not truncated.

\paragraph{Step 1. Bound the rows for Gaussian embeddings.}
Use the covariance eigenbasis as coordinates on $P_r\Hh$ and put
$\tau:=\log(8/\delta)$. Then $T:=S_D|_{P_r\Hh}=G/\sqrt D$, where
$G\in\RR^{D\times r}$ has independent standard Gaussian entries.
On the event
\[
 s_{\min}(T)\ge\tfrac12,\qquad
 \|T-T(T^*T)^{-1/2}\|_{\mathrm{op}}
       \le C\sqrt{(r+\tau)/D},
\]
provided by Lemma~\ref{lem:geometry}, put
$A:=T(T^*T)^{-1/2}:P_r\Hh\to\Hh_D$.
Its matrix satisfies $A^*A=I_r$ and
$\|a_i\|_2\le2\|G_{i,:}\|_2/\sqrt D$.
Lemma~1 of~\cite{laurent2000adaptive} gives
\[
 \PP\{\|G_{i,:}\|_2^2>r+2\sqrt{ru}+2u\}\le e^{-u}.
\]
Taking $u=\log(4D/\delta)$ and using a union bound over the $D$ rows
shows, with additional failure probability at most $\delta/4$, that
\begin{equation}\label{eq:random-coherence}
 \alpha^2\le C\frac{r+\log(4D/\delta)}D,
 \qquad \alpha^{m-2}\le C N(r,D,\delta).
\end{equation}

\paragraph{Step 2. Compare the Gaussian embeddings with rank $r$.}
In addition to the inequalities in Step~1, use the event
\[
 \|S_D(I-P_r)W_{\tin,0}\|_{L^2}^2
 +\|S_D(I-P_r)W_{\tout,0}\|_{\HS}^2
 \le(1+2\sqrt{\tau/D}+2\tau/D)\sum_{j>r}\lambda_j,
\]
whose probability is bounded in Lemma~\ref{lem:geometry}.
Equation~\eqref{eq:capped-gaussian-embedding-error} then gives
\[
 \|W_{\tin,0,D}-AP_rW_{\tin,0}\|_{L^2}
 +\|W_{\tout,0,D}-AP_rW_{\tout,0}\|_{\HS}
 \le C_B\left[(\textstyle\sum_{j>r}\lambda_j)^{1/2}
                      +\sqrt{\frac{r+\tau}D}\right].
\]
Condition on these embeddings. Compare successively
with the non-Gaussian rank-$r$ model initialized by
$AP_rW_{\tin,0},AP_rW_{\tout,0}$, its Gaussian counterpart, and the
Hilbert reference. Training stability,~\eqref{eq:rank-gaussianization}
and~\eqref{eq:rank-r-principal-comparison} give
\[
 \max_{k\le K}\|f_{k,D}^{\rm NG}-f_k\|_{L^2}
 \le C_{B,K}\left[(\textstyle\sum_{j>r}\lambda_j)^{1/2}
          +\sqrt{(r+\tau)/D}+N(r,D,\delta)\right].
\]

\paragraph{Step 3. Add the finite-network error.}
Conditionally on either choice of embeddings, the sub-Gaussian
version of Proposition~\ref{thm:finite-mean-ode} gives
\[
 \max_{k\le K}\|\widehat f_{k,D}^{\rm NG}-f_{k,D}^{\rm NG}\|_{L^2}
 \le C_{B,K}\left[\frac1L+t^{(K+2)/2}\sqrt{\frac D{LM}}\right].
\]
Add the population bound from the previous steps and allocate failure probability $\delta/4$ to each of the
finite-network estimate, the singular-value estimate, the discarded
embedding norm and the row estimate. Their intersection has probability
at least $1-\delta$ by the union bound. Increasing the absolute
constants inside the logarithms absorbs this allocation and gives
\eqref{eq:main-finite-rate}.
\end{proof}

\subsection{Empirical verification}
\begin{figure}[!htbp]
 \centering
 \includegraphics[width=0.38\linewidth]{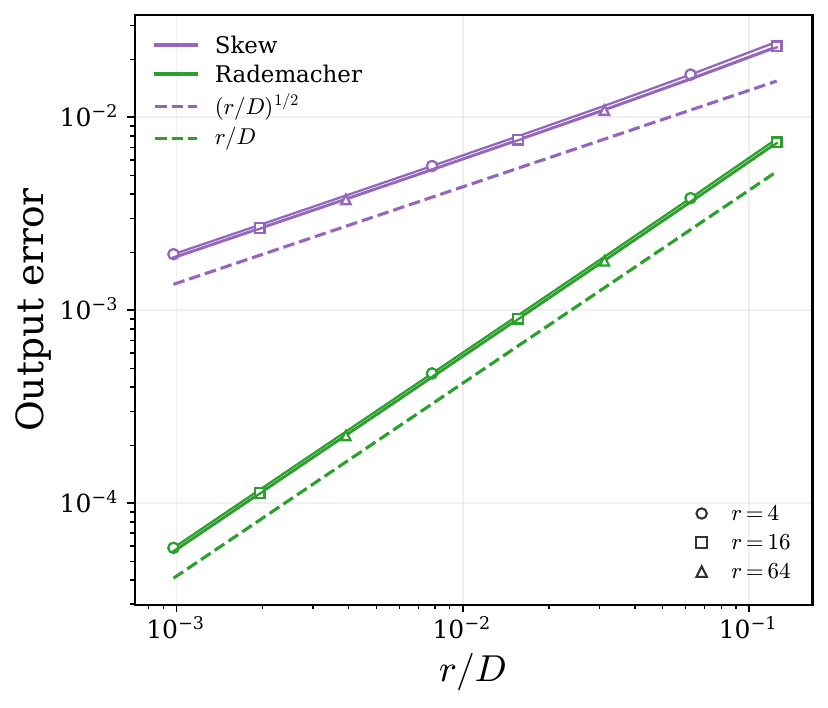}
 \caption{One-step population Gaussianization errors for~\eqref{eq:gaussianization-numerical-ode}. Colors distinguish the initialization laws and markers distinguish $r=4,16,64$. The dashed guides are $\sqrt{r/D}$ for the skew law and $r/D$ for the Rademacher law, suggesting tightness of the bounds of Proposition~\ref{prop:gaussianization}.}
 \label{fig:gaussianization-simple}
\end{figure}

Finally, Figure~\ref{fig:gaussianization-simple} isolates the two Gaussianization rates in Proposition~\ref{prop:gaussianization} in a different setting. The inputs are uniform on $\{\pm e_i:1\le i\le r-1\}\subset\RR^{r-1}$, the target is $Y=1$, and the loss is $(f-Y)^2/2$. For $D=rn$, let the columns of $A\in\RR^{D\times r}$ be normalized indicators of $r$ disjoint blocks of size $n$. Then $A^*A=I_r$ and every row has norm $\alpha=n^{-1/2}=\sqrt{r/D}$. We set $W_{\tin,0}=[Ae_1,\ldots,Ae_{r-1}]$, $W_{\tout,0}=Ae_r$, $v_0=0$, and $\rho(z)=1+\sin z+\cos z$. The coordinates of $u_0$ are independent, with either the standard Gaussian law, the Rademacher law, or the skew law assigning probabilities $1/4$ and $3/4$ to $\sqrt3$ and $-1/\sqrt3$, respectively. All three laws are centered with variance one, and the Rademacher law has zero third moment.

We take one GD step with $\eta=0.4$. Initially the output is zero and the gradients of $u$, $W_{\tin}$ and $W_{\tout}$ vanish. Writing $Z=A^*u_0$ and $q(Z)=(r-1)^{-1}\sum_{j=1}^{r-1}(1+\cos Z_j)$, the update is $v_1=\eta Ae_rq(Z)$. By permutation symmetry, the ensuing forward pass has the form $h_1(s;\pm e_i)=A(\pm e_i+c_\pm(s)e_r)$, where
\begin{equation}\label{eq:gaussianization-numerical-ode}
 c_\pm'(s)=\eta\,\EE\!\left[q(Z)\rho\big(\pm Z_1+c_\pm(s)Z_r\big)\right],
 \qquad c_\pm(0)=0.
\end{equation}
We compare the output $c_\pm(1)$ with its Gaussian counterpart $c_{\rm G}(1)$ through $\big(\tfrac12\sum_\pm|c_\pm(1)-c_{\rm G}(1)|^2\big)^{1/2}$, which here also equals the terminal hidden-state error in $L^2(\mu;\Hh_D)$. Each $Z_j$ is a normalized sum of $n$ independent coordinates, so the expectations in~\eqref{eq:gaussianization-numerical-ode} can be computed exactly and it only remains to integrate two scalar ODEs. The observed errors indicate the tightness of Proposition~\ref{prop:gaussianization}.%We use $r\in\{4,16,64\}$ and $n=2^j$ for $j=3,\ldots,10$. Over $n\ge64$, the fitted slopes against $r/D$ are $0.506$--$0.508$ for the skew law and $1.0010$--$1.0012$ for the Rademacher law, consistent with the respective rates $\sqrt{r/D}$ and $r/D$.

\end{document}